\documentclass[11pt]{article}

\usepackage[preprint]{acl}

\usepackage{times}
\usepackage{latexsym}
\usepackage{siunitx}
\usepackage[T1]{fontenc}

\usepackage[utf8]{inputenc}
\usepackage{dblfloatfix}
\usepackage{microtype}

\usepackage{inconsolata}

\usepackage{graphicx}
\usepackage{xspace}
\usepackage{booktabs}
\usepackage{multirow}
\usepackage{wrapfig}
\usepackage{amsmath}
\usepackage{amssymb}
\usepackage{pifont}
\usepackage{colortbl}
\usepackage{enumitem}
\usepackage{comment}
\usepackage{makecell}
\usepackage{caption}
\makeatletter
\def\input@path{{./}{../}{latex/}}
\makeatother
\usepackage{lipsum}
\graphicspath{{./}{../}{latex/}}

\title{Apples on the Table? Evaluating Text-Guided 3D Scene Synthesis \\ via Fine-Grained Constraint Verification}

\author{
    \textbf{Minseok Kang}\textsuperscript{1*} \quad
    Dongwook Choi\textsuperscript{1*} \quad
    Gyeom Hwangbo\textsuperscript{2*} \\
    \textbf{Seungwon Lim}\textsuperscript{\textbf{1}} \quad
    \textbf{Kai Tzu-iunn Ong}\textsuperscript{\textbf{1}} \quad
    \textbf{Jinyoung Yeo}\textsuperscript{\textbf{1$\dagger$}} \\
    \\
    \textsuperscript{1}Department of Artificial Intelligence, Yonsei University \quad
    \textsuperscript{2}POSTECH \\
    \texttt{\{minskq,dongwook,jinyeo\}@yonsei.ac.kr} \\
    \texttt{gyeomhwangbo@postech.ac.kr}
}

\newcommand\blfootnote[1]{%
  \begingroup
  \renewcommand\thefootnote{}%
  \footnote{#1}%
  \addtocounter{footnote}{-1}%
  \endgroup
}

\begin{document}
\maketitle
\blfootnote{%
\begin{tabular}[t]{@{}l@{}}
\textsuperscript{*} Equal contribution \\
\textsuperscript{$\dagger$} Corresponding author
\end{tabular}%
}
\definecolor{green}{RGB}{36, 214, 36}
\definecolor{red}{RGB}{235, 30, 30}
\definecolor{lightredshade}{HTML}{dea9a9}
\definecolor{lightgreenshade}{HTML}{bce3bd}
\definecolor{lightblueshade}{HTML}{cacbe8}
\definecolor{MyDarkBlue}{rgb}{0,0.08,1}
\definecolor{MyDarkGreen}{rgb}{0.02,0.6,0.02}
\definecolor{MyDarkRed}{rgb}{0.8,0.02,0.02}
\definecolor{MyDarkOrange}{rgb}{0.40,0.2,0.02}
\definecolor{MyPurple}{RGB}{111,0,255}
\definecolor{MyRed}{rgb}{1.0,0.0,0.0}
\definecolor{MyGold}{rgb}{0.75,0.6,0.12}
\definecolor{MyDarkgray}{rgb}{0.66, 0.66, 0.66}

\definecolor{MyYellow}{rgb}{254, 246, 170}
\definecolor{MyBlue}{rgb}{170, 217, 251}
\definecolor{LuneBlue}{rgb}{0.11, 0.11, 0.43}
\newcommand{\colorDelta}[1]{%
  \ifnum#1>50\relax\cellcolor{green!65}+{#1}\%\else%
  \ifnum#1>40\relax\cellcolor{green!40}+{#1}\%\else%
  \ifnum#1>30\relax\cellcolor{green!30}+{#1}\%\else%
  \ifnum#1>20\relax\cellcolor{green!20}+{#1}\%\else%
  \ifnum#1>10\relax\cellcolor{green!10}+{#1}\%\else%
  \ifnum#1>0\relax\cellcolor{green!5}+{#1}\%\else%
  #1\%\fi\fi\fi\fi\fi\fi%
}
\newcommand{\greencheck}{\textcolor{green}{\ding{51}}}
\newcommand{\redcross}{\textcolor{red}{\ding{55}}}
\newcommand{\icon}[1]{\raisebox{-0.15em}{\includegraphics[height=1em]{icons/#1}}}

\newcommand{\bench}{\textsc{LEGO}}
\newcommand{\eval}{\textsc{LEGO-Eval}}
\newcommand{\jy}[1]{\textcolor{MyDarkBlue}{[Jinyoung: #1]}}
\newcommand{\sy}[1]{\textcolor{MyDarkBlue}{[Shunyu: #1]}}
\newcommand{\hj}[1]{\textcolor{MyPurple}{[HJ: #1]}}
\newcommand{\sj}[1]{\textcolor{LuneBlue}{[SJ: #1]}}

\newcommand{\mcal}[1]{{\cal{#1}}}
\newcommand{\calA}{\mbox{${\cal A}$}}
\newcommand{\calB}{\mbox{${\cal B}$}}
\newcommand{\calC}{\mbox{${\cal C}$}}
\newcommand{\calD}{\mbox{${\cal D}$}}
\newcommand{\calE}{\mbox{${\cal E}$}}
\newcommand{\calF}{\mbox{${\cal F}$}}
\newcommand{\calG}{\mbox{${\cal G}$}}
\newcommand{\calH}{\mbox{${\cal H}$}}
\newcommand{\calI}{\mbox{${\cal I}$}}
\newcommand{\calJ}{\mbox{${\cal J}$}}
\newcommand{\calK}{\mbox{${\cal K}$}}
\newcommand{\calL}{\mbox{${\cal L}$}}
\newcommand{\calM}{\mbox{${\cal M}$}}
\newcommand{\calN}{\mbox{${\cal N}$}}
\newcommand{\calO}{\mbox{${\cal O}$}}
\newcommand{\calP}{\mbox{${\cal P}$}}
\newcommand{\calQ}{\mbox{${\cal Q}$}}
\newcommand{\calR}{\mbox{${\cal R}$}}
\newcommand{\calS}{\mbox{${\cal S}$}}
\newcommand{\calT}{\mbox{${\cal T}$}}
\newcommand{\calU}{\mbox{${\cal U}$}}
\newcommand{\calV}{\mbox{${\cal V}$}}
\newcommand{\calW}{\mbox{${\cal W}$}}
\newcommand{\calX}{\mbox{${\cal X}$}}
\newcommand{\calY}{\mbox{${\cal Y}$}}
\newcommand{\calZ}{\mbox{${\cal Z}$}}


\newcommand{\todocccc}[2]{{\textcolor{#1}{[[#2]]}}}
\newcommand{\todogreen}[1]{\todocccc{green}{[[#1]]}}
\newcommand{\haejuu}[1]{\todogreen{haejuu: #1}}
\newcommand{\yeo}[1]{\textcolor{purple}{#1}}
\newcommand{\dragon}[1]{\textcolor{brown}{#1}}
\newcommand{\bwoo}[1]{\textcolor{olive}{#1}}
\newcommand{\kyle}[1]{\textcolor{blue}{#1}} 
\newcommand{\cheris}[1]{\textcolor{teal}{#1}}
\newcommand{\chatgpt}[1]{\textcolor{gray}{#1}}
\newcommand{\minus}[1]{\textcolor{red}{#1}}
\newcommand{\plus}[1]{\textcolor{ForestGreen}{#1}}
\newcommand{\person}{{$\mathbb{P}$}}
\newcommand{\thought}{$\mathbb{Z}_{\mathbb{P}}$}
\newcommand{\ty}[1]{\textcolor{darkgreen}{[TY: #1]}}

\newcommand{\se}{{\it SE}}%
\newcommand{\eg}{{\it e.g.},~}%
\newcommand{\ie}{{\it i.e.},~}%
\newcommand{\etal}{{\it et al.}}%
\newcommand{\etc}{{\it etc}}%

\newcommand{\worldmodel}{$\mathcal{W}_{\theta}$}
\newcommand{\ours}{\textsc{Think-and-Execute}\xspace}
\newcommand{\coffeegym}{\textsc{Coffee-Gym}\xspace}
\newcommand{\cf}{\textsc{Coffee}\xspace}
\newcommand{\editevalbf}{\textbf{\textsc{CoffeeEval}}\xspace}
\newcommand{\coffeewemoji}{\coffee\xspace\cf}
\newcommand{\coffeewemojibf}{\coffee\xspace\textbf{\cf}}
\newcommand{\cfwemoji}{\coffee\xspace\cf}

\newcommand{\bluetext}[1]{\textcolor{blue}{#1}}
\newcommand{\redtext}[1]{\textcolor{red}{#1}}
\newcommand{\greenText}[1]{\textcolor{darkgreen}{#1}}
\newcommand{\graytext}[1]{\textcolor{gray}{#1}}

\begin{abstract} 
Accurately synthesizing 3D scenes from user-provided text descriptions is crucial for developing embodied agents.
Despite the importance of scene-description alignment, existing evaluation methods for such text-guided 3D scene synthesis either capture only coarse similarity between the synthesized scene and the user description, or ignore the spatial reasoning for verifying object placement. None of them addressed the fine-grained constraints (\eg X needs to be in the scene in a Y manner) implied by the description from users.
To address this, we introduce \bench, a benchmark dataset that pairs each user description with human-annotated constraints and a reference scene, and \eval, an evaluation framework that decomposes a description into atomic constraints and verifies each one using tools that ground textual references to 3D objects and reason about their spatial relationships. 
We show that (i) \eval~evaluates misalignment far more accurately than existing methods and (ii) current scene synthesis approaches achieve at most 10\% success rate in \eval. Our code and dataset are publicly available at \url{https://gyeomh.github.io/LEGO-Eval/}.


\end{abstract}
\section{Introduction}

Training embodied agents in simulators with realistic 3D environments is necessary for narrowing the sim-to-real gap~\citep{shridhar2020alfred, li2023behavior, puig2024habitat, Genesis, mittal2025isaac}.
For generalization to the real world, such realistic scenes are required at large scales, which is infeasible to achieve solely by manual crafting~\citep {deitke2022, xia2026sage}.
A growing body of work has thus explored automated text-guided 3D scene synthesis~\citep{feng2023layoutgpt, ccelen2025design, yang2024holodeck, sun2025layoutvlm, che2026mansion}, which produces 3D scenes, including their layout, objects, and spatial arrangement based on the text descriptions from users.


\begin{figure}[t]
    \centering
    \includegraphics[width=0.95\columnwidth]{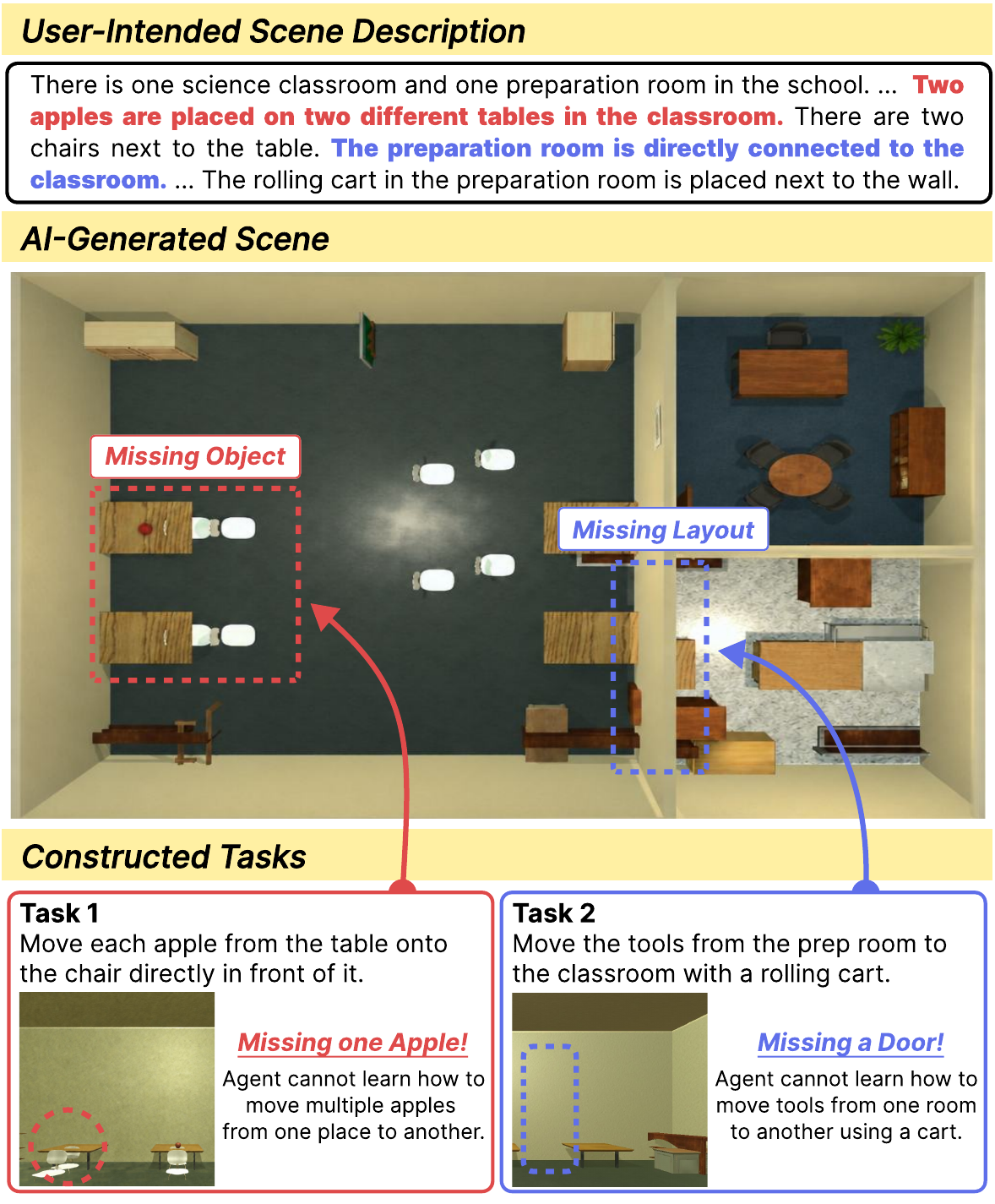}
    \caption{An example of misalignment in text-guided 3D scene synthesis. The generated scene spawns only one of two apples and omits the door, making the constructed tasks unexecutable and unlearnable.
    \vspace{-1.5em}}
    \label{fig:motivation}
\end{figure}


While it is important to generate scenes that accurately align with the user description, current methods often fail to do so~\citep{ling2025scenethesis}.
Such misalignment easily arises because the scene description often imposes diverse, fine-grained constraints on the scene\textemdash from individual object properties to spatial relations among them\textemdash that must be jointly satisfied during synthesis.
However, training on such flawed scenes introduces misleading supervision (Figure~\ref{fig:motivation}), shifting the training distribution away from the user's intent and degrading agent performance~\citep{nasiriany2026robocasa365}.
To prevent this and improve scene synthesis, it is necessary to conduct fine-grained diagnosis of where and why such scenes fail to align with user descriptions.

However, existing efforts toward such fine-grained diagnosis remain limited.
Recent scene synthesis methods adopt CLIPScore~\citep{hessel2021clipscore} or VLM-based judges~\citep{li2024topviewrs} for evaluation, but
CLIPScore captures only coarse-grained text-scene similarity and VLM-based judges lack the 3D spatial reasoning required to verify object placement~\citep{scenecritic}.
Furthermore, no existing dataset for text-guided 3D scene synthesis pairs text descriptions with the fine-grained constraints they imply, a foundation needed to evaluate and improve the field.

To address these, we introduce \textbf{\bench} (\textbf{L}anguage-guided \textbf{E}nvironment
\textbf{G}eneration for emb\textbf{O}died agents), a human-curated dataset for evaluating text-guided 3D scene synthesis.
\bench~contains fine-grained natural language descriptions with corresponding constraints, and a manually constructed scene.
It also contains rich scene metadata (including positions, orientations, scales, and material properties of scene components), making \bench~a valuable resource not only for evaluation but also for the broader 3D scene synthesis community.

Alongside \bench, we propose \textbf{\eval}, an evaluation framework that diagnoses scene-description alignment in text-guided 3D scene synthesis.
\eval~systematically decomposes the scene description into individual constraints and verifies each against the synthesized scene using a diverse set of tools, enabling precise diagnosis of the misalignment.
This tool-based design moves \eval~beyond vision-only judgments, turning synthesized scenes into a reliable foundation for embodied agent training.
Our contributions are summarized as follows:
\vspace{-0.5em}
\begin{itemize}
    \item We introduce \bench, a human-curated dataset that pairs each natural-language scene description with fine-grained annotated constraints and a reference scene for text-guided 3D scene synthesis evaluation.

    \item We propose \eval, a tool-augmented evaluation framework that decomposes each scene description into individual constraints and verifies them one by one, enabling constraint-level diagnosis of misalignment.

    \item We show that \eval~delivers substantially more accurate fine-grained evaluation than existing methods, and reveal that current synthesis approaches fall short of satisfying fine-grained scene descriptions.
\end{itemize}

\section{Background}
\label{sec:background}

\paragraph{Text-guided 3D scene synthesis.}

The goal of text-guided 3D scene synthesis is to construct a 3D environment that faithfully reflects a user's intent.
To this end, users express their requirements through a natural language description $l$.
A synthesis method $\mathcal{F}$ processes $l$ and produces a structured scene representation $\mathcal{S}=\mathcal{F}(l)$.

\paragraph{Why faithful synthesis matters.}

Using simulators, we train and evaluate embodied agents on tasks curated within diverse 3D scenes.
Since tasks are inherently grounded in the scene, users specify their target task distribution through a set of scene descriptions $l$, and construct or synthesize the scenes $\mathcal{S}$ accordingly~\citep{yang2024holodeck, xia2026sage}.
Any misalignment between $\mathcal{S}$ and $l$ thus shifts the task distribution the agent is exposed to, undermining the agent's ability to learn the intended behaviors.
Such misalignment makes any observed success or failure uninterpretable, confounding agent failures with environment errors.
\graytext{
}

\graytext{
}


\vspace{-2.5em}
\paragraph{Challenges in evaluating text-guided 3D scene synthesis.}

However, evaluating whether the scene $\mathcal{S}$ align with the description $l$ is inherently difficult, as it requires verifying every constraint encoded in $l$.
Grounding these constraints in a 3D scene is challenging, as it requires identifying the relevant scene components and reasoning about their properties and relationships—for instance, assessing \textit{"a blue chair placed next to a black desk"} requires multi-hop grounding across object identity, attributes, and spatial relationships. 
This underscores the need for an evaluation framework capable of verifying diverse constraints through explicit, fine-grained grounding.


\section{\bench: Dataset for Text-Guided 3D Scene Synthesis Evaluation}
\label{sec:LEGO-Bench}

In this section, we introduce \textbf{\bench}, a human-curated dataset for fine-grained evaluation of text-guided 3D scene synthesis.

\subsection{Dataset Overview} 
\label{ssec:dataset_overview}

\begin{figure}[t]
    \centering
    \includegraphics[width=0.95\linewidth]{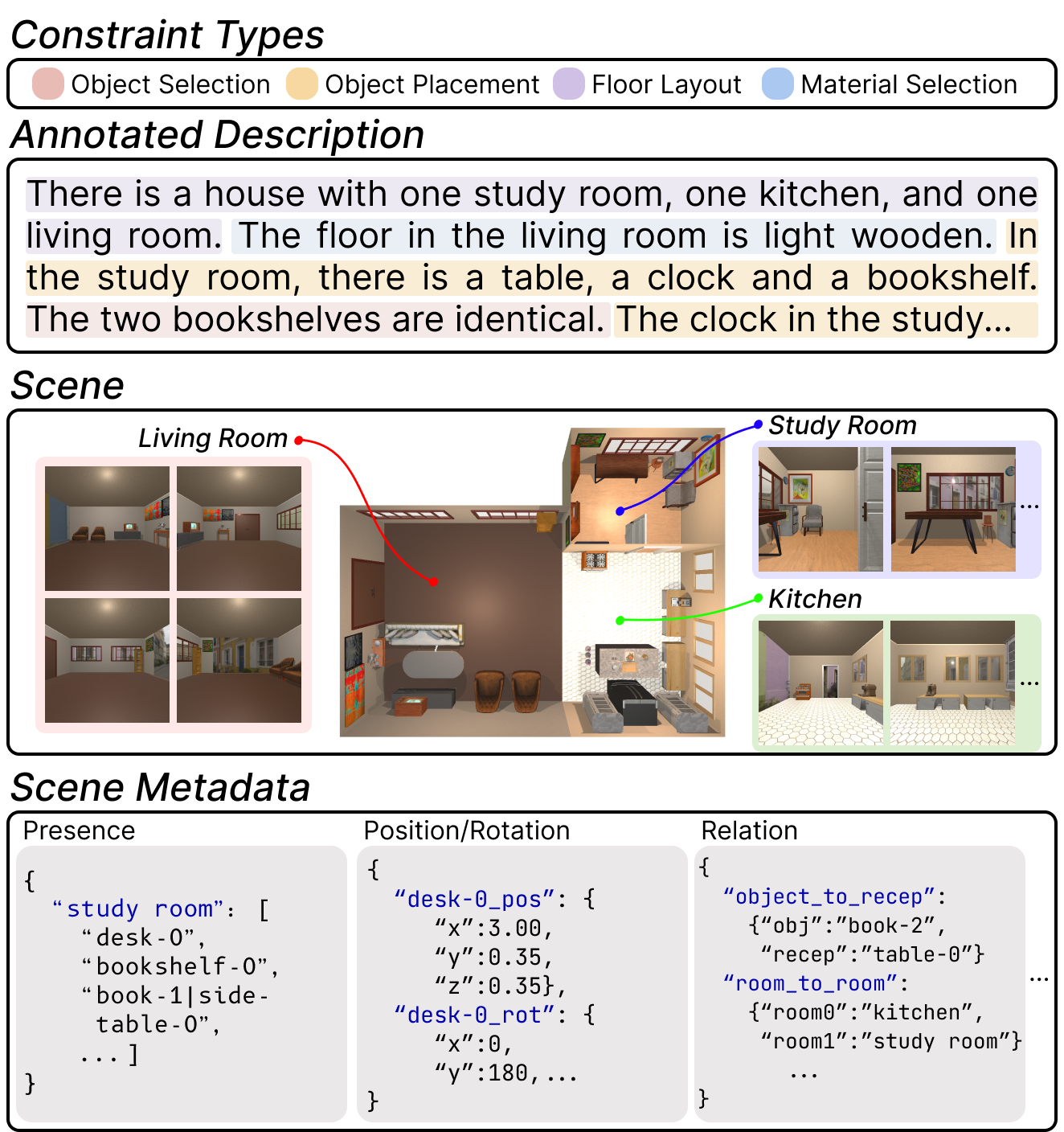}
    \caption{Overview of \bench}
    \label{fig:sample_dataset}
    \vspace{-1.5em}
\end{figure}

\begin{figure}[t]
    \centering
    \includegraphics[width=0.98\columnwidth]{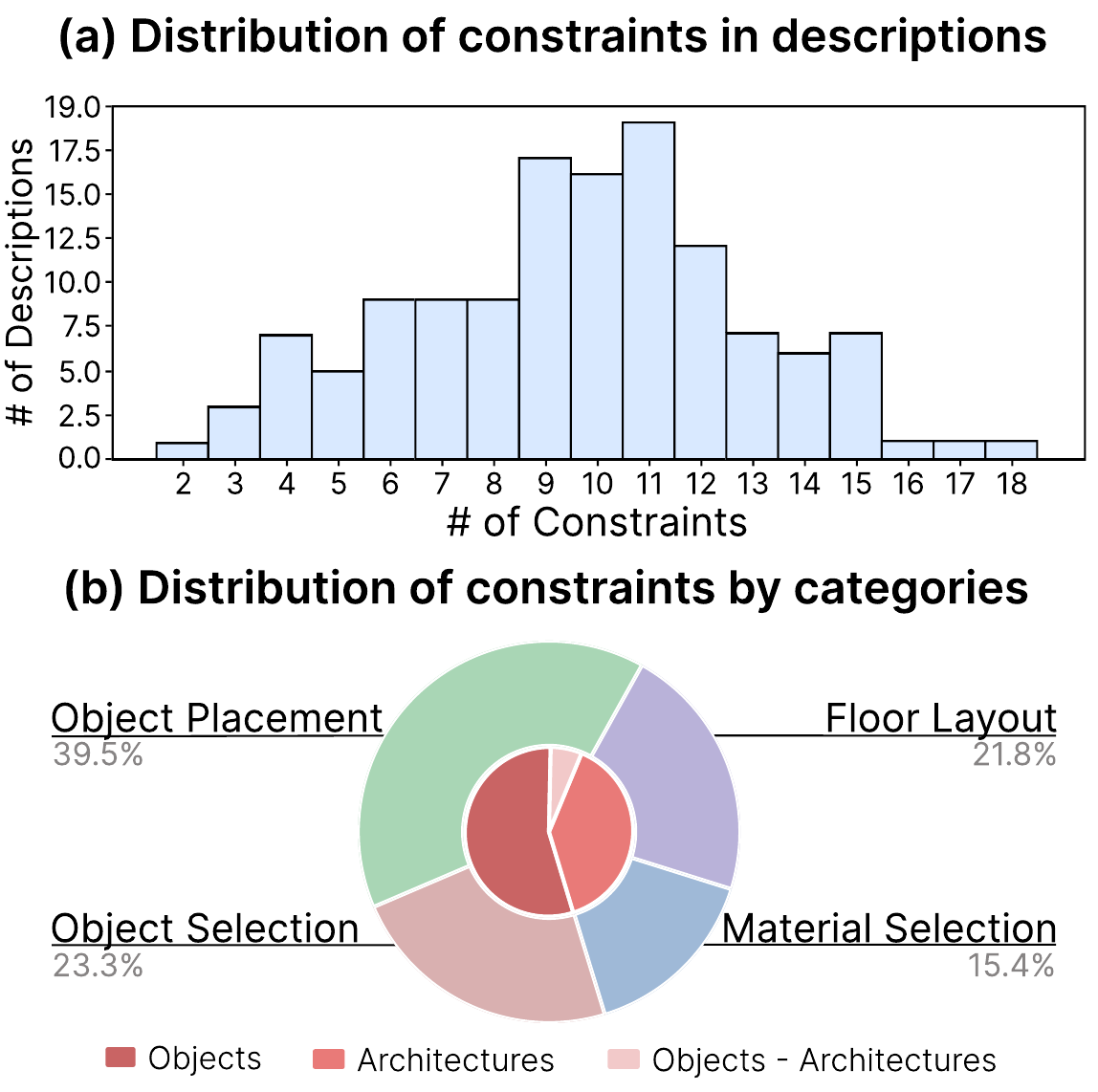}
    \vspace{-0.2em}
    \caption{
    Statistics of \bench.
    }
    \label{fig:BenchStatistics}
    \vspace{-1.5em}
\end{figure}

\bench~is a fully human-curated dataset spanning the entire pipeline from collecting description to scene construction and constraint annotation.
As illustrated in Figure~\ref{fig:sample_dataset}, each instance comprises a fine-grained natural language description with annotated constraints, a manually constructed scene, and rich scene metadata. 
The dataset contains 130 descriptions paired with 1,250 constraints in total, averaging 9.6 per description, with diverse constraint types as shown in Figure~\ref{fig:BenchStatistics}.
Beyond constraints, \bench~also encompasses diverse rooms, receptacles, objects, and structural elements, as detailed in Table~\ref{tab:lego_statistics}, reflecting the complexity and diversity of real-world indoor environments.

\subsection{Dataset Construction} 
\label{ssec:dataset_construction}
Here, we present the construction process of \bench, where four human annotators manually construct descriptions, annotate constraints, and build corresponding scenes for 3D scene synthesis.
Full details are provided in Appendix~\ref{sec:appendix_HumanAnnotation}.

\paragraph{Description Construction and Constraint Annotation.}

First, four annotators create fine-grained descriptions based on their empirical observations and real-world images of indoor spaces.
Subsequently, annotators identify all constraints embedded in each description using a dedicated annotation tool.
Constraints with dependencies are restructured into conditional expressions, and multiple attributes of the same entity are merged into unified statements.
Each constraint is then classified into one of four types---floor layout, material selection, object selection, or object placement---and mapped to its corresponding text segment in the description.

\paragraph{Scene Construction and Verification.}
Using the collected descriptions and constraints, annotators first generate initial scenes using Holodeck~\citep{yang2024holodeck}, which produces scenes as structural JSON representations containing object positions, rotations, and other scene properties.
Object instances are sourced from Objaverse~\citep{NEURIPS2023_70364304}, while structural components—including walls, floors, doors, windows, and materials—are drawn from ProcTHOR~\citep{deitke2022} and AI2-THOR~\citep{kolve2017ai2} asset libraries.
Annotators then manually refine the generated scenes to ensure full alignment with all constraints.
Finally, annotators cross-validate each other's work across two iterations, verifying both annotation accuracy and scene-description alignment, with any discrepancies corrected before completion.

\begin{table}[t]
    \centering
    \scriptsize
    \resizebox{0.96\linewidth}{!}{    
    \begin{tabular}{c|cccc}
    \toprule
        &Room &Receptacle &Object&\makecell{Structural \\ Elements} \\
    \midrule
    \# Total & 295 &2,320 &4,518 &3,974 \\
    \midrule
    \# Types & 161 & 446 &812 &3 \\
    \midrule
    Avg./scene & 2.27 &17.85 &34.75 &30.57 \\
    \bottomrule
    \end{tabular}
    }
    \caption{Statistics for rooms, receptacles, objects, and structural elements in \bench. Room types reflect functional categories rather than layout; receptacle and object types are identity-based. Structural elements include doors, windows, and walls.}
    \label{tab:lego_statistics}
\end{table}


\begin{figure*}[t]
\centering
\includegraphics[width=0.95\textwidth]{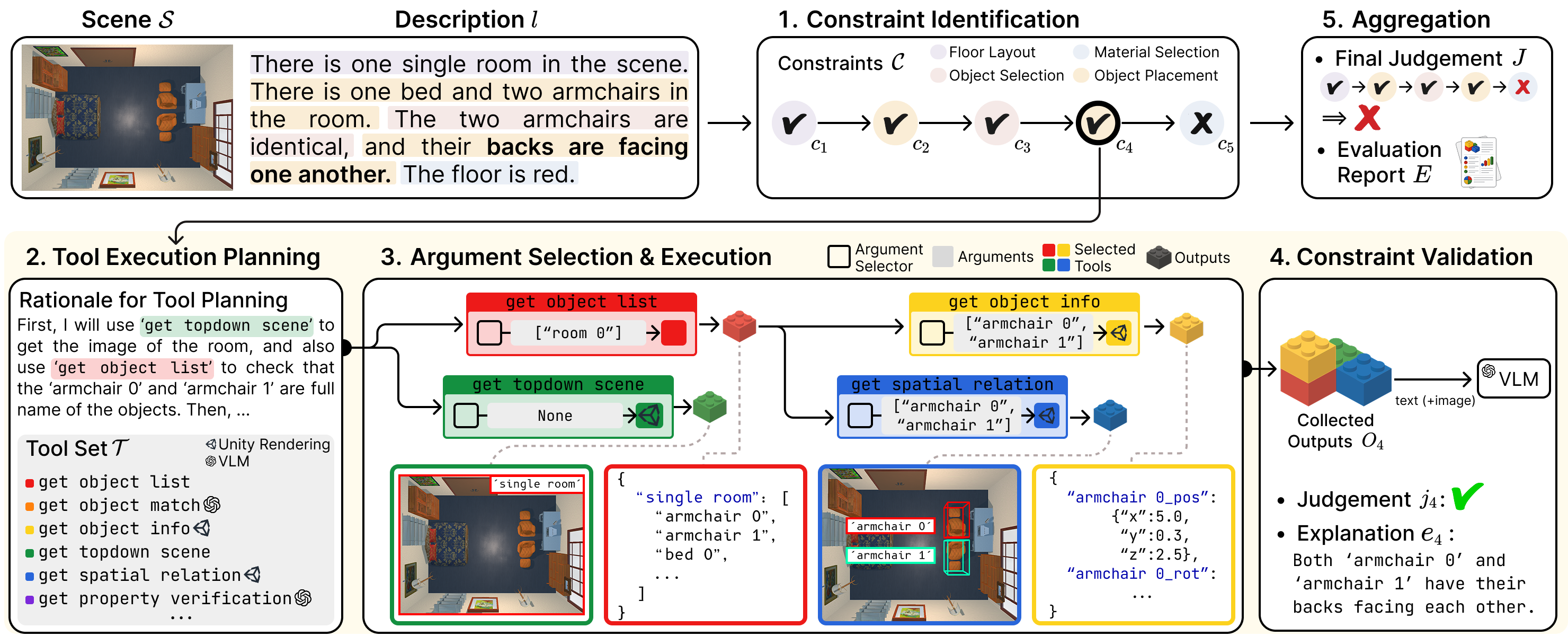}
\caption{\eval. All steps are performed by a VLM judge $\mathcal{M}$, which plans tool execution, selects arguments, executes tools, evaluates each constraint, and aggregates the results into a final judgment $J$ and a report $E$.
}
\vspace{-1em}
\label{fig:lego_eval}
\end{figure*}

\section{\eval: Evaluation with Tool-Augmented VLMs}\label{sec:LEGO-Eval}

We propose \textbf{\eval}, a tool-augmented evaluation framework for fine-grained diagnosis of scene-description alignment in text-guided 3D scene synthesis. 
We first describe the framework in detail, followed by its implementation details.

\subsection{\eval}

\eval~decomposes a description $l$ into individual constraints and evaluates each sequentially in an iterative manner, with each evaluation informed by the results of prior constraints. 
To do so, \eval~leverages tools from tool set $\mathcal{T}$ to retrieve the necessary information from a generated scene $\mathcal{S}$, and employs a VLM judge $\mathcal{M}$ to provide binary judgment $J$ along with evaluation report $E$:

$$J, E \leftarrow \eval(l \mid \mathcal{S})$$

As shown in Figure~\ref{fig:lego_eval}, our evaluation framework consists of the following five steps:

\paragraph{Step 1: Constraint Identification.}
Since a description $l$ is expressed as a natural language description, it may encode multiple atomic requirements, each of which must be satisfied in the scene and evaluated separately.
Therefore, the judge $\mathcal{M}$ first decomposes $l$ into a set of constraints $\mathcal{C} = \{c_i\}^k_{i=1}$.
Since different constraints require different validation strategies (\eg \textit{verifying spatial layout} vs. \textit{checking object appearance}), each constraint $c_i$ is assigned a constraint type $\tau_i$ that determines how it should be evaluated;
$\tau_i$ belongs to one of four types\footnote{Inspired by Holodeck~\citep{yang2024holodeck}.}:

\begin{itemize}[left=0pt,labelsep=0.5em,itemsep=3pt,topsep=0.5em,parsep=0pt]
    \item \textbf{Floor Layout}: Spatial layout of rooms, walls, doors, and windows.
    \item \textbf{Material Selection}: Visual appearance of floors and walls.
    \item \textbf{Object Selection}: Appearance of objects, including doors and windows.
    \item \textbf{Object Placement}: Object placement and rotation within the scene.
\end{itemize}

\paragraph{Step 2: Tool Execution Planning.}
\label{par:step2}
The set of constraints $\mathcal{C} = \{c_i\}^k_{i=1}$ are evaluated sequentially.
For each constraint $c_i$, evaluating a single constraint often requires multi-hop reasoning—for instance, verifying ``\textit{the cup is on the red table}'' first requires confirming the existence and color of the table before assessing the cup's placement.
To capture such dependencies, the judge $\mathcal{M}$ generates a graph-structured tool execution plan, selecting a subset of tools $\mathcal{T}_i \subseteq \mathcal{T}$ based on the assigned constraint type $\tau_i$.
Furthermore, if the information needed to evaluate the constraint $c_i$ has already been retrieved while processing prior constraints $\{c_\ell\}_{\ell<i}$, the model reuses prior results to avoid redundant tool executions.

\paragraph{Step 3: Tool Argument Selection \& Execution.}
While Step 2 determines the logical order of tool execution, accurate validation further requires precise grounding of the relevant objects within the scene $\mathcal{S}$.
For this, for each tool $t \in \mathcal{T}$ in the execution plan, the judge $\mathcal{M}$ selects tool arguments $\alpha$ by conditioning on the current constraint $c_i$ of type $\tau_i$, along with all prior judgements $\{j_\ell\}_{\ell<i}$ and explanations $\{e_\ell\}_{\ell<i}$.
The resulting outputs are collected as $O_i=\{{t(\alpha; \mathcal{S})}\}_{t \in \mathcal{T}_i}$ for constraint validation.

\begin{table*}[ht]
\centering
\footnotesize
\setlength{\tabcolsep}{3.5pt}
\resizebox{0.85\textwidth}{!}{
\begin{tabular}{>{\raggedright\arraybackslash}m{3cm}*{4}{>{\centering\arraybackslash}m{1.4cm}}|*{4}{>{\centering\arraybackslash}m{1.4cm}}}
\toprule
\multicolumn{1}{c}{\multirow{2}{*}{\textbf{Methods}}}
& \multicolumn{4}{c|}{\textbf{Holistic} $\uparrow$}
& \multicolumn{4}{c}{\textbf{Partial} $\uparrow$} \\
\cmidrule(lr){2-5} \cmidrule(lr){6-9}
& \textbf{F1} & \textbf{Precision} & \textbf{Recall} & \textbf{Cohen's $\kappa$}
& \textbf{F1} & \textbf{Precision} & \textbf{Recall} & \textbf{Cohen's $\kappa$} \\
\midrule

\textbf{CLIPScore}
& 0.49 & 0.51 & 0.51 & 0.02
& 0.46 & 0.50 & 0.50 & 0.00 \\

\midrule
\multicolumn{9}{l}{\textbf{VLM-as-a-judge}} \\

\quad Gemini 2.5 Pro
& 0.38 & 0.76 & 0.52 & 0.05
& 0.60 & 0.66 & 0.70 & 0.28 \\

\quad GPT-4.1
& 0.40 & 0.67 & 0.53 & 0.05
& 0.68 & 0.65 & 0.73 & 0.35 \\

\quad GPT-5.1
& 0.38 & 0.68 & 0.52 & 0.04
& 0.60 & 0.71 & 0.71 & 0.28 \\

\midrule
\multicolumn{9}{l}{\textbf{SceneEval\textsuperscript{*}}} \\

\quad GPT-4o
& 0.47 & 0.74 & 0.58 & 0.15
& 0.45 & 0.64 & 0.58 & 0.12 \\

\midrule
\multicolumn{9}{l}{\textbf{\eval}} \\

\quad Qwen3-VL-8B
& 0.67 & 0.74 & 0.69 & 0.38
& 0.71 & 0.74 & 0.69 & 0.42 \\

\quad Qwen3-VL-32B
& 0.67 & 0.79 & 0.70 & 0.39
& 0.80 & 0.80 & \textbf{0.81} & 0.61 \\

\quad GPT-4.1-mini
& 0.70 & 0.78 & 0.72 & 0.43
& 0.78 & 0.80 & 0.76 & 0.56 \\

\quad GPT-4.1
& \textbf{0.81} & \textbf{0.84} & \textbf{0.82} & \textbf{0.63}
& \textbf{0.83} & \textbf{0.86} & \textbf{0.81} & \textbf{0.66} \\

\bottomrule
\end{tabular}
}
\caption{Human comparison of \eval~and existing evaluation methods.}
\vspace{-1em}
\label{tab:eval_method_comparison}
\end{table*}

\paragraph{Step 4: Constraint Validation.}
Given the collected tool outputs $O_i$, the judge $\mathcal{M}$ produces a binary judgement $j_i$ and explanation $e_i$ assessing whether the constraint $c_i$ is satisfied in the scene $\mathcal{S}$.
Steps 2–4 are repeated for each constraint $c_i \in \mathcal{C}$, accumulating per-constraint evaluation results $\{(j_i, e_i)\}_{i=1}^{k}$ for aggregation.
\paragraph{Step 5: Aggregation.}

After all constraints are evaluated via Steps 2-4, the judge $\mathcal{M}$ aggregates per-constraint evaluation results $\{(j_i, e_i)\}_{i=1}^{k}$ into a final judgment $J$ and a comprehensive evaluation report $E$.
Final judgement $J$ is valid only if all binary judgements $\{j_i\}_{i=1}^k$ are valid, ensuring that no fine-grained requirement is overlooked.
Furthermore, the report $E$ provides comprehensive per-constraint explanations that enable detailed analysis of scene deviations from the description $l$.

\subsection{Implementation Details}
\label{ssec:toolset}
\paragraph{Environment.}
\eval~is built upon the AI2THOR simulator~\citep{kolve2017ai2}, a Unity-based 3D simulator, enabling direct access to object attributes and spatial configurations via the rendering engine.

\paragraph{Tool set.}
\eval~is augmented with 21 diverse tools, grouped into three types: 
\begin{itemize}[left=0pt,labelsep=0.5em,itemsep=3pt,topsep=0.5em,parsep=0pt] 

\item \textbf{Environment Interaction}: Visual information retrieval via direct Unity interface (\eg \texttt{get topdown scene}, \texttt{get spatial relation}). 

\item \textbf{Textual Reasoning}: Attribute retrieval from structured textual scene representation (\eg \texttt{get room list}, \texttt{get object info}). 

\item \textbf{Multimodal Reasoning}: Visual-to-text conversion using VLMs~\citep{wang2024muirbench} for reliable image-based reasoning (\eg \texttt{get object match}, \texttt{get property description}). 

\end{itemize}
More details on the environment and tools are in Appendix~\ref{appendix:env_setting} and ~\ref{sec:Tool Set}.

\section{Experiments}
\label{sec:Experiments}
To validate \eval, we evaluate its agreement with human judgment for different automatic evaluation methods using \bench. Furthermore, leveraging both \eval~and \bench, we benchmark existing LLM-based 3D scene synthesis methods, revealing their limitations.

\subsection{Benchmarking Evaluation Methods}\label{sec:comparison_of_eval_methods}
\subsubsection{Setup}

\paragraph{Evaluation Dataset.}
We use the descriptions, annotated constraints, and scenes from \bench~to compare the performance of evaluation methods. 
To enrich the dataset, we additionally construct a misaligned pair for each of the 130 original descriptions by manually crafting a scene that violates the description, resulting in 260 pairs in total.


\begin{figure*}[t]
\centering
\includegraphics[width=0.93\textwidth]{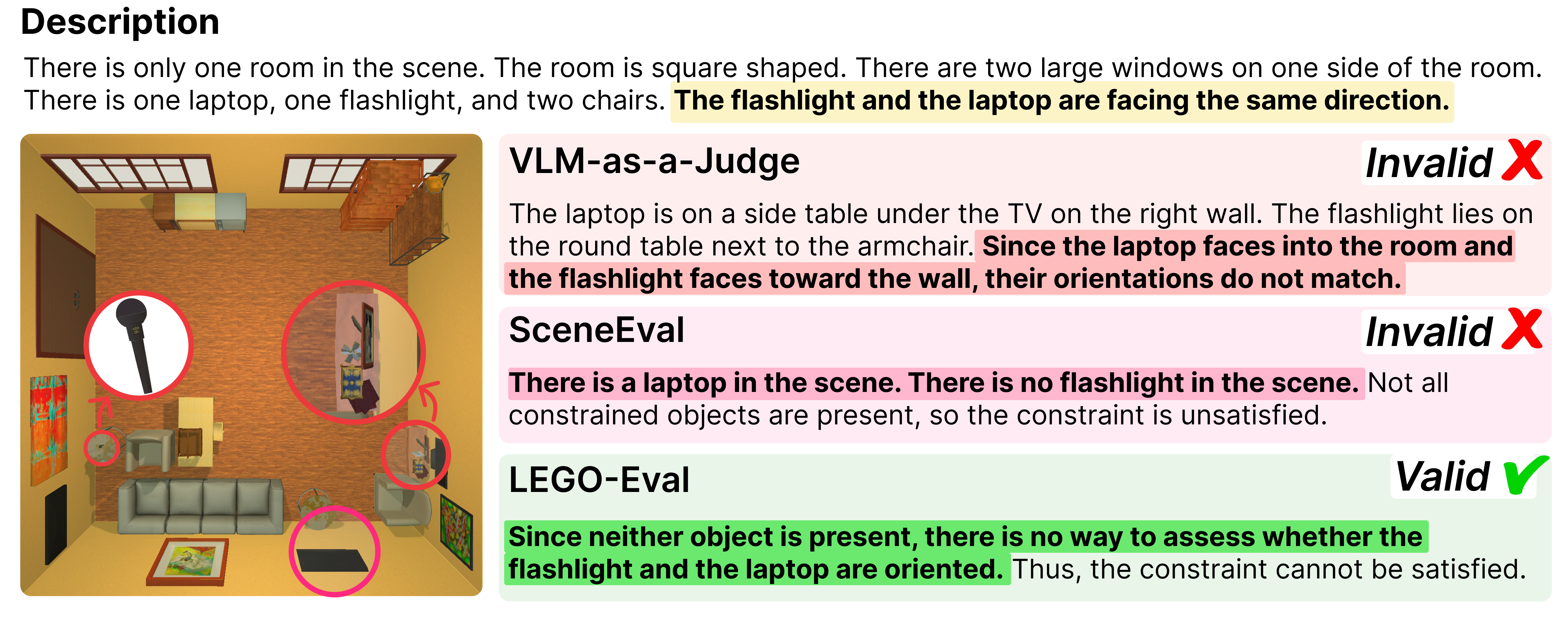}
\vspace{-0.2em}
\caption{Comparison of evaluation results from VLM-as-a-judge, SceneEval, and \eval.
\vspace{-1em}
}
\label{fig:case_study}
\end{figure*}

\paragraph{Baselines.}
We compare \eval~against CLIPScore~\citep{hessel2021clipscore}, VLM-as-a-judge, and SceneEval~\citep{tam2025sceneeval}. 
CLIPScore computes the similarity between the description and the top-down scene image, and judges the scene as aligned if the score exceeds a threshold of 20. 
For VLM-as-a-judge, we provide scene images from four perspectives and use Gemini-2.5-Pro, GPT-4.1, and GPT-5.1 as base models with self-consistency across 3 samples. 
For SceneEval, we leverage GPT-4o as the backbone VLM\footnote{SceneEval cannot assess 41\% of the constraints in \bench~due to its fixed evaluation criteria; we evaluate it only on the constraints it can handle for a fair comparison.}. 
Following \eval, all baseline methods consider a scene to be aligned with the description if all constraints are satisfied.


\paragraph{Metrics.}
We evaluate the methods using F1, precision, recall, and Cohen's $\kappa$. Cohen's $\kappa$ measures agreement with human judgments beyond chance, a chance-corrected complement to F1.
We compute these metrics at two levels: (1) Holistic, which measures agreement with human judgments on the full description, and (2) Partial, which measures agreement on each constraint.
More details on Cohen's $\kappa$ are presented in Appendix~\ref{appendix:metric_details}


\subsubsection{Results}
As shown in Table~\ref{tab:eval_method_comparison}, existing methods all suffer from distinct limitations. 
CLIPScore shows unreliable results for evaluating fine-grained alignment with 3D scenes.
This may stem from the as CLIP~\citep{radford2021learning} being trained on 2D image-text pairs, which limits its ability to capture 3D scene structure.
VLM-as-a-judge fails to identify and ground the scene components referenced in the description, leading to incorrect evaluation.
SceneEval does not overcome the grounding limitation of VLM-as-a-judge, and its evaluation suffers from restricted scope—it applies rigid rule-based criteria for spatial relations such as 'beside' and only handles object-level relations.


In contrast, \eval~significantly outperforms all baselines, achieving the highest F1 and Cohen's $\kappa$ at both the holistic and partial levels. 
Notably, even with a lightweight backbone such as Qwen3-VL-8B, \eval~still outperforms all baselines.
This advantage stems from \eval's tool-augmented design, which integrates diverse tools for multi-hop grounding to locate and interpret scene components more reliably than the baselines. 
These results demonstrate that \eval~provides fine-grained and reliable evaluation of text-guided 3D scene synthesis.



\paragraph{Case Study.}
Figure~\ref{fig:case_study} illustrates an example comparing evaluation results from VLM-as-a-judge, SceneEval, and \eval, revealing that while all methods achieve accurate judgments, their reasoning processes differ significantly. Although the flashlight and laptop do not exist in the scene, the VLM-as-a-judge locates them and determines they are not facing the same direction. Similarly, SceneEval misidentifies the black painting on the wall as a laptop. In contrast, \eval~accurately recognizes the absence of both objects and determines the constraint cannot be satisfied.

\begin{table}[t]
\centering
\small
\setlength{\tabcolsep}{4pt}
\resizebox{0.85\columnwidth}{!}{%
\begin{tabular}{lcc}
\toprule
\textbf{Tools} & \textbf{Holistic F1 ($\Delta$)} & \textbf{Partial F1 ($\Delta$)} \\
\midrule
w/o \textbf{M}\,\icon{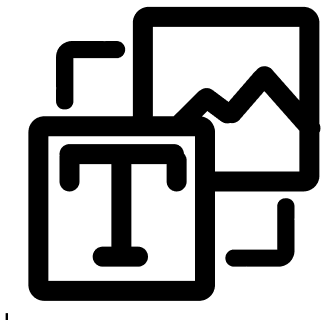}
& $-0.04\%$ & $-1.02\%$ \\
w/o \textbf{T}\,\icon{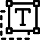}
& $-5.05\%$ & $-2.65\%$ \\
w/o \textbf{T}\,\icon{text.png} + \textbf{M}\,\icon{mmr.png}
& $-6.46\%$ & $-2.81\%$ \\
w/o \textbf{E}\,\icon{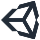} + \textbf{M}\,\icon{mmr.png}
& $-24.90\%$ & $-5.34\%$ \\
\bottomrule
\end{tabular}%
}
\caption{Performance drop with tool types disabled.
(\textbf{M}: Multimodal Reasoning,
\textbf{T}: Textual Reasoning,
\textbf{E}: Environment Interaction)}
\label{tab:ToolTypeAblation}
\end{table}
\begin{table*}[t]
\centering
\resizebox{0.9\linewidth}{!}{
\begin{tabular}{>{\raggedright\arraybackslash}m{2.6cm}|c|ccccc}
\toprule
\multicolumn{1}{c|}{\multirow{2}{*}{\textbf{Method}}} & \multirow{2}{*}{\textbf{Holistic SR $\uparrow$}} & \multicolumn{5}{c}{\textbf{Partial SR $\uparrow$}} \\
\cmidrule(lr){3-7}
& & \textbf{Floor Layout} & \textbf{Material Selection} & \textbf{Object Selection} & \textbf{Object Placement} & \textbf{Avg.} \\
\midrule
I-Design   & 3.8  & 92.7 & 63.7 & 11.0 & 4.1  & 34.2 \\
MANSION  & 5.4 & 67.7 & 53.9 & \textbf{55.7} & \textbf{56.1} & 58.2 \\
LayoutGPT  & 6.9  & 96.0 & \textbf{65.3} & 40.9 & 37.3 & 55.2 \\
Holodeck   & 8.4  & \textbf{96.3} & 61.6 & 46.3 & 43.7 & 58.5 \\
LayoutVLM  & \textbf{10.0} & 95.6 & \textbf{65.3} & 49.8 & 46.0 & \textbf{60.6} \\

\bottomrule
\end{tabular}
}
\caption{Evaluation results of LLM-based 3D scene synthesis methods on \bench.}
\label{tab:scene_generation_methods}
\end{table*}

\subsubsection{Ablation Study}\label{sec:ablation_studies}

We conduct an ablation study to verify that each tool type in our toolset contributes substantially to evaluation performance.
Specifically, we disable each tool type individually and measure the resulting performance of \eval, while keeping tools that return lists of scene components enabled, as these are necessary for argument selection.
As shown in Table~\ref{tab:ToolTypeAblation}, disabling any single tool type degrades performance, confirming that all are indispensable for comprehensive evaluation.
The largest drop occurs when textual tools are disabled, underscoring that visual inputs alone—on which conventional evaluation methods often rely—are insufficient.
Textual tools capture subtle scene attributes, such as colors of small objects that are hard to identify from images alone. 
They also convey key information about scene components extracted from retrieved images.


\subsection{Evaluation via \eval}
\label{sec:comparison_of_generation_methods}

\begin{figure}[t]
    \centering
    \includegraphics[width=0.85\columnwidth]{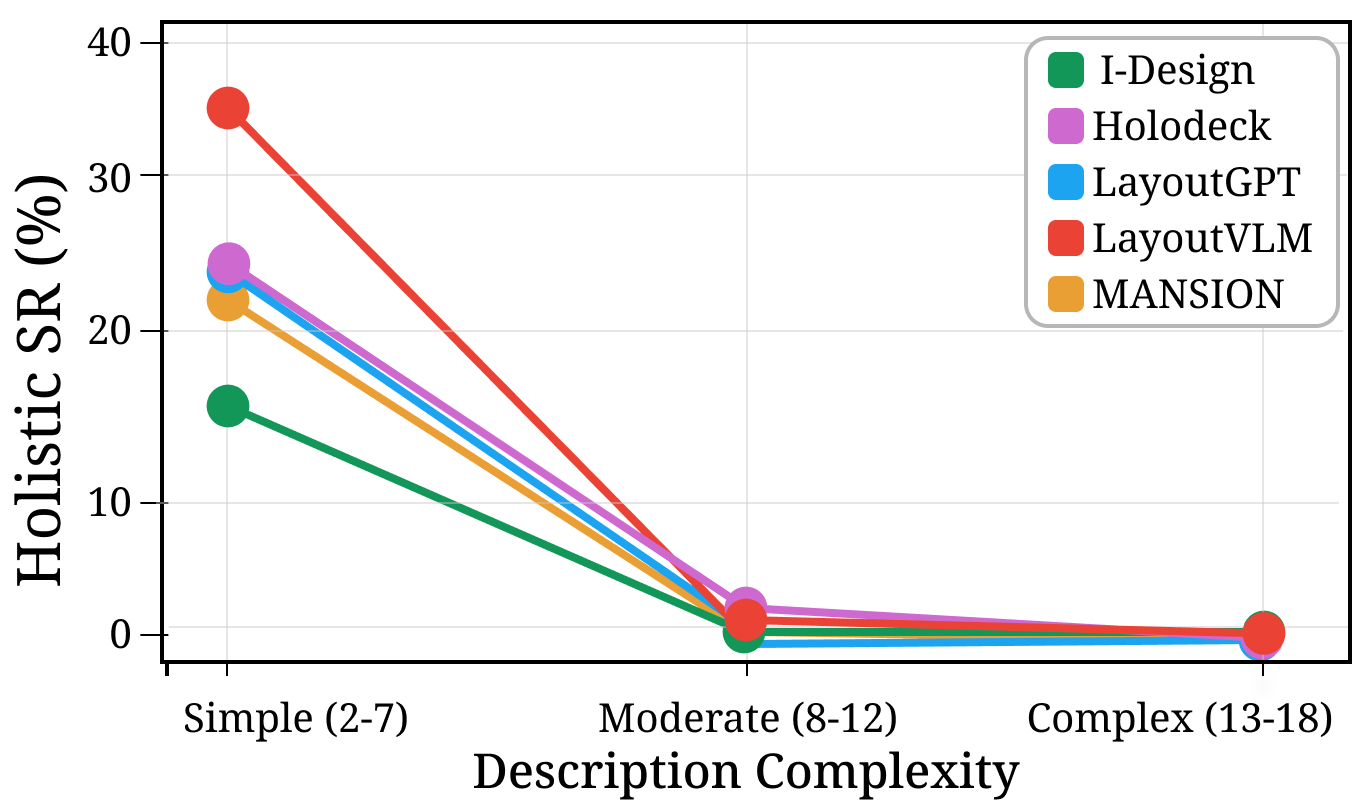}
    \caption{Description complexity vs. Holistic SR.}
    \label{fig:constraint_count}
\end{figure}

\subsubsection{Setup}
\paragraph{Baselines.} 
We evaluate five LLM-based text-guided 3D scene synthesis methods on \bench~with \eval: LayoutGPT~\citep{feng2023layoutgpt}, Holodeck~\citep{yang2024holodeck}, I-Design~\citep{ccelen2025design}, LayoutVLM~\citep{sun2025layoutvlm} and MANSION~\citep{che2026mansion}. These methods vary in functionality: Holodeck and MANSION generate complete scenes with object selection, attributes, and placement; I-Design only selects and places objects but omits attributes; LayoutGPT and LayoutVLM only position a given set of objects. 
For fair comparison, we augment the latter three with Holodeck to produce full scenes for evaluation. 
Details on the methods are in Appendix~\ref{appendix:Scene Generation Methods}.

\paragraph{Metrics.} We measure performance with two Success Rates (SR) metrics: (1) Holistic SR, the proportion of descriptions with all specified constraints satisfied;
(2) Partial SR, the proportion of valid constraints. 
For partial SR, we report success rates by constraint type and their overall average.

\subsubsection{Results}
\label{ssec:comparison_of_generation_methods_results}
As shown in Table~\ref{tab:scene_generation_methods}, no current method achieves a holistic SR above 10\%, meaning that nearly no synthesized scene satisfies the full description.
While most methods reach average partial SRs exceeding 50\%, their performance on object selection and placement falls below this level. 
This shows that current methods still struggle with two fundamental aspects of scene synthesis—choosing the right objects and placing them appropriately. 

\paragraph{Existing methods fail on complex descriptions.}

Real-world descriptions are inherently complex, as evidenced by our fully human-curated \bench, where user-written room descriptions (Appendix~\ref{appendix:Survey}) contain an average of 18.2 constraints per room.
To analyze how this complexity affects performance, we categorize descriptions based on the number of constraints into three groups: simple (2–7 constraints), moderate (8–12), and complex (more than 12). 
As Figure~\ref{fig:constraint_count} shows, existing methods consistently fail on complex descriptions—precisely what real-world use demands. 
This means that scenes synthesized by current methods are likely to deviate from the real-world training distribution, undermining their usefulness as training environments.

\subsubsection{Human Evaluation}
\label{ssec:humanl_eval}
To further validate the scene synthesis results in Section~\ref{ssec:comparison_of_generation_methods_results}, we conduct a human evaluation of scene-description alignment on generated scenes.
We randomly sample 20 scenes from each of the five scene synthesis methods used in Section~\ref{sec:comparison_of_generation_methods} and ask three undergraduate evaluators to independently verify all associated constraints. 
The evaluators are given the same scene information available to \eval, consisting of scene images and scene graph information. 
The final label for each constraint is determined by majority voting among the human evaluators~\footnote{Inter-evaluator agreement is 80.3\%.}.

\paragraph{\eval~shows strong agreement with human judgments.}
As shown in Table~\ref{tab:human_eval}, \eval~shows strong agreement with human judgments (72.6-89.8\% across methods) and yields similar relative trends across synthesis methods.
Moreover, the human-judged scores remain comparably low, indicating that the low absolute scores in Section~\ref{sec:comparison_of_eval_methods} are not because \eval~is overly strict.
It rather reflects the difficulty of satisfying the multiple constraints specified for each scene, underscoring the importance of \eval~in enabling reliable progress on this challenging task.

\begin{table}[t]
\centering
\small
\setlength{\tabcolsep}{5pt}
\renewcommand{\arraystretch}{1.08}

\resizebox{0.98\linewidth}{!}{
\begin{tabular}{llccc}
\toprule
\textbf{Method}
& \textbf{Evaluator}
& \textbf{Hol. SR}
& \textbf{Part. SR}
& \textbf{Agreement} \\
\midrule

\multirow{2}{*}{iDesign}
& Human
& 5.0\%
& 29.3\%
& \multirow{2}{*}{89.8\%} \\
& \eval
& 5.0\%
& 31.4\%
& \\
\midrule

\multirow{2}{*}{LayoutVLM}
& Human
& 15.0\%
& 56.9\%
& \multirow{2}{*}{77.7\%} \\
& \eval
& 20.0\%
& 61.7\%
& \\
\midrule

\multirow{2}{*}{LayoutGPT}
& Human
& 5.0\%
& 54.8\%
& \multirow{2}{*}{72.6\%} \\
& \eval
& 10.0\%
& 55.9\%
& \\
\midrule

\multirow{2}{*}{Holodeck}
& Human
& 10.0\%
& 55.9\%
& \multirow{2}{*}{82.1\%} \\
& \eval
& 10.0\%
& 57.4\%
& \\
\midrule

\multirow{2}{*}{Mansion}
& Human
& 0.0\%
& 52.7\%
& \multirow{2}{*}{76.5\%} \\
& \eval
& 5.0\%
& 59.6\%
& \\

\bottomrule
\end{tabular}
}
\caption{
Human Evaluation Results.
}
\label{tab:human_eval}
\end{table}

\section{Exploring the Reliability and Utility of \eval}
\label{sec:FurtherAnalyses}


In this section, we take a closer look at \eval, examining \textit{why} it reliably works and \textit{whether} its usefulness extends beyond diagnosis.
We first identify the source of its reliability (\S\ref{ssec:legoeval_reliable}) and assess its utility as a feedback signal for refinement (\S\ref{ssec:feedback})

\subsection{Examining the Source of \eval’s Reliability}
\label{ssec:legoeval_reliable}

To understand why \eval~reliably works, we examine whether its advantage merely stems from privileged access to scene information.
\eval~can retrieve relevant information for each constraint directly through its tools, whereas other baselines must infer all of this from rendered images alone.
We therefore narrow this gap from two directions: more access to scene information (\S\ref{ssec:scene_info}), and better access to scene information through identification modules (\S\ref{ssec:failure_mode}).

\subsubsection{Does More Access to Scene Information Improve Evaluation?}
\label{ssec:scene_info}


We introduce two variants with \textit{full access to scene information}: one that receives the complete text scene graph alone, and one that receives both the scene graph and images. 
We use VLM-as-a-judge for our baseline, with GPT-4.1 as the base model for all variants and \eval. 
Note that we also include SceneEval for comparison, since it also has access to a structured scene representation.


\paragraph{Even full access to scene information does not lead to better evaluation.}
As shown in Table~\ref{tab:vlm_info}, providing VLM-as-a-judge with the full scene graph fails to improve its evaluation capability.
SceneEval, despite its access to a structured scene representation, likewise remains far below \eval.
These results suggest that simply giving more scene information is insufficient for reliable evaluation. 
Rather than simply relying on more scene information, \eval~determines which information each constraint requires and selectively invokes the appropriate tools to retrieve it. 
Thus, the result highlights that the ability to identify and leverage constraint-relevant scene information matters for reliable evaluation.


\subsubsection{Does Better Access to Scene Information Improve Evaluation?}
\label{ssec:failure_mode}
As identifying constraint-relevant scene information is a prerequisite for leveraging it, we next ask whether better access at the identification stage improves evaluation.
We augment VLM-as-a-judge with three identification modules: bounding box detection (Grounding DINO~\citep{liu2023grounding}), scene graph generation (LASER~\citep{huang2025laser}), and 3D language models (Point-LLM~\citep{guo2023pointbindpointllmaligning}).
To locate where evaluation still fails, we manually annotate each incorrect judgment into one of three error types--Identification, Spatial, and Attribute.
These correspond to failures in locating or identifying scene elements, inferring spatial relationships between them, and inferring their properties such as color, size, or shape, respectively.
Details of the modules and the experimental setup are in Appendix~\ref{appendix:vlm_modular}.



\begin{table}[t]
\centering
\small
\setlength{\tabcolsep}{4pt}
\renewcommand{\arraystretch}{1.08}

\resizebox{0.98\linewidth}{!}{
\begin{tabular}{lcccc}
\toprule
\multirow{2}{*}{\textbf{Method}} 
& \multicolumn{2}{c}{\textbf{Holistic} $\uparrow$} 
& \multicolumn{2}{c}{\textbf{Partial} $\uparrow$} \\
\cmidrule(lr){2-3} \cmidrule(lr){4-5}
& F1 & Cohen's $\kappa$ & F1 & Cohen's $\kappa$ \\
\midrule

VLM-as-a-judge (Images-only)
& 0.40 & 0.05 & 0.68 & 0.35 \\

VLM-as-a-judge (SG-only) 
& 0.37 & 0.04 & 0.66 & 0.36 \\

VLM-as-a-judge (Images + SG) 
& 0.37 & 0.04 & 0.65 & 0.35 \\

SceneEval 
& 0.47 & 0.15 & 0.45 & 0.12 \\

\midrule
\textbf{\eval} 
& \textbf{0.81} & \textbf{0.63} & \textbf{0.83} & \textbf{0.66} \\

\bottomrule
\end{tabular}
}

\caption{Effect of more access to scene information on evaluation reliability. SG denotes the scene graph in text form.}
\label{tab:vlm_info}
\end{table}

\paragraph{Identification support alone worsens evaluation.}
Surprisingly, all variants increase total errors rather than reducing them. 
As shown in Table~\ref{tab:error_comparison}, Identification errors in particular worsen substantially despite being the very type each module was designed to address.
One likely reason is that the judge cannot assess whether the provided information is relevant to the constraint being verified.
\eval~instead decides which information each constraint requires before retrieving it. 
These results suggest that the gap does not come from better access to identification results, but from how \eval~decides which information each constraint requires before retrieving it.
\begin{table}[t]
\centering
\footnotesize
\setlength{\tabcolsep}{4pt}

\resizebox{0.85\columnwidth}{!}{
\begin{tabular}{>{\raggedright\arraybackslash}m{3cm}cccc}
\toprule

\textbf{Method}
& \textbf{Total} 
& \textbf{Ident.} 
& \textbf{Spa.} 
& \textbf{Attr.}  \\

\midrule

VLM-as-a-judge
& 376
& 150
& 76
& 150 \\

\quad + \textbf{Grounding DINO}
& 451
& 288
& 42
& 121 \\

\quad + \textbf{LASER}
& 497
& 233
& 100
& 164 \\


\quad + \textbf{Point-LLM}
& 447
& 203
& 72
& 172 \\

\midrule

LEGO-Eval
& \textbf{119}
& \textbf{53}
& \textbf{20}
& \textbf{46} \\

\bottomrule
\end{tabular}
}

\caption{Number of errors across different evaluation methods, all based on GPT-4.1.}
\label{tab:error_comparison}

\end{table}

\subsection{Applying \eval~to Improve Scene Synthesis}
\label{ssec:feedback}

Beyond evaluation, fine-grained diagnosis can directly support improvement when fed back to the generator.
To examine this, we use \eval's rationale as feedback to refine the results of Holodeck (Section~\ref{sec:comparison_of_generation_methods}) over 3 refinement iterations, with GPT-4.1 as the baseline VLM judge.

\begin{figure}[t]
    \centering
    \includegraphics[width=0.90\columnwidth]{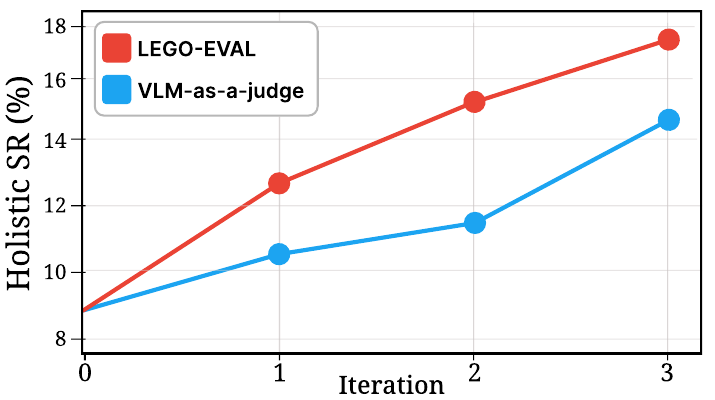}
    \caption{Comparison of refinement results using VLM-as-a-judge and \eval~as feedback signal.}
    \label{fig:figure_refinement}
\end{figure}

\paragraph{Fine-grained diagnosis improves text-guided 3D scene synthesis.}
As shown in Figure~\ref{fig:figure_refinement}, \eval's feedback leads to greater improvements in Holodeck than VLM-as-a-judge, demonstrating its higher utility as a refinement signal.
Since feedback quality ultimately depends on accurate grounding, even a stronger base model cannot compensate when grounding itself is unreliable.
In contrast, \eval~grounds each constraint via dedicated tools, producing detailed and interpretable feedback. 
Together, these establish \eval~as a reliable signal for both evaluating and improving text-guided 3D scene synthesis.

\section{Related Work}
\paragraph{Text-guided 3D scene synthesis.}
Recently, LLMs and VLMs have been adopted for 3D scene synthesis, leveraging their real-world priors.
Prior works generally fall into three categories: approaches that generates complete 3D environments from descriptions~\citep{feng2023layoutgpt, yang2024holodeck, bucher2025respace}, approaches that focus on selecting relevant objects and determining their placements~\citep{yang2024llplace, ccelen2025design}, and layout optimization methods that refine the spatial arrangement of predefined assets~\citep{sun2025layoutvlm, ran2025direct}.

\paragraph{Automatic evaluation for 3D scene synthesis.}
Prior work evaluates text-guided 3D scene synthesis with CLIPScore between top-down scene images and descriptions~\citep{ocal2024sceneteller, fu2024anyhome, deng2025global}, or by prompting VLMs with multi-view images~\citep{wang2024architect, ccelen2025design, sun2025layoutvlm, ling2025scenethesis}. 
However, these methods fail to reliably assess fine-grained scene fidelity, such as object counts, attributes, and spatial relations~\citep{ma2023crepe, li2024topviewrs}. 
Recently, SceneEval~\citep{tam2025sceneeval} predefines evaluation criteria but cannot evaluate attributes or spatial relations involving architectural elements (\eg ``\textit{A sliding window is on the orange wall}''), and restricts spatial relations to basic ones like ``\textit{left of}'' or ``\textit{near}'', excluding comparative expressions such as ``\textit{The table is closer to the chair than the bed}''. 
In contrast, \eval~supports spatial reasoning across all scene components and handles diverse natural-language relationships by leveraging various tools.

\section{Conclusion} 
We present \bench, a human-curated dataset for evaluating text-guided 3D scene synthesis, and \eval, a tool-augmented framework that evaluates scene-description misalignment at the constraint level.
We show that \eval~substantially outperforms existing evaluators, and further reveal that current scene synthesis methods achieve holistic success rates below 10\%—highlighting the gap between current capabilities and real-world demands.
We hope \bench~and \eval~together drive progress toward more faithful text-guided 3D scene synthesis, providing a foundation of diverse and reliable 3D environments for training more capable embodied agents.

\section*{Limitations}
Despite the contributions of our work, several limitations remain.  
First, \bench~currently focuses on scenes composed of rectangular or square rooms. 
This limits the benchmark's coverage of more diverse spatial layouts and scene descriptions.
Extending the benchmark to handle richer scene geometries remains an important direction for future work.
Second, the environment interaction tools and multimodal reasoning tools used in \eval~are simulator-dependent. 
Consequently, when the simulator is unavailable, evaluation must rely only on textual tools. 
Extending \eval~to support evaluation without simulator-specific tools remains an important direction for future work.
Finally, evaluating the \bench~benchmark with \eval~requires approximately two hours, limiting evaluation efficiency.
Improving efficiency through faster model inference and more efficient tool execution remains an important direction for future work.

\section*{Acknowledgements}

This work was supported by Institute for Information \& Communications Technology Planning \& Evaluation (IITP) grant funded by the Korea government (MSIT) (No. RS-2026-25585074, Development of Agentic AI Technology for Enterprise-wide Business Innovation Based on Goal-Oriented Long-term Context Reasoning, 50\%). This work was supported by Institute of Information \& communications Technology Planning \& Evaluation (IITP) grant funded by the Korea government(MSIT) (No. RS-2024-00457882, National AI Research Lab Project, 30\%). This research was supported by the MSIT(Ministry of Science, ICT), Korea, under the Global Research Support Program in the Digital Field program(RS-2024-00436680) supervised by the IITP(Institute for Information \& Communications Technology Planning \& Evaluation, 20\%). This project is partially supported by Microsoft Research Asia.

\bibliography{custom}
\appendix

\section{\bench}

\subsection{Data Annotation Tool}
\begin{figure*}[t]
\centering
\includegraphics[width=0.8\textwidth]{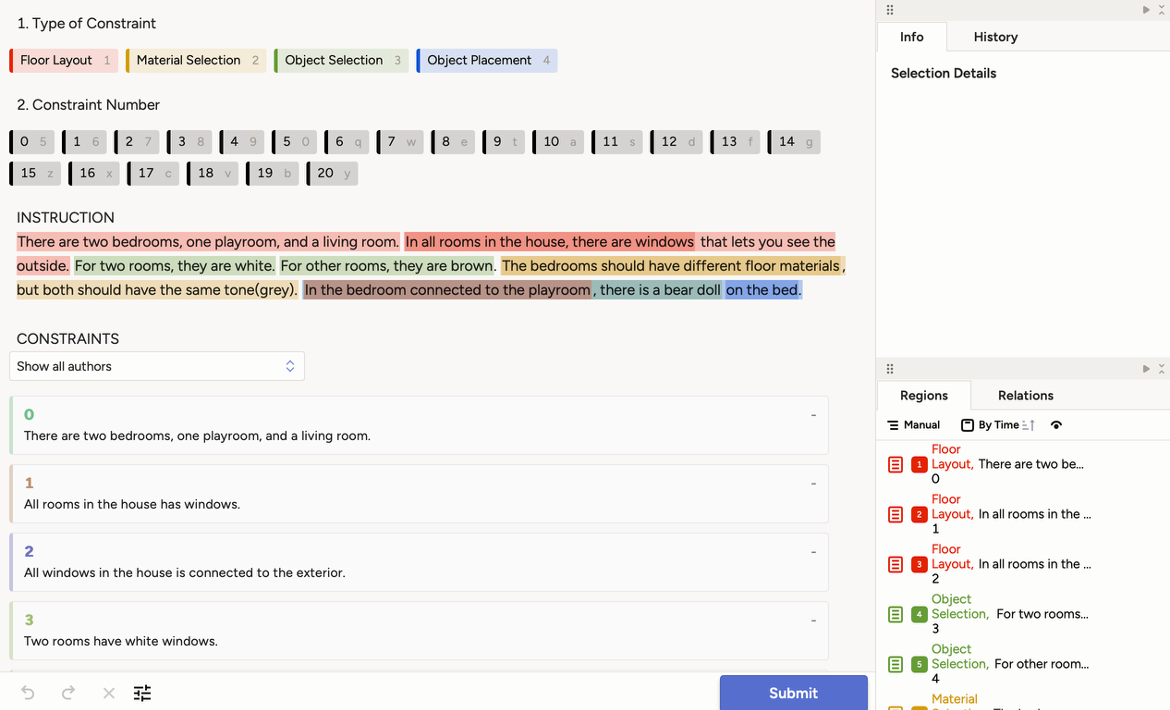}
\caption{Interface of Label Studio for constraint classification.
}
\label{fig:Annotation Tool 2}
\end{figure*}

To reduce burden of human annotation, we developed a tool for constraint identification, and use Label Stuio for constraint classification, Figure~\ref{fig:Annotation Tool 1} shows the interface of the tool for annotating the constraints within the descriptions. This tool allows the annotators to track if all constraints are accurately identified. Figure~\ref{fig:Annotation Tool 2} shows the interface of Label Studio for annotating the constraint types and mapping each constraint with the corresponding parts of the description. The constraint type is for annotating the constraints to the correct type, and constraint number is to map the constraint to the appropriate part of the description.

\subsection{Details of Human Annotation}\label{sec:appendix_HumanAnnotation}
Our data collection process follows the process below:

\textbf{Step 1: Annotator recruiting and education.} 
To collect data, we recruit human annotators to construct the dataset. All annotators completed a two-hour education session prior to annotation. This education covers a detailed explanation about guidelines for writing quality descriptions and identifying constraints, data annotation interface, and generating scenes that align with the description. After completing the training, each annotator is assigned to create descriptions, annotate them, and generate corresponding aligned scenes.

\textbf{Step 2: Data annotation.}
The annotation process is structured into three phases. In the first phase, we ask each annotator to create 30 to 50 fine-grained descriptions describing the scenes. Annotators are instructed to create descriptions that reflect real-world scenes relying on their empirical experiences.
To avoid ambiguity, we additionally require annotators to explicitly specify the viewpoint in the descriptions. For example, when the fridge is placed against a wall, we ask annotators to use expressions such as ''facing the fridge,” so that ''left” and ''right” are interpreted from that reference perspective.
\begin{figure}[t]
\centering
\includegraphics[width=0.5\textwidth]{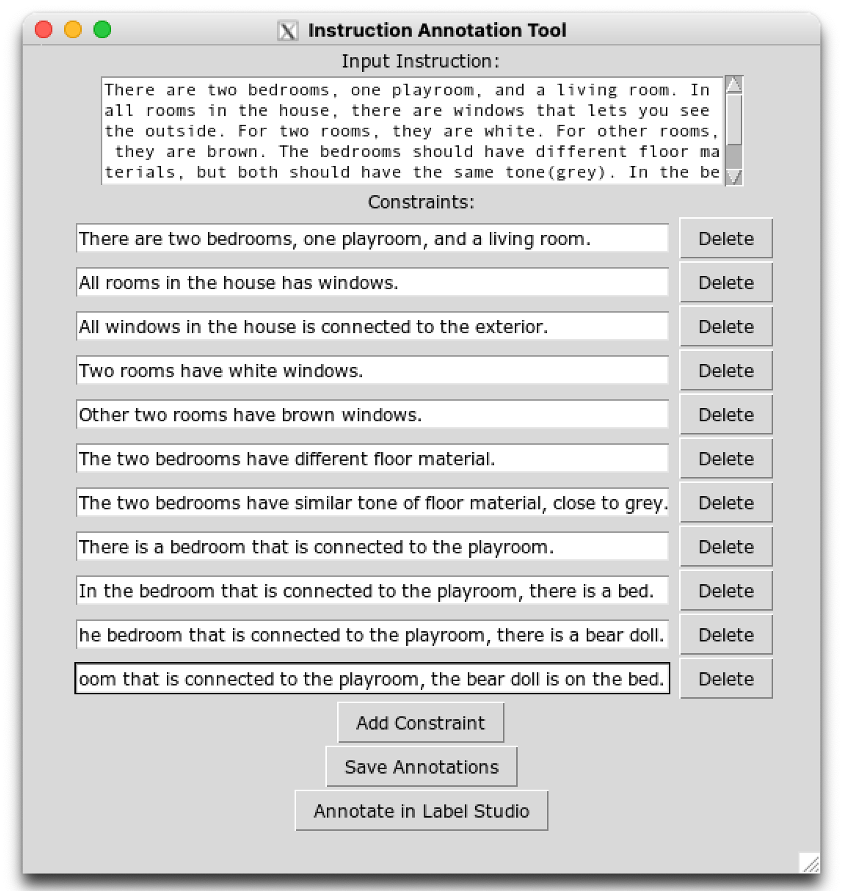}
\caption{Our annotation tool used for annotating constraints.
}
\label{fig:Annotation Tool 1}
\end{figure}

In the second phase, annotators annotate each description by identifying the constraints included in the description using our annotation tool (See Figure~\ref{fig:Annotation Tool 1}) and Label Studio(See Figure~\ref{fig:Annotation Tool 1}). 
As some constraints can only be satisfied when others are fulfilled, we express these dependencies explicitly. We achieve this by restructuring the constraints into conditional expressions that directly reference their prerequisites. For example, "\textit{There is a room with a red floor. There is a bed in that room. The bed is white.}" The sentence "The bed is white" relies on the preceding context, as it refers to the bed in the room with the red floor. This constraint can then be expressed as "The bed in the room with the red floor is white." We also merge multiple attributes of the same entity into unified statements, such as combining "the three walls are yellow" and "the three walls have patterns" into "\textit{the three walls are yellow and have patterns.}" Once the constraints have been revised, each constraint is annotated with one of four constraint types: floor layout, material selection, object selection, or object placement. We also annotate which parts of each description correspond to specific constraints, similar to entity annotation in Named Entity Recognition (NER).


In the third phase, annotators create a scene that fully validates the constraints within the description. To achieve this, we first input the descriptions into Holodeck~\citep{yang2024holodeck}, which generates scenes in textual representations (\ie JSON scripts) that can be rendered in the 3D simulator. The textual representations contain comprehensive scene information, including the asset ID of each component as well as the exact position and rotation of all scene components. As shown in Table~\ref{tab:scene_generation_methods}, Holodeck hardly generates scenes that fully align with the descriptions. Thus, annotators manually modify the text representation of the scene so that the scene fully aligns with the description. 

\textbf{Step 3: Verification.}
To ensure the quality of \bench, we conduct manual verification. Specifically, annotators conduct mutual reviews of the annotated descriptions and corresponding scenes. Annotators first verify whether the constraints have been correctly identified and whether the annotations comply with the description annotation guidelines. The descriptions are re-annotated if the constraints are not accurately identified. Annotators also verify whether the generated scene fully aligns with the description. If misalignment is found, the annotators modify the scene to ensure the alignment. The procedure is repeated for two iterations.

\subsection{Statistics}\label{sec:AppendixStatisticsLegobench}
Figure~\ref{fig:BenchStatistics} summarizes key statistics of our collected dataset. \bench~includes 130 natural language descriptions paired with manually annotated scenes, encompassing a total of 1,250 constraints. On average, each description specifies 9.6 constraints, with the majority falling between 9 and 11. These constraints span a broad range of scene elements: 55\% involve objects, while 39\% target architectural components. We further categorize constraints based on their semantics—approximately 40\% relate to material and object selection, and the remaining
60\% involve floor layout and object placement. 

\begin{figure}[t] 
    \centering
    \includegraphics[width=1\linewidth]{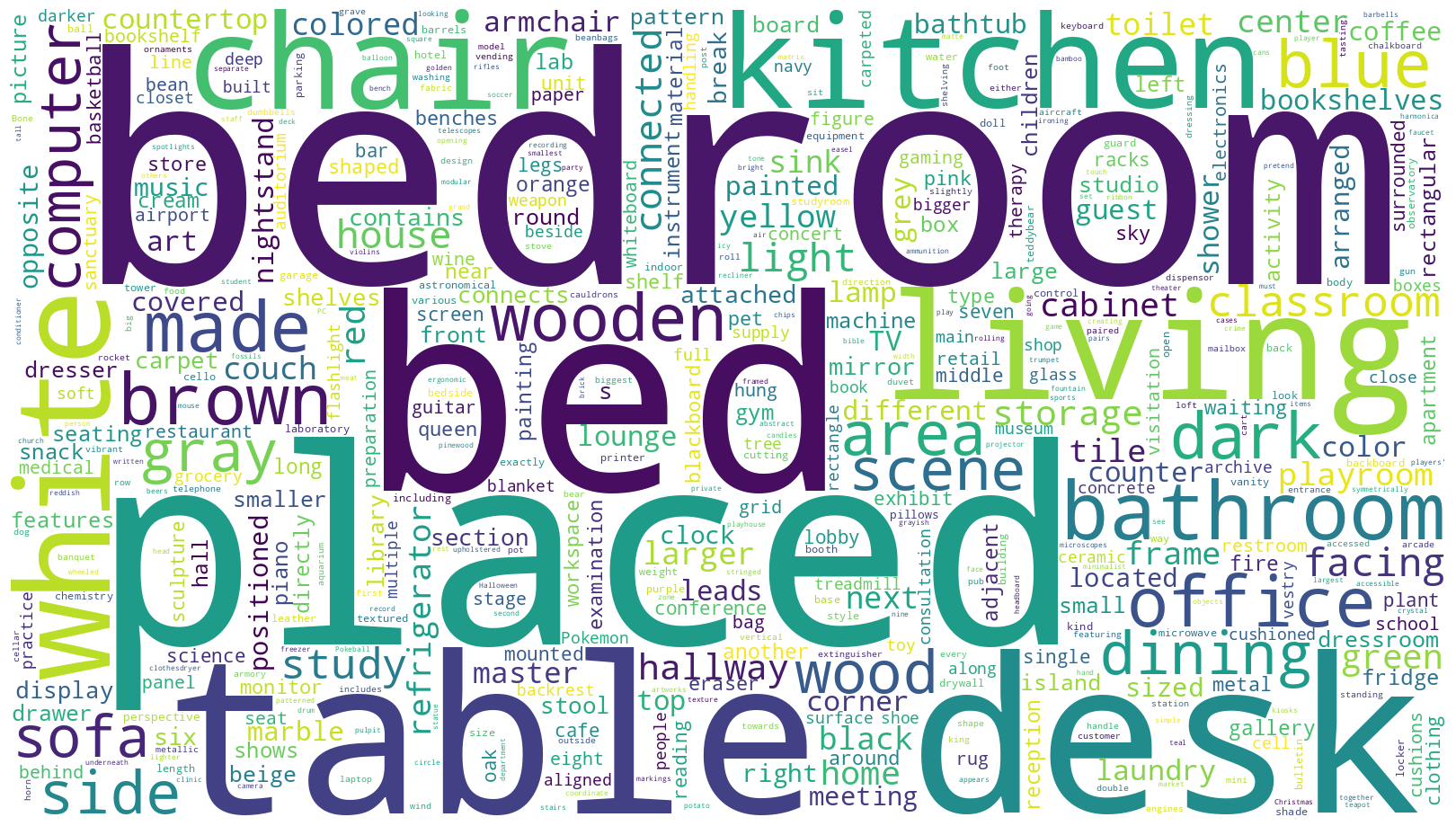}
    \caption{Word cloud illustrating the diversity of instructions in \bench.}
    \label{fig:wordcloud}
\end{figure}

To illustrate the diversity of our collected descriptions, we utilize a word cloud representation in Figure~\ref{fig:wordcloud}. The visualization highlights the diverse linguistic forms and semantic categories present in the \bench, indicating that the descriptions span across multiple domains and levels of abstraction.

Together, these distributions highlight the complexity and diversity of real-world environments captured by \bench.



\section{\eval}



We provide additional implementation details of \eval~in the following.

\subsection{Environment Setting}\label{appendix:env_setting}
We use the AI2-THOR simulator~\citep{kolve2017ai2}, which is built on the Unity engine, as our simulation environment. The simulator provides a flexible and realistic platform for embodied AI research. The environment assets are sourced from Objaverse~\citep{NEURIPS2023_70364304}, ProcTHOR~\citep{deitke2022}, and AI2-THOR~\citep{kolve2017ai2}, providing a wide range of object categories. Together, these assets enable comprehensive environment generation for our experiments.

\subsection{Model Inference}
We prompt the closed-source models, including gemini-2.5-pro, gpt-4.1, gpt-4.1-mini, and gpt-5.1 for both LLM and VLM inference. The input images are resized to 1200px resolution, and we use the default hyperparameter settings provided by the model for inference.

For the open-source model, including Gemma3-27B, Qwen2.5VL-32B, Qwen3-8B, and Qwen3-32B, the temperature is fixed at 0 to ensure deterministic outputs. The input images are resized to 335px resolution; when the number of images exceeds the model’s processing capacity, two images are concatenated into a single input, with their corresponding names displayed at the top-left corner of each image.

\subsection{Metric Details}\label{appendix:metric_details}
To aid readers’ understanding, we provide detailed explanations of the evaluation metrics below.

\begin{itemize}
    \item\textbf{F1 Score}: This metric evaluates the balance between precision and recall, serving as the harmonic mean of the two. A higher F1 score indicates better performance in correctly identifying positive instances while avoiding false positives and false negatives.
    \begin{equation}
    F1 = \frac{2 \cdot \text{Precision} \cdot \text{Recall}}{\text{Precision} + \text{Recall}}
    \end{equation}
    \item\textbf{Cohen's Kappa ($\kappa$)}: This metric measures the level of agreement between two raters, adjusting for the amount of agreement that could occur by chance. A higher $\kappa$ value indicates stronger agreement, where $\kappa = 1$ denotes perfect agreement and $\kappa = 0$ corresponds to chance-level agreement. We use Cohen’s kappa to measure the agreement with human judgments.
    \begin{equation}
    \kappa = \frac{p_o - p_e}{1 - p_e}
    \end{equation}
    where $p_o$ denotes the observed agreement between the two raters, and $p_e$ denotes the expected agreement under the assumption of statistical independence between the raters.
\end{itemize}
\subsection{Tool Set}\label{sec:Tool Set}

\begin{figure}[t]
    \centering
    \includegraphics[width=\columnwidth]{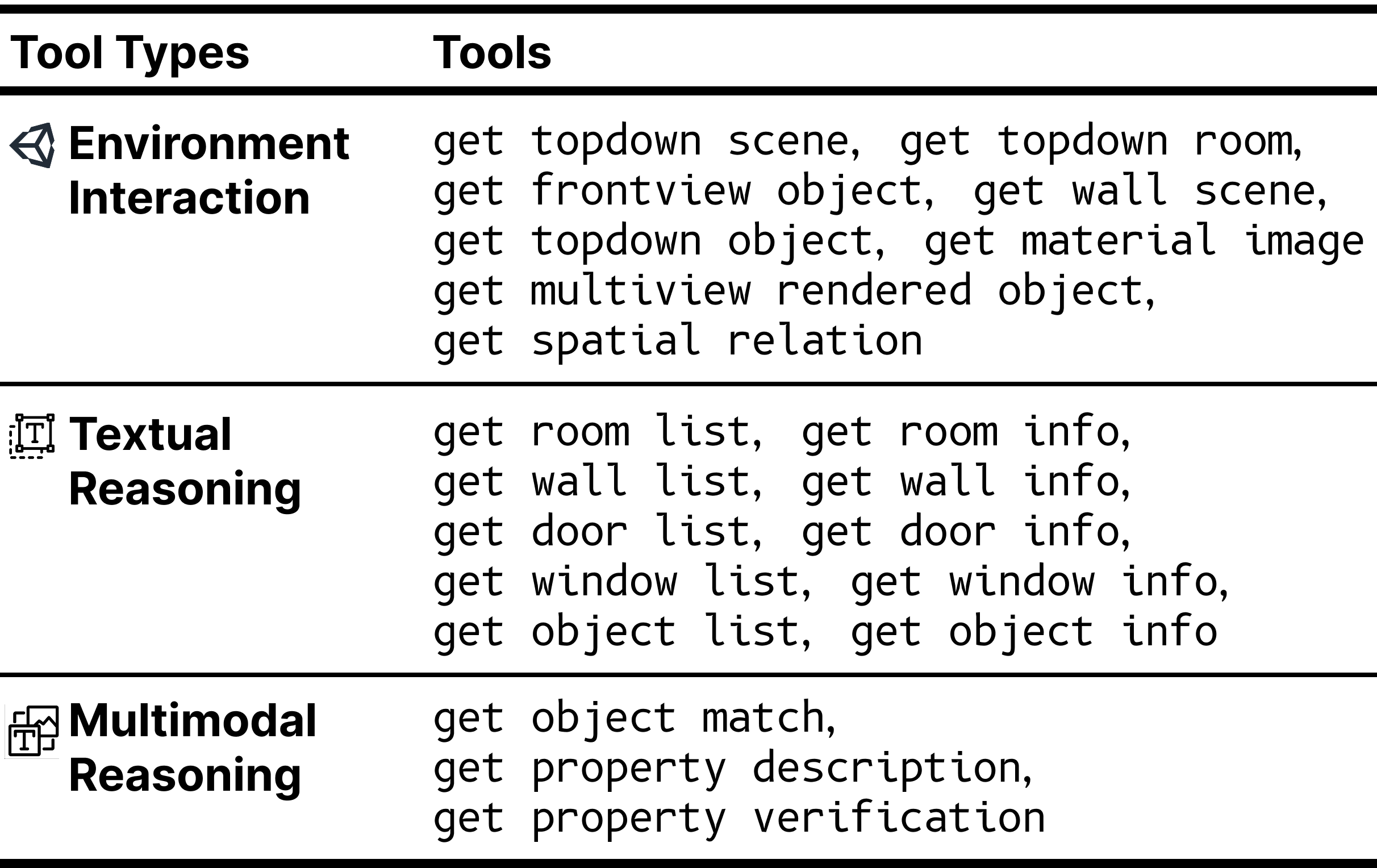}
    \caption{
    Diverse tools included in our tool set.
    \vspace{-1em}
    }
    \label{fig:toolset}
\end{figure}

As shown in Figure~\ref{fig:toolset}, \eval~utilizes total 21 tools for robust evaluation. In this section, detailed descriptions for each tool are provided. For tools that output images, example outputs of the images for each tool are also provided.

\subsubsection{Environment Interaction}

\begin{figure}[htbp]
    \centering
    \includegraphics[width=0.4\textwidth]{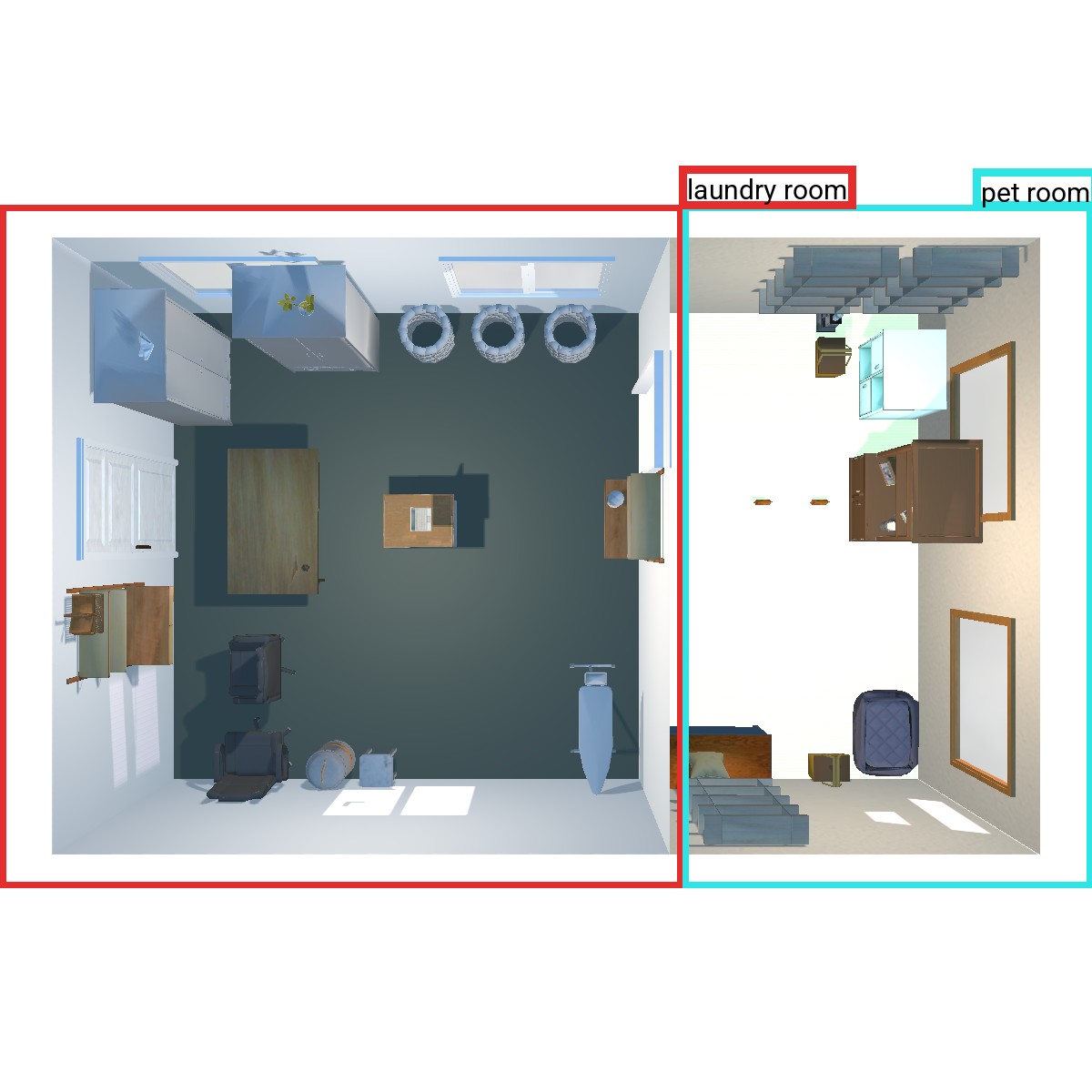}
    \caption{Example Image Output of get topdown scene.}
    \label{fig:get_topdown_scene_ex}
\end{figure}

\begin{figure}[htbp]
    \centering
    \includegraphics[width=0.4\textwidth]{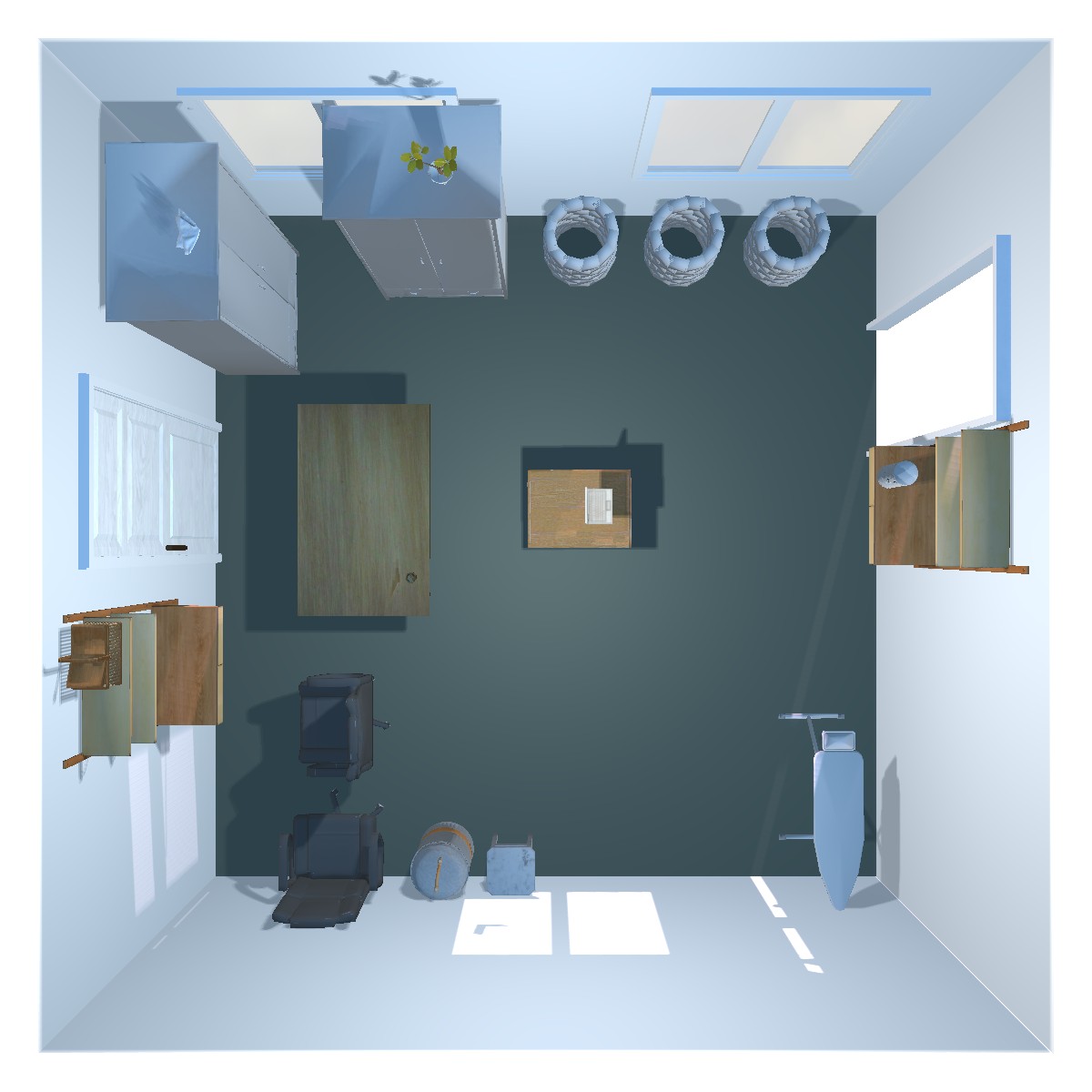}
    \caption{Example Image Output of get topdown room.}
    \label{fig:get_topdown_room_ex}
\end{figure}

\begin{figure}[htbp]
    \centering
    \includegraphics[width=0.4\textwidth]{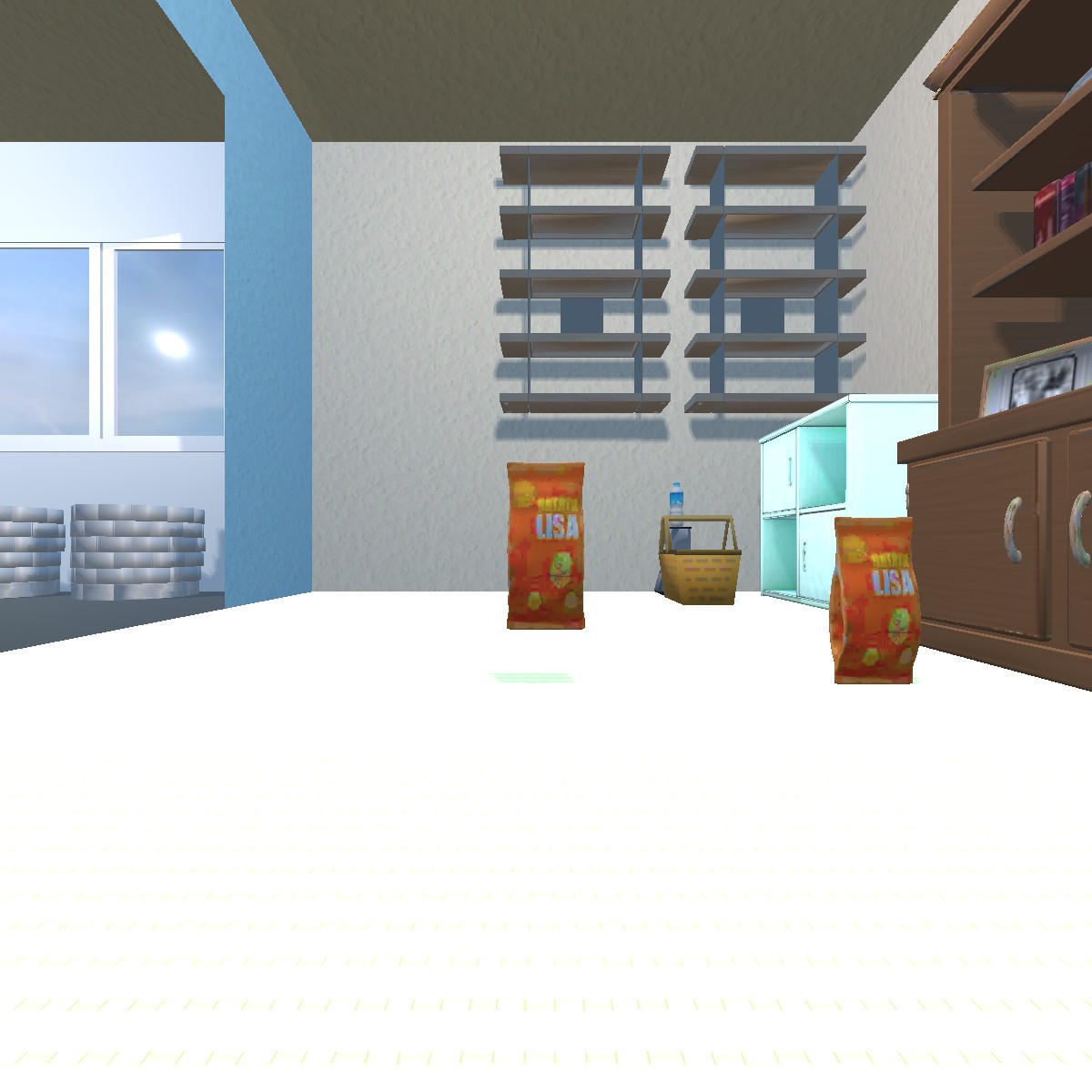}
    \caption{Example Image Output of get frontview object.}
    \label{fig:get_frontview_obj_ex}
\end{figure}

\begin{figure}[htbp]
    \centering
    \includegraphics[width=0.4\textwidth]{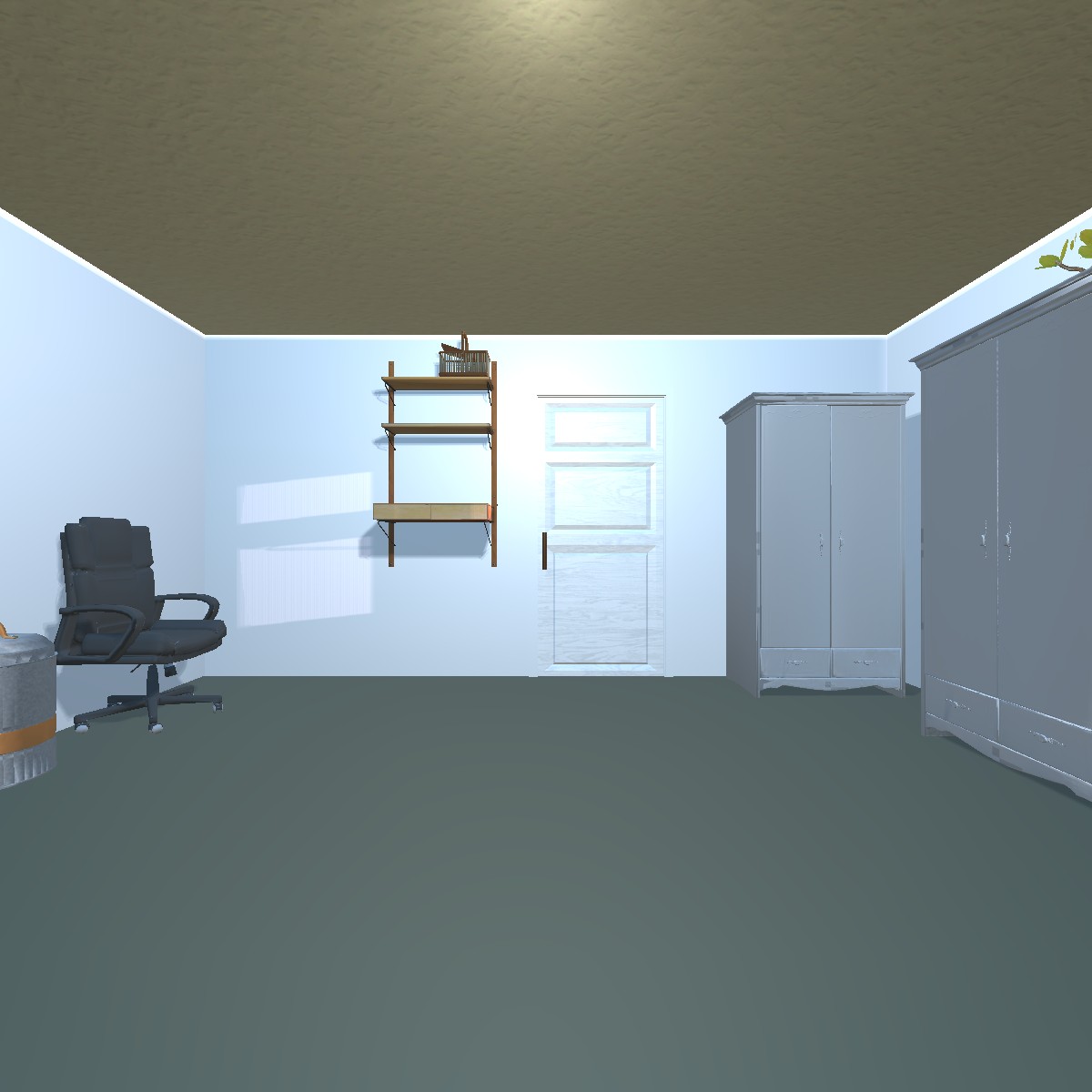}\hspace{0.05\textwidth}
    \includegraphics[width=0.4\textwidth]{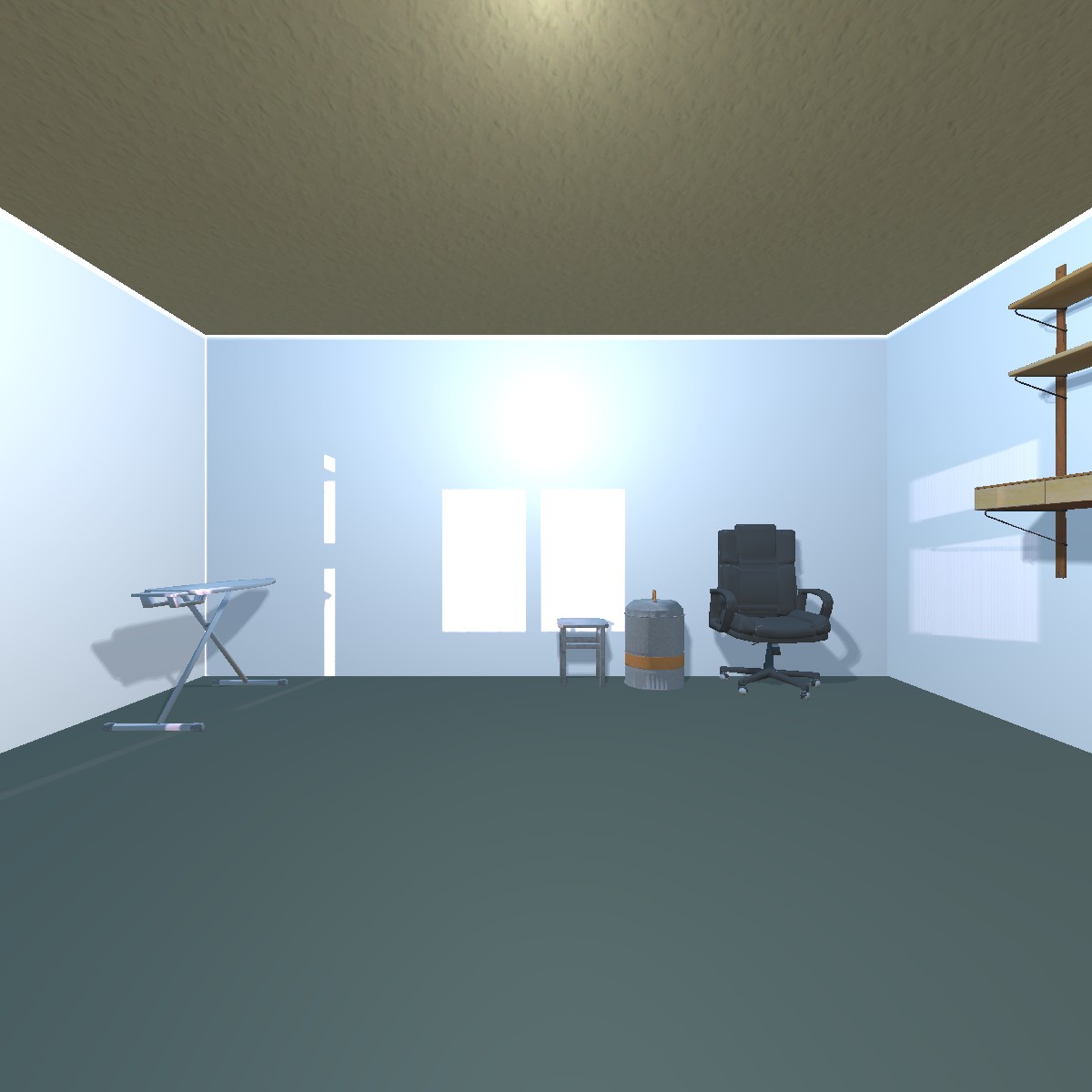}
    \caption{Example Image Output of get wall scene.}
    \label{fig:get_wall_scene_ex}
\end{figure}

\begin{figure}[htbp]
    \centering
    \includegraphics[width=0.4\textwidth]{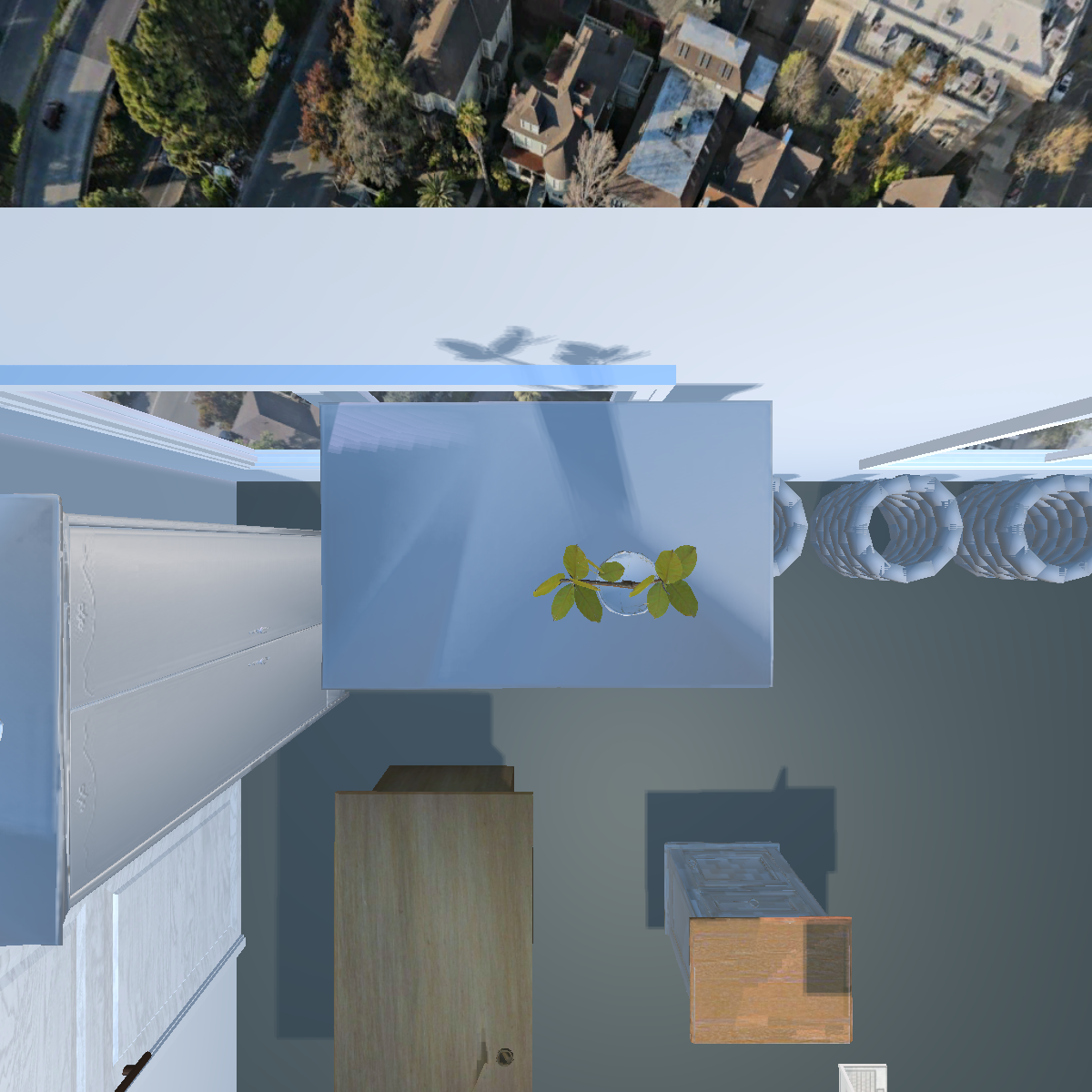}
    \caption{Example Image Output of get topdown object.}
    \label{fig:get_topdown_obj_ex}
\end{figure}

\begin{figure}[htbp]
    \centering
    \includegraphics[width=0.4\textwidth]{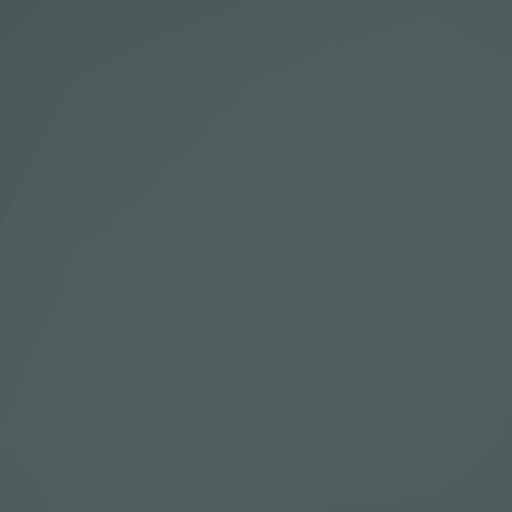}
    \caption{Example Image Output of get material image.}
    \label{fig:get_material_image_ex}
\end{figure}

\begin{figure}[htbp]
    \centering
        \includegraphics[width=0.22\textwidth]{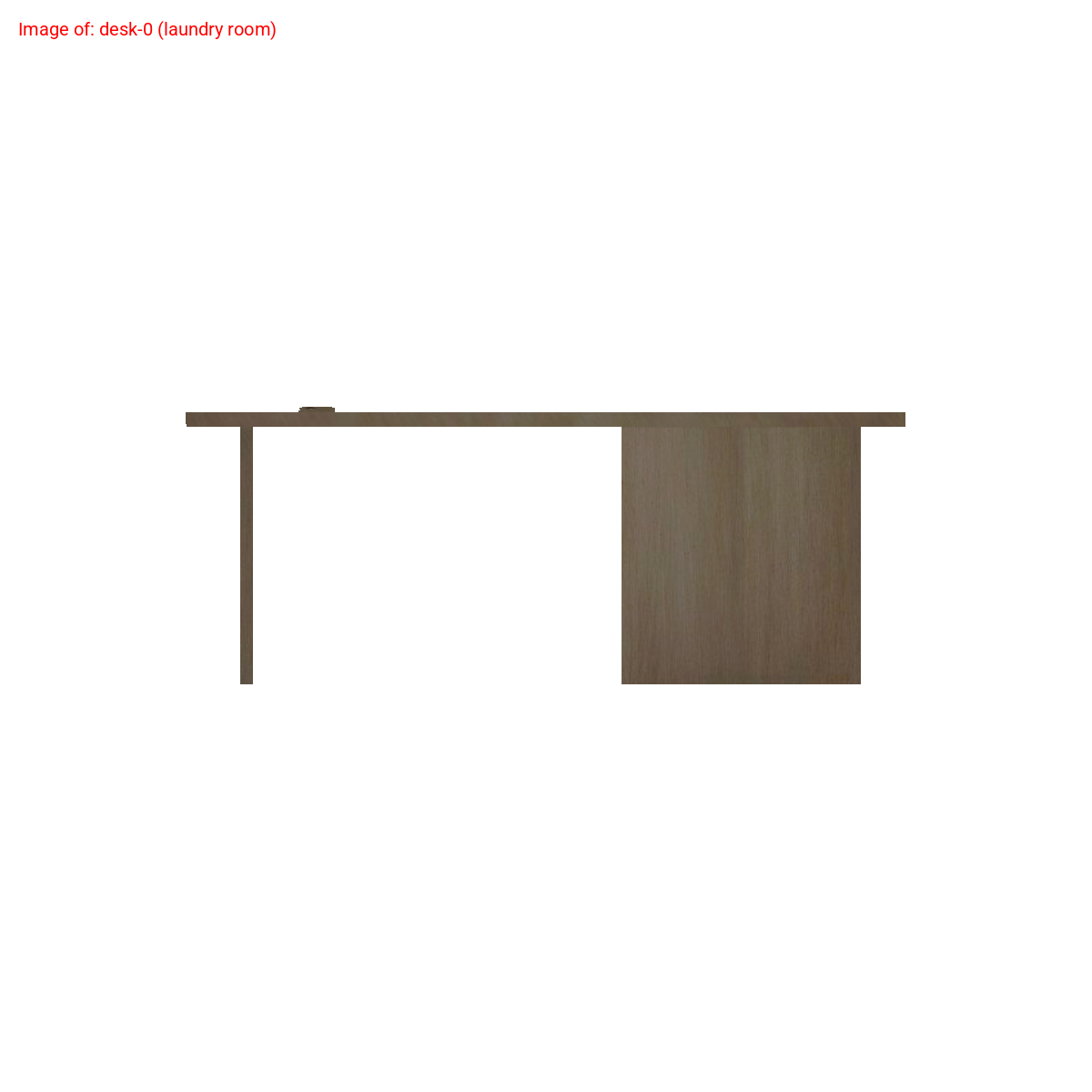}\hfill
    \includegraphics[width=0.22\textwidth]{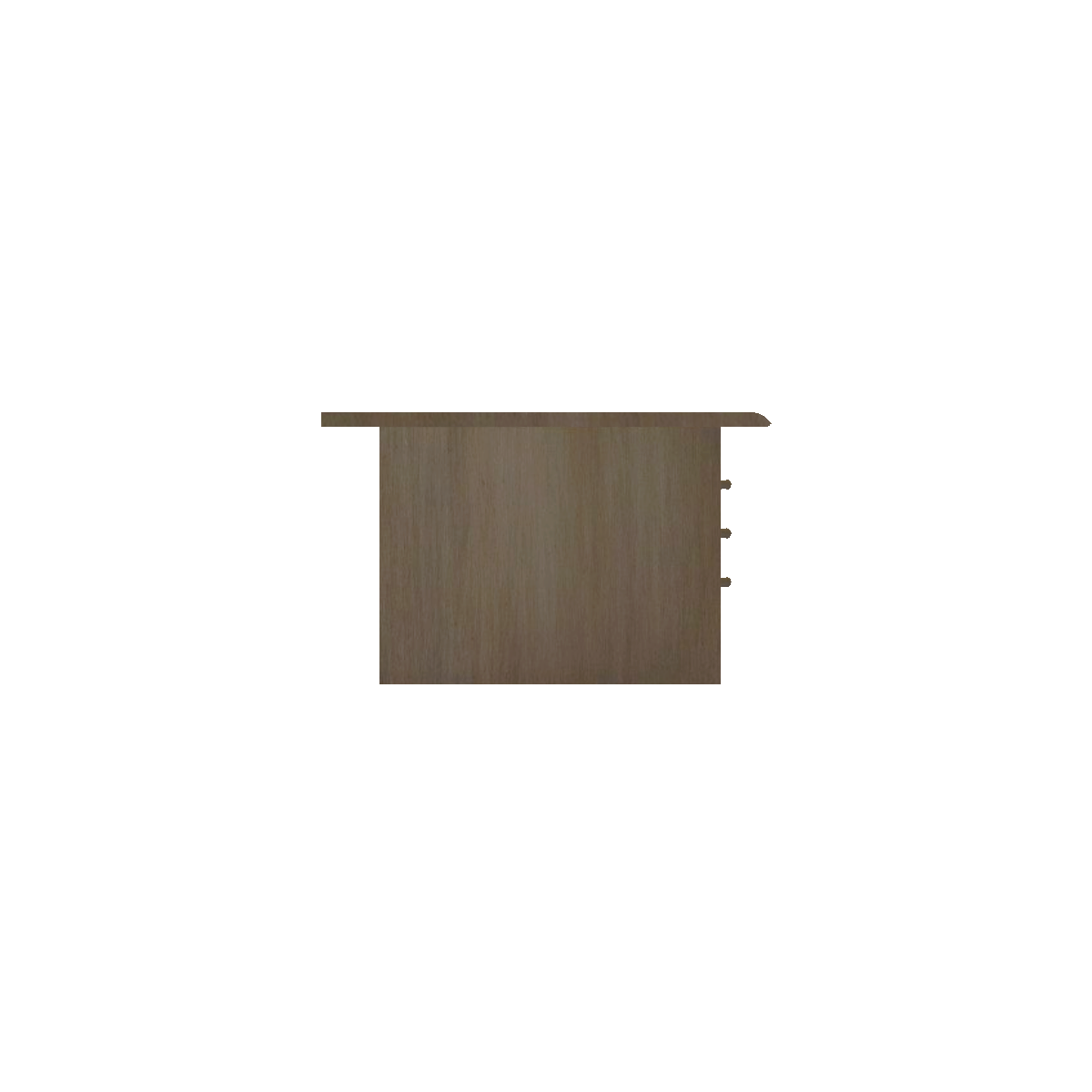}\hfill
    \includegraphics[width=0.22\textwidth]{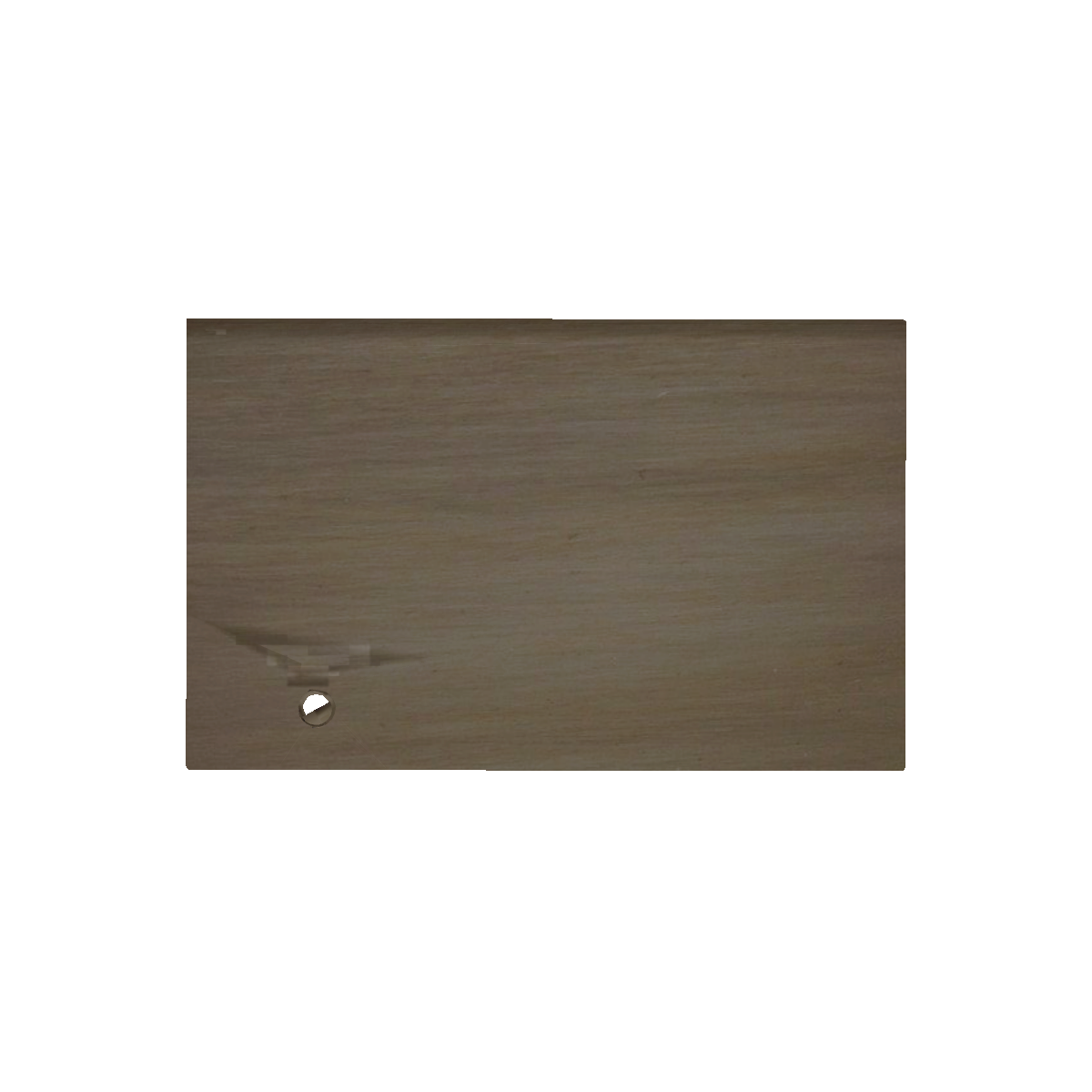}\hfill
    \includegraphics[width=0.22\textwidth]{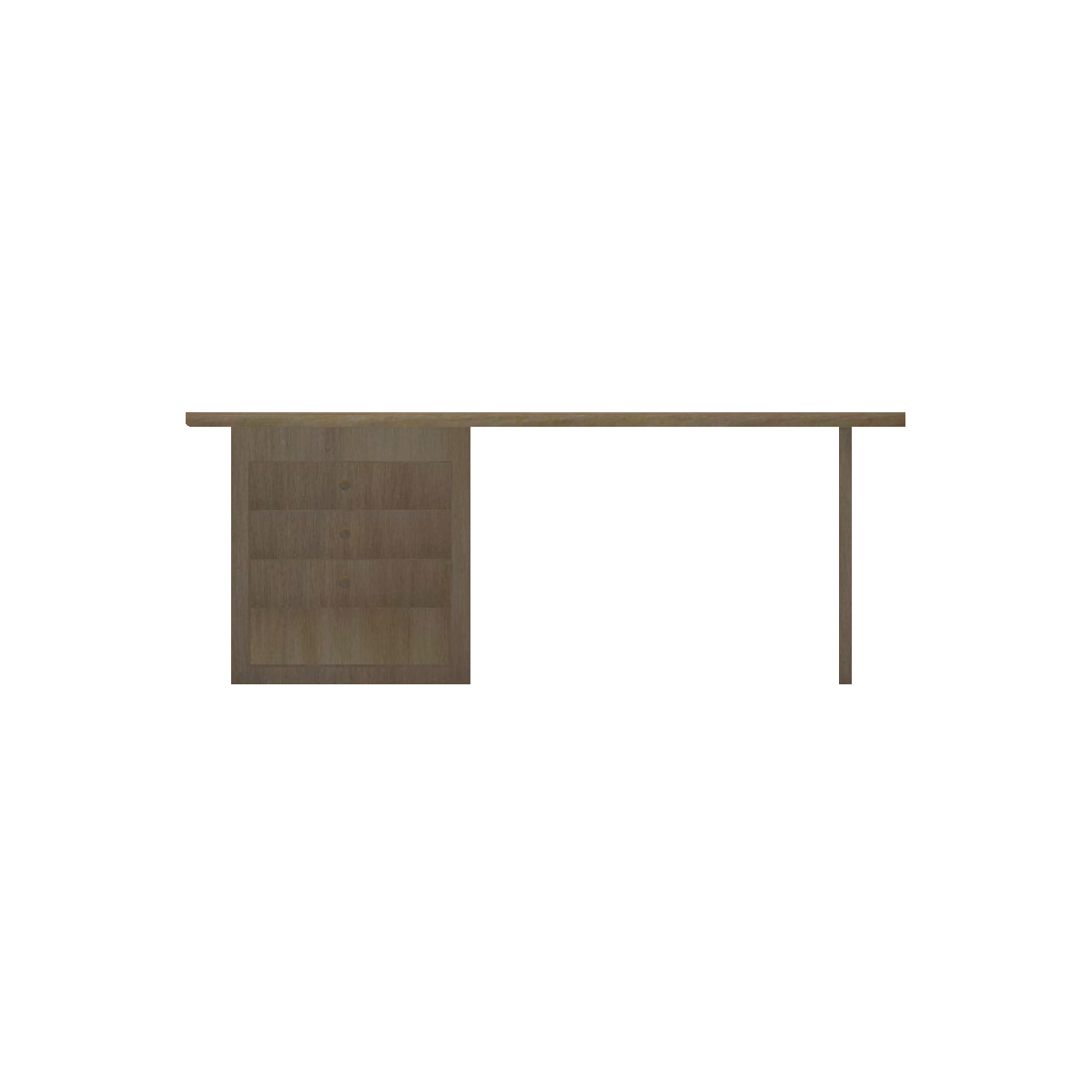}
    \caption{Example Image Output of get multiview rendered object.}
    \label{fig:get_multiview_rendered_obj_ex}
\end{figure}

\begin{figure}[htbp]
    \centering
    \includegraphics[width=0.4\textwidth]{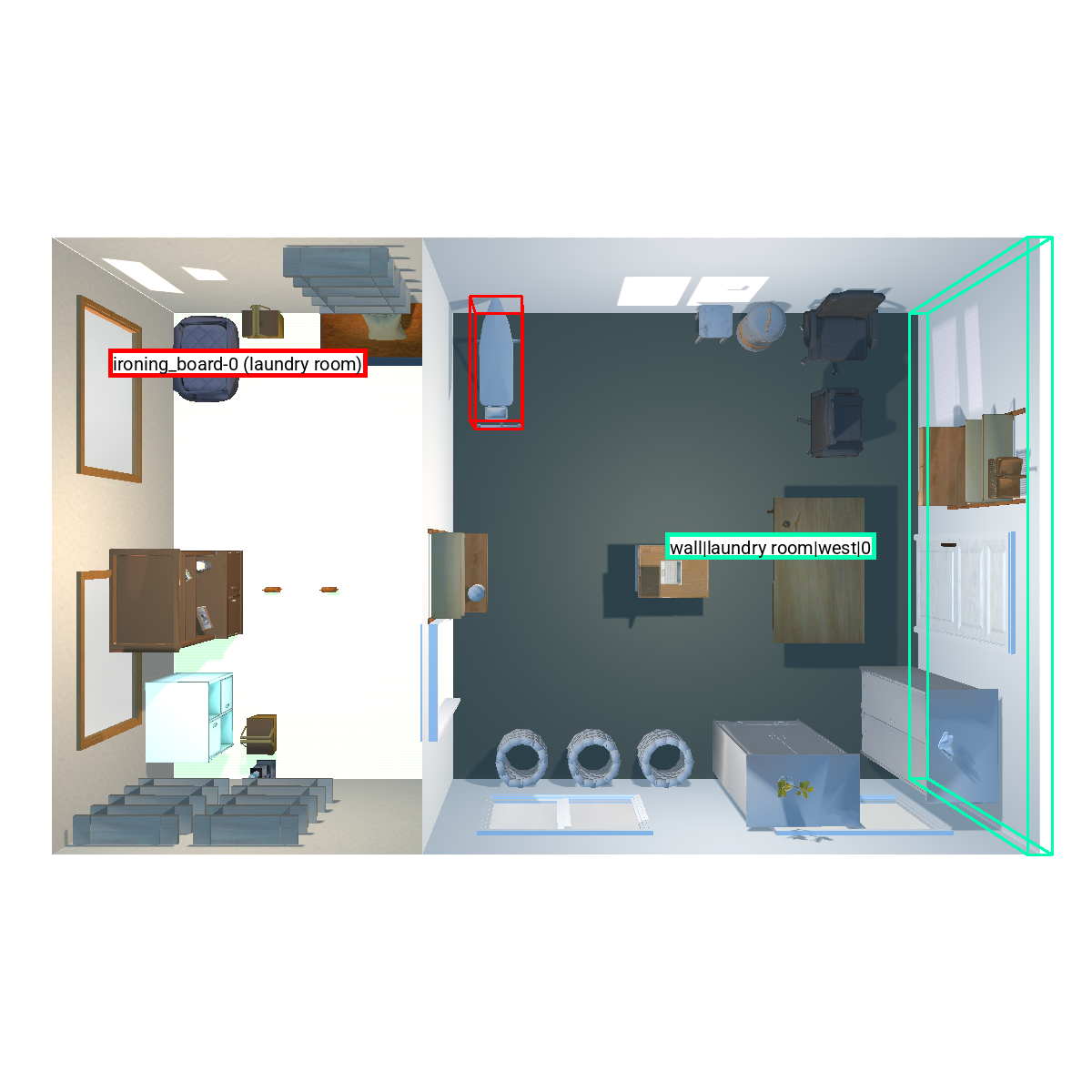}
    \caption{Example Image Output of get spatial relation.}
    \label{fig:get_spa_relation_ex}
\end{figure}

\begin{itemize}
    \item \textbf{get topdown scene:}
      This tool generates a bird’s-eye (top-down) view of the environment.
      It produces a dictionary containing an image array that represents
      the entire scene layout, including walls, doors, and objects. The example output is shown in Figure~\ref{fig:get_topdown_scene_ex}.
    \item \textbf{get topdown room:}
      This tool generates a top-down visual representation of a specific room in the environment.
      By providing the room ID as input, it produces a structured dictionary containing an image array
      that depicts the layout of the room, including walls, doors, and objects. The example output is shown in Figure~\ref{fig:get_topdown_room_ex}.
    \item\textbf{get frontview object:}   
       This tool produces front-view images of specified objects within a scene. Given a list of object IDs, it centers each target object in the frame while preserving surrounding context, such as nearby objects, for richer spatial awareness. The output is a structured dictionary mapping each object ID to its corresponding front-view image array. The example output is shown in Figure~\ref{fig:get_frontview_obj_ex}.
    \item\textbf{get wall scene:}   
       This tool generates clear wall-view images of a room from specified directions. By taking a list of wall IDs as input, it produces perspective images where objects in the middle of the room are removed to prevent occlusion, ensuring unobstructed visibility of the walls. The output is a dictionary mapping each wall to its corresponding image array. The example output is shown in Figure~\ref{fig:get_wall_scene_ex}.
    \item\textbf{get topdown object:}   
       This tool produces object-centric images from an overhead perspective. Given a list of object IDs, it generates top-down views where each target object is centered in the frame, while preserving surrounding context such as nearby objects. The resulting dictionary maps each object to image arrays, allowing inspection at different distances. The example output is shown in Figure~\ref{fig:get_topdown_obj_ex}.
    \item\textbf{get material image:}   
        This tool generates visual representations of specified materials, restricted to walls and floors within the scene. By providing a list of material names, it returns a dictionary mapping each material to an RGBA image array. The tool only includes materials that are actually present in the designated environment. The example output is shown in Figure~\ref{fig:get_material_image_ex}.
    \item\textbf{get multiview rendered object:}   
        This tool produces rendered images of specified objects using their underlying asset IDs. Given a list of object IDs, it returns a dictionary where each object ID maps to an image array showing the rendered appearance of that object. The example output is shown in Figure~\ref{fig:get_multiview_rendered_obj_ex}.
    \item\textbf{get spatial relation:}   
        This tool generates scene images that isolate and display only specified object pairs (or sets), such as objects, windows, doors, or walls. By taking a list of tuples containing object IDs, it produces a dictionary mapping each group of objects to an image array. Values of the dictionary provide visualizations that highlight the spatial relationships among the selected elements. The example output is shown in Figure~\ref{fig:get_spa_relation_ex}.
\end{itemize}

\subsubsection{Multimodal Reasoning}

\begin{itemize}
    \item \textbf{get object match:}
        This tool provides semantic descriptions for object IDs by mapping them to their corresponding object types. It is designed to be used after retrieving rendered views with \texttt{get multiview rendered object}, ensuring accurate alignment between visual assets and their categorical labels. The output is a dictionary that links each object ID to a descriptive type.

    \item \textbf{get property description:}
        This tool generates semantic descriptions of objects by identifying their color, shape, and material from rendered images and optional textual metadata. It is intended for use after retrieving visual inputs with \texttt{get material image} and \texttt{get multiview rendered object}, ensuring accurate grounding of appearance-based attributes. The output is a dictionary summarizing the derived text information, along with reasoning.
        
    \item \textbf{get property verification:}
        This tool validates descriptive attributes of a subject—either an object or material—based on a given instruction. First, an LLM parses the instruction to determine which attributes should be checked (e.g., surface patterns or textures). Then, a VLM analyzes the corresponding images to extract those attributes. It is designed for use after obtaining visuals with \texttt{get material image} and \texttt{get multiview rendered object}. The output is a dictionary summarizing the verified textual information.  
\end{itemize}

\subsubsection{Textual Reasoning}

\begin{table*}[!t]
\centering
\footnotesize
\setlength{\tabcolsep}{5pt}
\begin{tabular}{>{\raggedright\arraybackslash}m{3cm}|cccc|cccc}
\toprule
\multicolumn{1}{c|}{\multirow{2}{*}{\textbf{Methods}}} &
\multicolumn{4}{c|}{\textbf{Holistic SR}} &
\multicolumn{4}{c}{\textbf{Partial SR}} \\
\cmidrule(lr){2-5}\cmidrule(lr){6-9}
& \textbf{M1} & \textbf{M2} & \textbf{M3} & \textbf{M4}
& \textbf{M1} & \textbf{M2} & \textbf{M3} & \textbf{M4} \\
\midrule
Oracle Constraint     & 0.12 & 0.05 & 0.05 & 0.14 & 0.66 & 0.57 & 0.35 & 0.65 \\
Identified Constraint & 0.13 & 0.07 & 0.05 & 0.12 & 0.63 & 0.57 & 0.35 & 0.64 \\
\midrule
Difference in SR      & +0.01 & +0.02 & +0.00 & -0.02 & -0.03 & +0.00 & +0.00 & -0.01 \\
\bottomrule
\end{tabular}
\caption{End-to-end evaluation maintains consistent results with annotated constraint evaluation.}
\label{tab:end_to_end}
\end{table*}
\begin{itemize}
    \item \textbf{get room list:}
        This tool extracts all room identifiers available within a given scene. It returns a dictionary which contains a list of room IDs as strings.
        
    \item \textbf{get room info:}
        This tool provides detailed metadata for specific rooms in a scene. Given a dictionary from \texttt{get room list} containing room IDs, it returns structured information about each room. The output includes the polygon vertices defining the room’s geometry and the identified floor material.
        
    \item \textbf{get wall list:}
        This tool extracts the wall identifiers associated with specific rooms in a scene. Given a list of room IDs, it returns a dictionary mapping each room ID to a list of its connected wall IDs.
        
    \item \textbf{get wall info:}
        This tool provides comprehensive metadata for specified walls in a scene. Using a list of wall IDs (retrieved via \texttt{get wall list}), it returns the values including details such as the wall’s unique ID, connected room IDs, geometric coordinates, material type, width, height, and directional orientation. 

    \item \textbf{get door list:}
        This tool extracts the door identifiers connected to specific rooms within a scene. Given a list of room IDs, it returns a dictionary mapping each room ID to its associated door IDs.
        
    \item \textbf{get door info:}
        This tool provides detailed metadata for doors within a scene. Using a list of door IDs (obtained via \texttt{get door list}), it returns the values capture attributes such as the door’s unique ID, associated asset ID, connected rooms, adjoining walls, 3D position, and openness state. 
        
    \item \textbf{get window list:}
        This tool extracts the window identifiers associated with specific rooms in a scene. Given a list of room IDs, it returns a dictionary mapping each room ID to its corresponding window IDs. 
        
    \item \textbf{get window info:}
        This tool provides detailed metadata for windows in a scene. Given a list of window IDs (retrieved via \texttt{get window list}), it outputs the values include attributes such as the window’s unique ID, associated asset ID, connected rooms, adjoining walls, and 3D position. 
        
    \item \textbf{get object list:}
        This tool extracts the identifiers of objects contained within specific rooms of a scene. Given a list of room IDs, it returns a dictionary mapping each room ID to the corresponding list of object IDs present in that room.

    \item \textbf{get object info:}
        This tool provides detailed metadata about objects in a scene. Using a list of object IDs (retrieved via \texttt{get object list}), it returns the values include attributes such as the object’s unique ID, associated asset ID, room assignment, 3D position, rotation, and geometric representation through coordinates or mesh vertices.

\end{itemize}

\section{Additional Results}
We provide a series of analyses on \eval~to better understand its behavior.

\subsection{\eval~Provides End-to-End Automated Evaluation}
To demonstrate the effectiveness of \eval, we show that it performs reliably even without human-annotated constraints. Accurate, fine-grained evaluation requires capturing all relevant constraints in a given description, so we compare \eval's performance using its own automatically identified constraints against its performance with human-annotated ones. In our setup, constraints are extracted and categorized using GPT-4.1, then used to evaluate 3D scene generation outputs from Qwen2.5-VL-32B. These evaluations are directly compared with those conducted using human-annotated constraints. As shown in Table~\ref{tab:end_to_end}, \eval~exhibits only minor performance differences across four 3D scene synthesis methods when using automatic versus human-provided constraints. This close alignment demonstrates that \eval~can reliably identify and classify constraints, making it suitable as a fully automated, end-to-end evaluation tool. 

\subsection{\eval~Plans Tool Execution Effectively}
We analyze how two core components of the evaluation process—tool execution planning and argument selection—contribute to overall performance, using all constraints introduced in Section~\ref{sec:comparison_of_eval_methods}. For tool planning, \eval~generates execution plans conditioned on the given constraints. For argument selection, \eval~selects arguments given a human-annotated tool execution plan. We test multiple LLMs for each component to observe their contributions, using Qwen2.5VL-32B as a fixed validator. We use three metrics to assess component performance: (1) Tool F1, the macro F1 score over constraints, where a tool prediction is correct if it appears in the ground-truth set; (2) Argument F1, the macro F1 over tools, with correctness based on ground-truth arguments; and (3) Graph Edit Distance (GED), measuring the number of edits needed to align predicted and ground-truth constraint graphs. Results in Table~\ref{tab:component_analysis} show that tool execution planning correlates more strongly with overall evaluation performance than argument selection, highlighting its role as an orchestrator. When tool plans are fixed to ground truth, argument selection correlates more with performance, showing its dependence on accurate planning (See Section~\ref{appendix:ArgumentSelectionF1}). These results suggest that \eval~handles tool planning well, which in turn supports accurate argument selection and evaluation.



\begin{table*}[t]
\centering
\begin{tabular}{c c c c >{\columncolor{gray!15}}c >{\columncolor{gray!15}}c}
\toprule
\multirow{2}{*}{\textbf{Models}} &
\multicolumn{3}{c}{\textbf{Component Performance}} &
\multicolumn{2}{>{\columncolor{gray!15}}c}{\textbf{Evaluation Performance}} \\
\cmidrule(lr){2-4} \cmidrule(lr){5-6}
& \textbf{Tool F1} $\uparrow$
& \textbf{GED} $\downarrow$
& \textbf{Argument F1} $\uparrow$
& \textbf{Holistic F1} $\uparrow$
& \textbf{Partial F1} $\uparrow$ \\
\midrule
Gemma3-27B    & 0.50 & 3.01 & 0.38 & 0.61 & 0.68 \\
Qwen2.5-VL-32B & 0.57 & 2.87 & 0.46 & 0.64 & 0.72 \\
Qwen3-32B     & 0.62 & 2.55 & 0.42 & 0.69 & 0.73 \\
\bottomrule
\end{tabular}
\caption{The performance of components are highly correlated with the evaluation performance. Tool F1 and argument F1 measure prediction accuracy; GED measures graph structural similarity.}
\label{tab:component_analysis}
\end{table*}


\subsection{Argument Selection Plays a Critical Role in Evaluation Performance}\label{appendix:ArgumentSelectionF1}
To further analyze the contribution of argument selection in our evaluation pipeline, we conduct an experiment where the tool sequence is fixed to human-annotated ground truth, and models are only responsible for selecting appropriate arguments for each tool. All other components of \eval, including tool execution and validation, use Qwen2.5-VL-32B as the backbone.

We evaluate three open-source models Gemma3-27B, Qwen2.5-VL-32B, and Qwen3-32B on their ability to perform argument selection. As shown in Table~\ref{tab:arg_selection_with_human}, Qwen2.5-VL-32B achieves the highest Argument F1 score, aligning with superior holistic and partial evaluation performance. Qwen3-32B also performs competitively, while Gemma3-27B lags behind across all metrics.

These results highlight that accurate argument selection is crucial for reliable evaluation. Since argument identification directly determines the quality of evidence retrieved by tools, improvements in this component significantly enhance both holistic and partial agreement with human judgments.

\begin{table}[t]
\centering
\footnotesize
\begin{tabular}{c c c c}
\toprule
\multirow{2}{*}{\textbf{Models}} &
\multicolumn{3}{c}{\textbf{Argument Selection}} \\
\cmidrule(lr){2-4}
& \textbf{Arg. F1} $\uparrow$
& \textbf{Hol. F1} $\uparrow$
& \textbf{Part. F1} $\uparrow$ \\
\midrule
Gemma3-27B    & 0.38 & 0.65 & 0.66 \\
Qwen2.5VL-32B & 0.46 & 0.68 & 0.68 \\
Qwen3-32B     & 0.42 & 0.68 & 0.67 \\
\bottomrule
\end{tabular}
\caption{Evaluation performance of models on argument selection.}
\label{tab:arg_selection_with_human}
\end{table}

\subsection{Tool Usage Distributions Differ Across Constraint Types}
\begin{figure*}[t]
\centering
\includegraphics[width=0.8\textwidth]{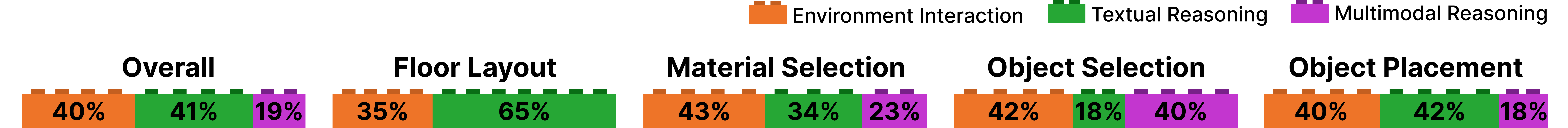}
\caption{Distribution of tool types executed by \eval~during evaluation.
}
\label{fig:tool_dist}
\end{figure*}

Extending the analysis in Section~\ref{sec:ablation_studies}, we examine which tool types are used in \eval~across different constraint categories. As shown in Figure~\ref{fig:tool_dist}, environment interaction tools and textual reasoning tools constitute the majority of tool usage, indicating that most constraints require both visual and textual information from the generated scene for evaluation. In contrast, multimodal reasoning tools are used less frequently on average. However, their usage becomes substantially higher for object placement constraints, where they account for approximately 40\% of all tool calls. This increase is primarily driven by the need for object identification and attribute verification, which rely on aligning object descriptions with visual observations. These results suggest that different constraint types require distinct combinations of tool capabilities, highlighting the importance of supporting diverse tool modalities for 3D scene and instruction alignment evaluation.

\subsection{Error Analysis of VLM-as-a-Judge with Different Modules}\label{appendix:vlm_modular}
We provide details of the experiment conducted in Section~\ref{ssec:failure_mode}. We conduct a error analysis of VLM-as-a-Judge using different perception modules. Specifically, we compare the vanilla VLM-as-a-Judge with three variants that incorporate additional scene understanding modules: Grounding DINO, LASER, and Point-LLM. We construct a fixed evaluation set by randomly sampling 50 instruction–scene pairs with aligned instructions and 50 with misaligned instructions from the evaluation dataset from Section~\ref{sec:comparison_of_eval_methods}, resulting in 100 total cases. As shown in Table~\ref{tab:error_comparison}, the vanilla VLM-as-a-Judge exhibits the highest number of identification and attribute errors. Although Grounding DINO is used to localize scene components mentioned in the instruction by adding bounding boxes to the input image, it unexpectedly increases identification errors. Similarly, incorporating LASER to generate scene graphs as additional inputs does not improve performance and instead leads to more errors across all categories compared to the vanilla setting. Finally, we convert scenes into point clouds and feed them into Point-LLM, which slightly reduces spatial errors but increases both identification and attribute errors. Detailed descriptions of each method are provided in Appendix~\ref{appendix:VLM-as-a-Judge with Different Modules}.

\subsection{Effect of \eval~on Embodied Agent Training}\label{appendix:Practical}
To show the practical value of instruction-aligned scene generation and the effectiveness of \eval, we conduct a small-scale embodied-agent training experiment, where we train embodied agents on generated scenes with varying levels of instruction-scene alignment. 
\subsubsection{Setup}
\paragraph{Task.} We consider an object pickup task in which multiple identical objects are placed on a table, and the agent must pick up the object that satisfies the spatial relation specified in the instruction (e.g., \textit{Pick up the {A} that is closer to the {B}} or \textit{Pick up the {A} that is between the {B} and the {C}}). The agent uses the same high-level action space as ALFRED {(e.g., \textit{MoveAhead}, \textit{RotateLeft}, \textit{RotateRight}, and \textit{PickupObject})}.

\paragraph{Scene generation.} We first generate 300 unique task instructions by filling the placeholders in 12 predefined instruction templates (e.g., \textit{Pick up the {A} that is closer to the {B}}) with 15 predefined object categories (e.g., \textit{apple} and \textit{wallet}), and convert each into a scene description. Each scene generated from these descriptions contains a table with 2–4 objects on it, varying only in object placement. Of the 300 scenes, 200 are instruction-aligned and 100 are intentionally misaligned (e.g., the target object is placed in an incorrect spatial relation, or a reference object is omitted), while always leaving at least two candidate targets.

\paragraph{Training trajectories.} We split the 200 aligned scenes into 100 training scenes and 100 test scenes. For each training scene, we predefine three starting locations and collect expert trajectories in which the agent navigates to the table and picks up the correct object. For misaligned scenes, we similarly collect trajectories, except that one of the candidate target objects is selected at random because the instruction cannot be satisfied uniquely. This results in 300 training trajectories for the aligned set and 300 training trajectories for the misaligned set.

\paragraph{Training configurations.}
We use the Episodic Transformer (E.T.)~\citep{pashevich2021episodic}, a strong baseline for ALFRED, and further fine-tune its released checkpoint on our collected trajectories for 50 epochs. To select training scenes, we run \eval{} with 200 scenes, which labels each scene as aligned or misaligned. We then randomly sample 100 scenes that~\eval{} labeled as aligned. We compare the following four settings: (1) zero-shot evaluation using the original E.T. checkpoint; 
(2) fine-tuning on 100 misaligned scenes; 
(3) fine-tuning on 100 scenes consisting of 50 aligned and 50 misaligned scenes; and 
(4) fine-tuning on 100 scenes that \eval{} predicts as instruction-aligned.

\subsubsection{Results}
\begin{table}[t]
    \centering
    \begin{tabular}{lc}
        \toprule
        \textbf{Training Data} & \textbf{Success Rate (\(\uparrow\))} \\
        \midrule
        None (zero-shot) & 0\% \\
        100 Misaligned & 4\% \\
        50 Aligned + 50 Misaligned & 9\% \\
        100 \eval-Selected & \textbf{10\%} \\
        \bottomrule
    \end{tabular}
    \caption{Success rate of E.T. with different training data.}
    \label{tab:practical_training_data}
\end{table}
The results in Table~\ref{tab:practical_training_data} shows that increasing the proportion of instruction-aligned scenes consistently improves downstream task performance. The success rate increases from 4\% when training with only misaligned scenes to 9\% with 50 aligned scenes, and further improves to 10\% when training on scenes selected by \eval.

These results suggest two key findings: (1) instruction-scene alignment has a measurable impact on embodied-agent training, and (2) \eval reliably identifies the scenes that are most beneficial for downstream policy learning. While limited in scale given the rebuttal period, it demonstrates that \eval{} is not merely an intrinsic evaluation metric, but a practical tool for selecting higher-quality training data that translates into improved downstream embodied-agent performance.

\subsection{Humans Usually Provide Detailed Descriptions of the Room}\label{appendix:Survey}
\begin{figure}[htbp]
    \centering
    \includegraphics[width=0.45\textwidth]{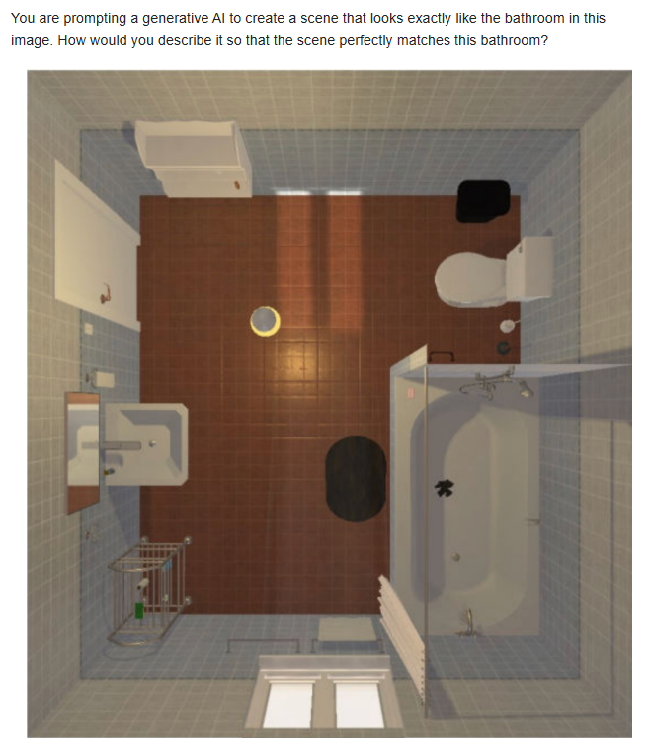}
    \includegraphics[width=0.45\textwidth]{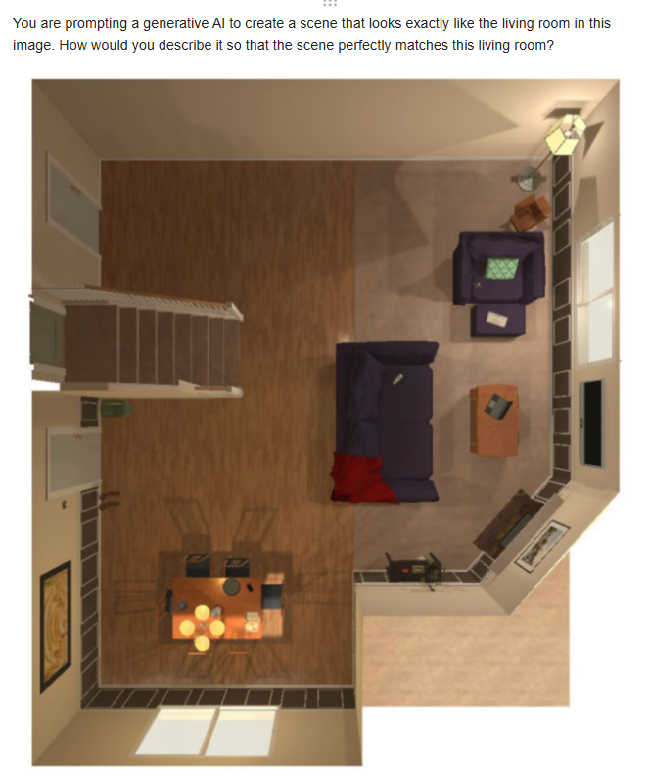}
    \caption{Our survey examples for asking human to describe the room.}
    \label{fig:survey_example}
\end{figure}

We verify whether people actually provide descriptions to an LLM for room generation that are as long as we claimed. Therefore, as shown in Figure~\ref{fig:survey_example}, we conduct a survey where participants are asked to describe a room photo in the form of a prompt. The results show that, on average, the scene descriptions contain 18.2 constraints. This finding confirms that people indeed provide complex descriptions, and highlights the importance of \eval, which can robustly evaluate generated rooms based on such descriptions.

\section{Method Details} 
In this section, we describe in detail the methods employed for scene evaluation and scene generation in our experiments.

\subsection{Scene Evaluation}
\subsubsection{Overview of Existing Evaluation Methods}
\begin{figure}[t]
    \centering
    
    \includegraphics[width=0.48\columnwidth]{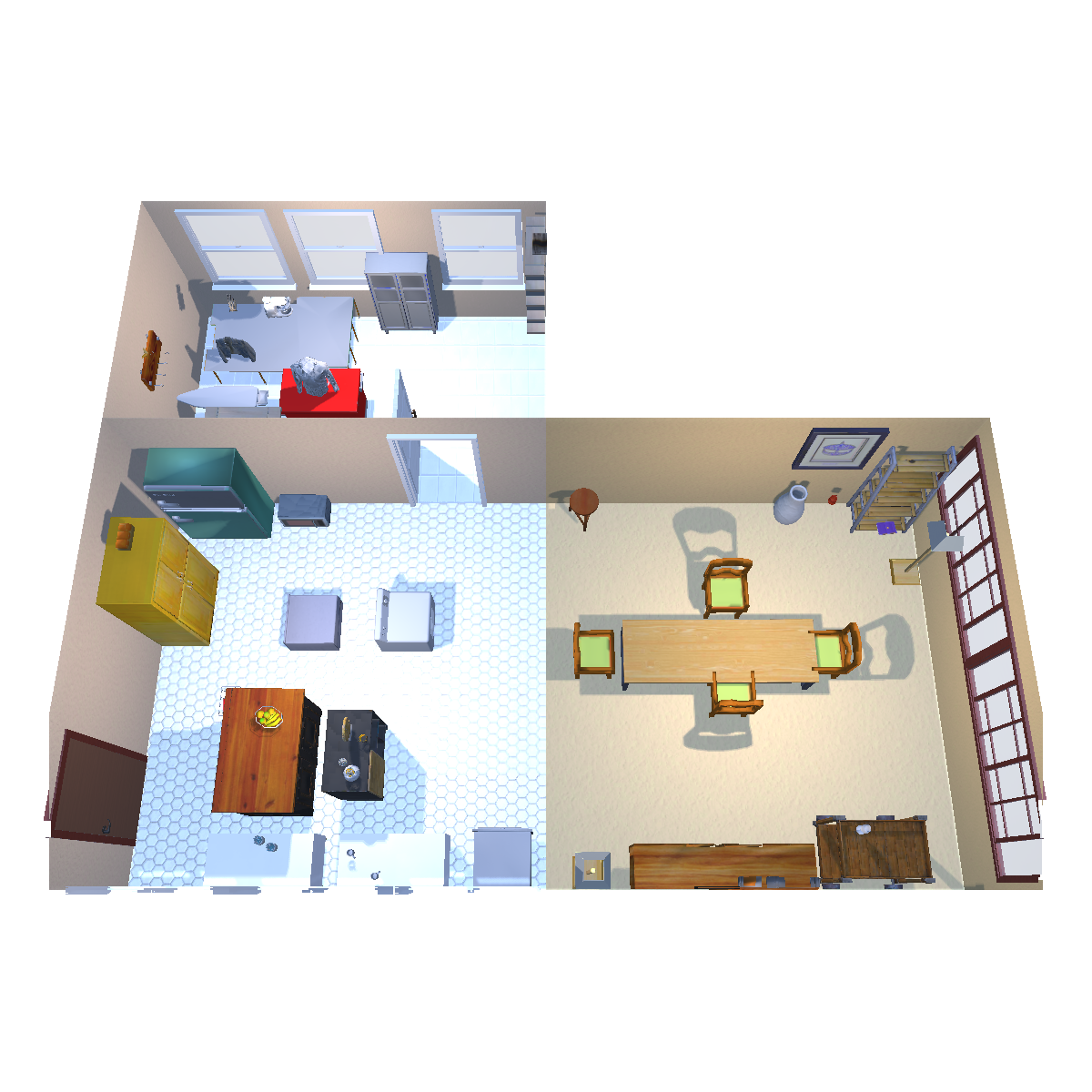}
    \includegraphics[width=0.48\columnwidth]{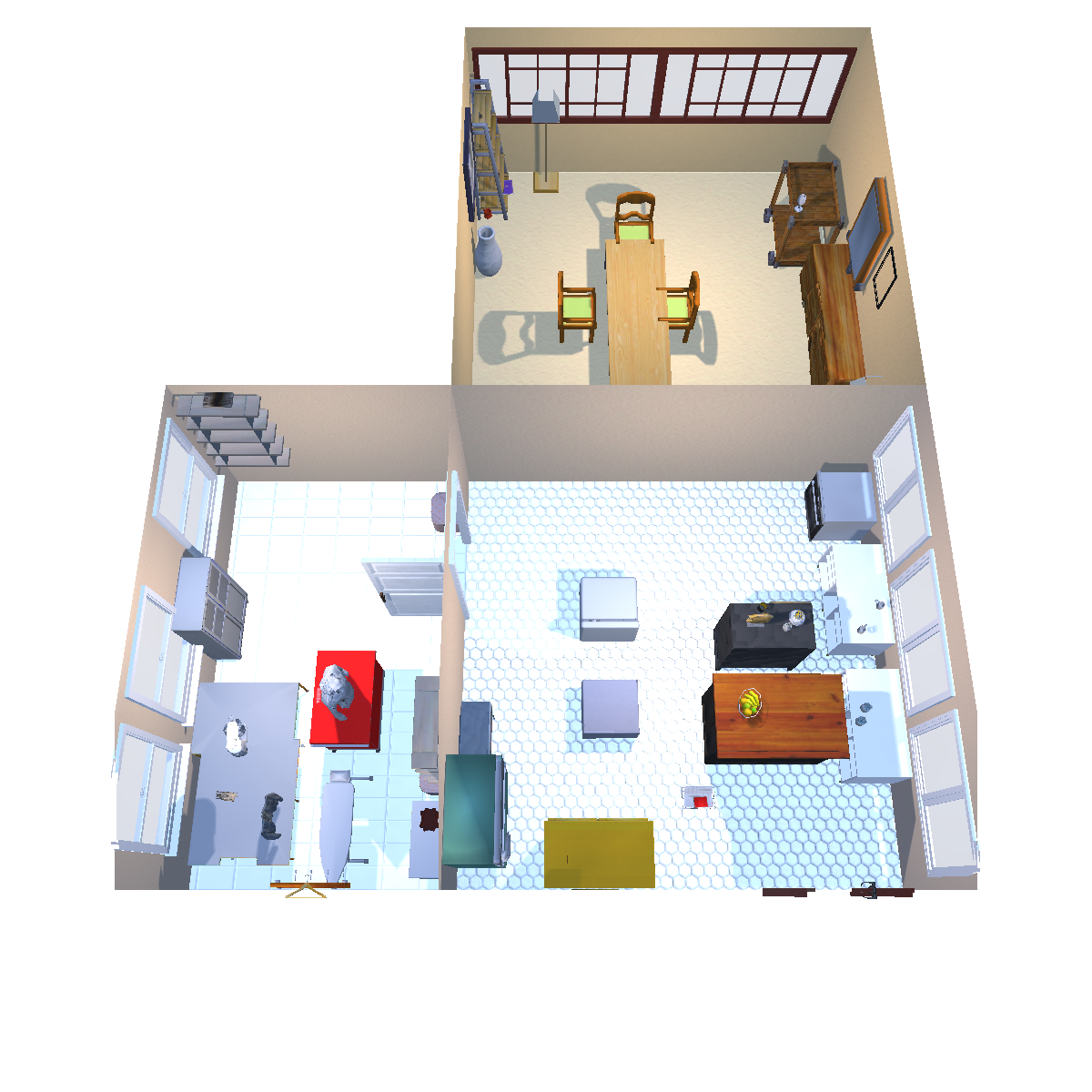}
    
    \vspace{1.5mm}
    
    \includegraphics[width=0.48\columnwidth]{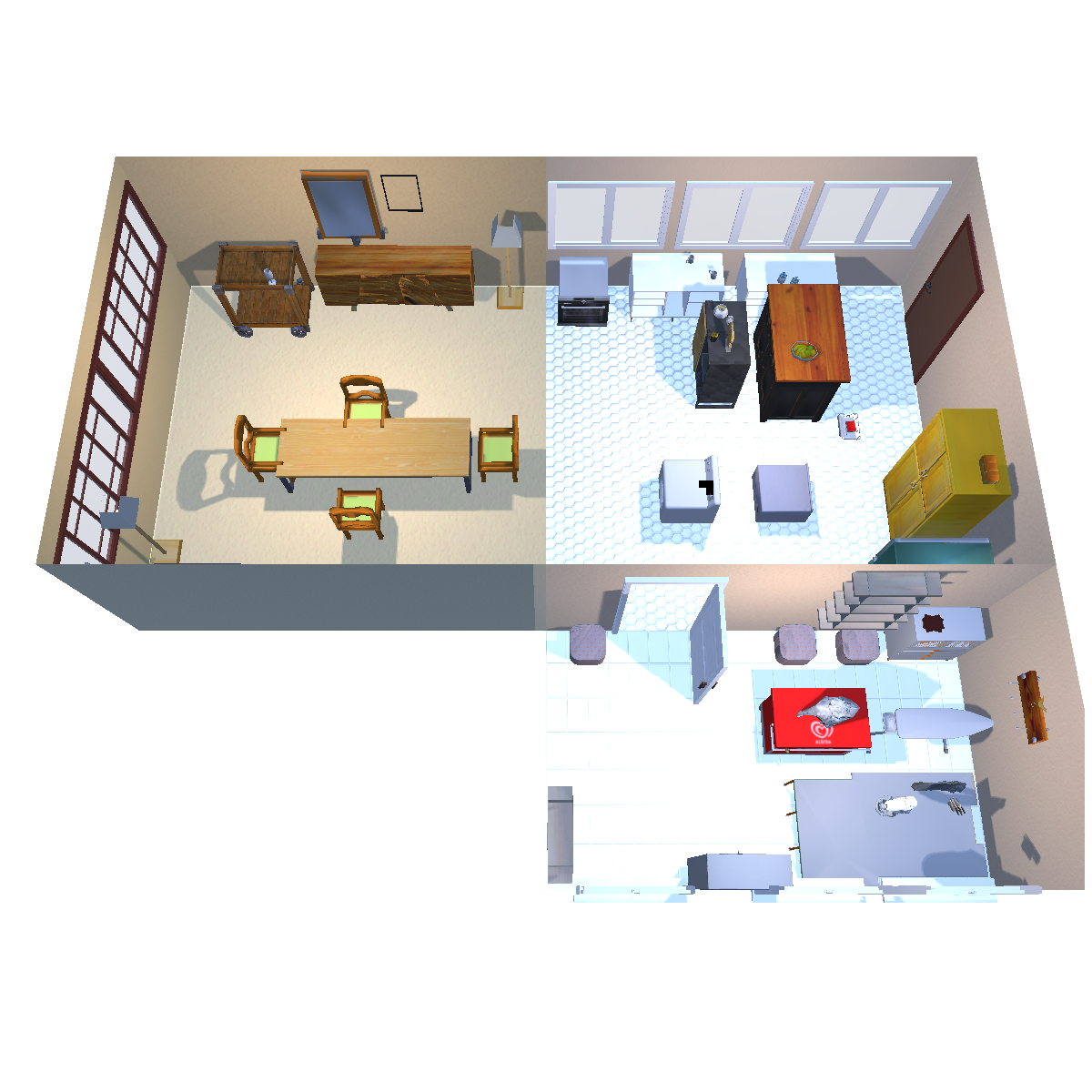}
    \includegraphics[width=0.48\columnwidth]{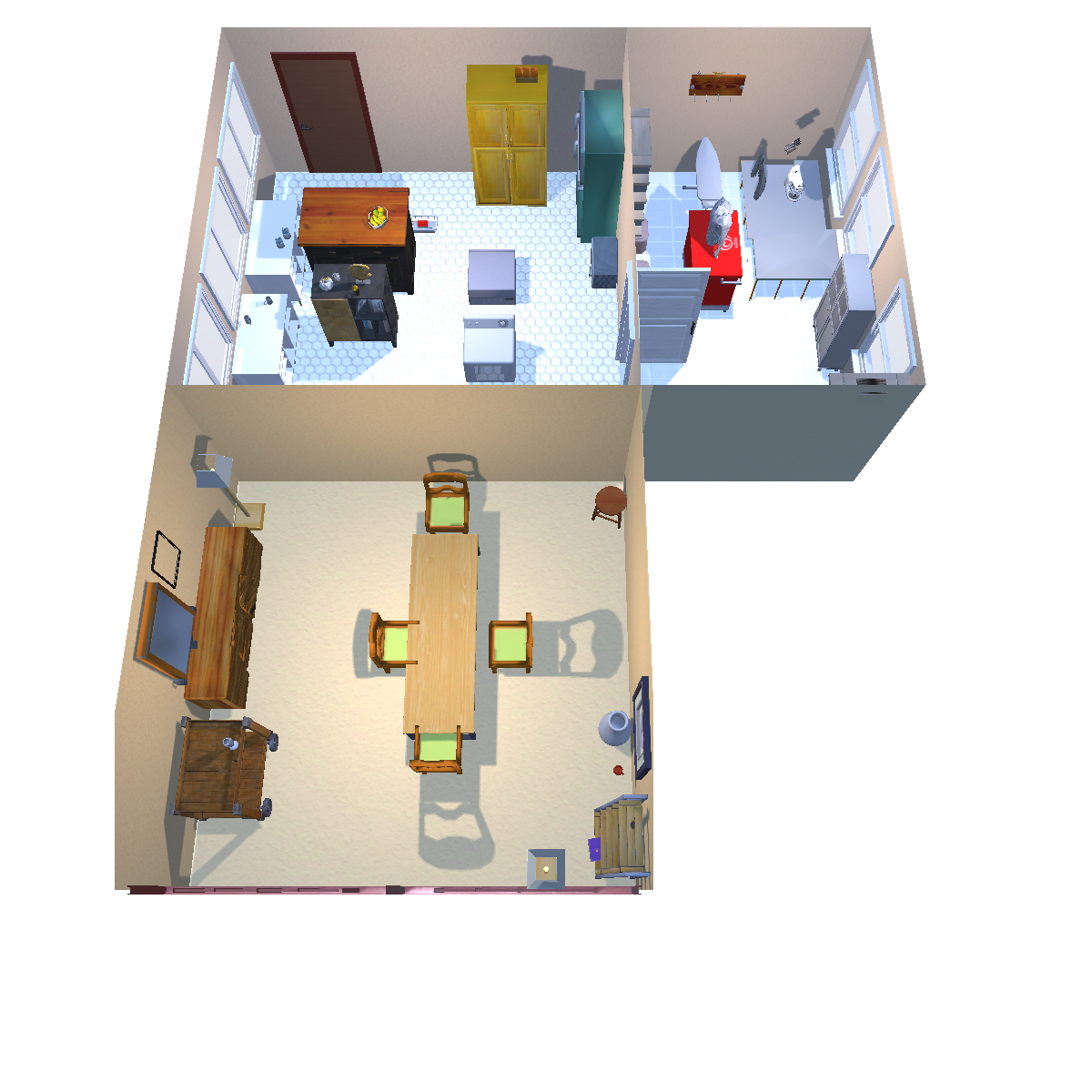}
    
    \caption{Example image input for VLM-as-a-judge}
    \label{appendix:input_VLM-as-a-judge}
\end{figure}

\noindent\textbf{VLM-as-a-judge} prompts VLM to evaluate the generated scene aligning with the instruction. Scene images from four perspectives are provided as model input for fair evaluation. The example input images are shown in Figure~\ref{appendix:input_VLM-as-a-judge}.

\vspace{0.2cm}

\begin{figure}[h]
    \centering
    \includegraphics[width=0.48\textwidth]{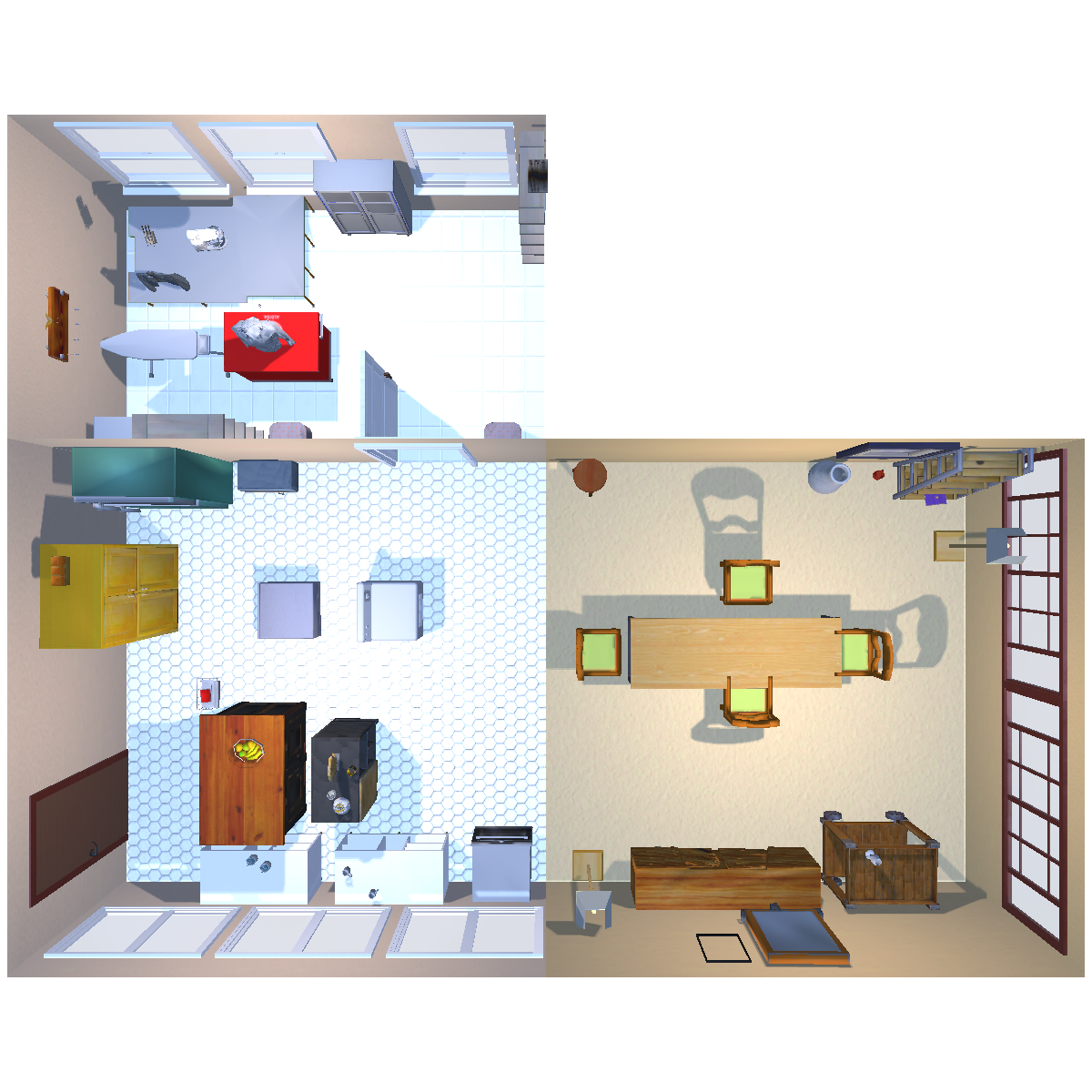}
    \caption{Example image input for CLIPSCORE}
    \label{appendix:input_CLIPSCORE}
\end{figure}

\noindent\textbf{CLIPSCORE}~\citep{hessel2021clipscore} leverages pretrained vision-language models to assess the semantic alignment between generated captions and reference images. Instead of relying on token-level matching or human references, CLIPSCORE computes similarity in the joint embedding space of CLIP. It performs binary judgment on the instruction and top-down scene image using thresholds of 15, 20, and 25. The example input images are shown in Figure~\ref{appendix:input_CLIPSCORE}.

\vspace{0.2cm}

\noindent\textbf{SceneEval}~\citep{tam2025sceneeval} introduces an evaluation framework for text-conditioned 3D indoor scene synthesis. Unlike prior metrics that primarily measure realism or distributional similarity, SceneEval directly evaluates how generated scenes satisfy both explicit user requirements (e.g., object counts, attributes, and spatial relationships) and implicit expectations (e.g., absence of collisions, navigability, accessibility). To support evaluation, the authors release SceneEval-500, a dataset of scene descriptions with annotated ground-truth properties. Together, these metrics provide a comprehensive assessment of fidelity and plausibility in 3D scene generation.

\subsubsection{VLM-as-a-Judge with Different Modules}\label{appendix:VLM-as-a-Judge with Different Modules}
\noindent\textbf{Grounding DINO}~\citep{liu2023grounding} is an open-set object detection model that extends the DINO detector with language grounding capabilities. It takes as input an image and a free-form text prompt describing target objects or scene elements, and outputs bounding boxes corresponding to regions that match the given text. In this work, we use Grounding DINO to localize scene elements mentioned in the instruction within the input scene images. Specifically, we apply it to detect and ground relevant objects, draw bounding boxes on the image, and then feed the annotated image into VLM-as-a-judge.

\vspace{0.2cm}

\noindent\textbf{LASER}~\citep{huang2025laser} is a neuro-symbolic framework that learns spatial-temporal scene graphs from visual inputs under weak supervision. It constructs structured representations that encode objects, their attributes, and their relationships within a scene, enabling explicit reasoning over spatial and temporal dependencies. In our setting, LASER is used to generate scene graphs for each individual room in a multi-room environment. We then provide the scene graphs from all rooms, together with the scene images, as additional inputs to the VLM-as-a-Judge.

\vspace{0.2cm}

\noindent\textbf{Point-LLM}~\citep{guo2023pointbindpointllmaligning} is a framework that enables large language models to understand and reason over 3D point cloud data. It encodes point cloud inputs into representations that can be processed by LLMs, allowing the model to perform spatial reasoning in 3D space. In our approach, we first convert the scene graphs into point cloud representations. We then uniformly downsample the point clouds to match the input size requirements. Finally, we use Point-LLM to evaluate the resulting 3D scene representations.

\subsection{Scene Synthesis}\label{appendix:Scene Generation Methods}
\subsubsection{Overview of Scene Synthesis Methods}
\noindent\textbf{LayoutGPT}~\citep{feng2023layoutgpt} introduces a training-free framework that leverages large language models as visual planners for layout-based generation. LayoutGPT composes in-context visual demonstrations in CSS-like structures to inject visual commonsense into LLMs. This enables accurate translation of challenging linguistic concepts—such as numerical and spatial reasoning—into 2D image layouts and 3D indoor scene arrangements.

\vspace{0.2cm}

\noindent\textbf{I-Design}~\citep{ccelen2025design}  introduces a personalized framework for 3D indoor scene synthesis driven by natural language input. I-Design employs a team of large language model agents to interpret unstructured user descriptions and reason about object selection, style, and spatial relationships. These preferences are represented as scene graphs, which are then transformed into complete room layouts through a backtracking placement algorithm and enriched with 3D assets retrieved from object databases. The system outputs interpretable, editable pipelines including scene graphs, floor plans, and rendered views, enabling flexible, user-centered interior design exploration.

\vspace{0.2cm}

\noindent\textbf{Holodeck}~\citep{yang2024holodeck} introduces a controllable simulation platform for embodied agents in diverse 3D environments. Unlike prior systems that often limit interaction fidelity or domain variety, Holodeck enables agents to operate within photorealistic virtual worlds featuring rich physics, object manipulation, and dynamic scenarios. The platform provides flexible interfaces for integrating natural language commands, sensory input, and reinforcement learning frameworks, making it suitable for studying grounded reasoning and task execution.

\vspace{0.2cm}

\noindent\textbf{LayoutVLM}~\citep{sun2025layoutvlm} introduces a framework for open-universe 3D layout generation guided by natural language instructions. Unlike prior approaches that either predict precise object poses or solve rigid constraint systems, LayoutVLM combines numerical pose estimates and spatial relations within a differentiable optimization process to achieve physically plausible and semantically coherent layouts. The method leverages vision-language models with visual prompting and a self-consistent decoding procedure to generate scene layout representations from unlabeled 3D assets and rendered images.

\vspace{0.2cm}

\noindent\textbf{MANSION}~\citep{che2026mansion} introduces a framework for generating multi-floor 3D indoor environments from natural language instructions. Unlike prior work that focuses on single-floor scene synthesis or local object arrangement, MANSION explicitly models hierarchical building structures to ensure global spatial consistency across rooms and floors. It generates structurally valid and navigable environments while preserving inter-floor connectivity and semantic coherence, and further supports large-scale procedural generation with open-vocabulary scene customization for embodied AI evaluation.

\subsubsection{Implementation of Scene Synthesis Methods}\label{appendix:implementation of generation methods}
\noindent\textbf{Scene layout generation.} I-Design, LayoutVLM, and LayoutGPT do not generate structural components such as walls or floors. Therefore, we use Holodeck to generate the basic room layout (i.e., walls and floors), and the compared methods are applied solely for object selection and/or object placement within each room.

Additionally, \eval~ currently supports only square or rectangular room layouts. To adhere to this evaluation setting, we modify the generation pipeline of MANSION so that it produces only rectangular or square scene layouts.
 
\vspace{0.2cm}

\noindent\textbf{Object Selection.} For LayoutVLM, although its supplementary material provides prompts for object selection, these prompts are designed for test-time data generation and are not part of the core LayoutVLM framework. Moreover, LayoutVLM relies on a human-in-the-loop procedure for object selection, making it incompatible with fully automated benchmarking. Therefore, we do not use these prompts. Instead, we provide the model with object names and asset UIDs selected by Holodeck.

For LayoutGPT, the model selects objects to place based on a given prompt; however, it is restricted to a fixed set of 21 object categories. In contrast, LEGO covers 51.7K object categories, which cannot be enumerated in a single prompt. To address this, we include only the categories selected by Holodeck in the prompt, allowing LayoutGPT to choose objects from this reduced set.

\section{Extended Related Work}
\paragraph{Embodied AI simulators and sim-to-real gap problem.}
Training and evaluating embodied agents directly in the real world is costly, time-consuming, and often unsafe~\citep{liu2025aligning}, motivating the use of simulation as a scalable substitute for real-world interaction data~\citep{shridhar2020alfred, li2023behavior, puig2024habitat, mittal2025isaac, Genesis}.
However, this reliance on simulation inevitably introduces a sim-to-real gap~\citep{zhao2020sim}, as discrepancies in visual appearance, physical dynamics, and scene configuration between simulated and real environments can degrade the performance of agents once deployed.

To narrow this sim-to-real gap, a growing body of work has developed simulators that provide increasingly diverse and realistic scenes~\citep{straub2019replica, deitke2020robothor, li2023behavior, khanna2024habitat}, enabling agents to be trained and evaluated across a broader range of environments and configurations.
Such diverse and realistic scenes, in turn, are essential for supporting a wide range of downstream embodied tasks, from everyday manipulation~\citep{brohan2022rt, brohan2023rt, driess2023palm, o2024open} and navigation~\citep{ahn2022can,shah2023lm,han2025dialnav} to safety-critical decision making~\citep{yin2024safeagentbench, choi2026embguard} and personalized assistance~\citep{wu2023tidybot, kwon2025embodied}, each of which depends on the simulator faithfully reflecting the specific conditions those tasks require.
Our work builds on this direction by accelerating the construction of such diverse, task-relevant scenes directly from user instructions, and by evaluating whether the resulting scenes preserve the fine-grained properties those instructions specify.

\newpage
\section{Examples of Tool Image Outputs}

\clearpage
\onecolumn
\section{Prompts Used in Our Work}

\begin{figure*}[!b]
    \centering
    \includegraphics[width=0.95\textwidth]{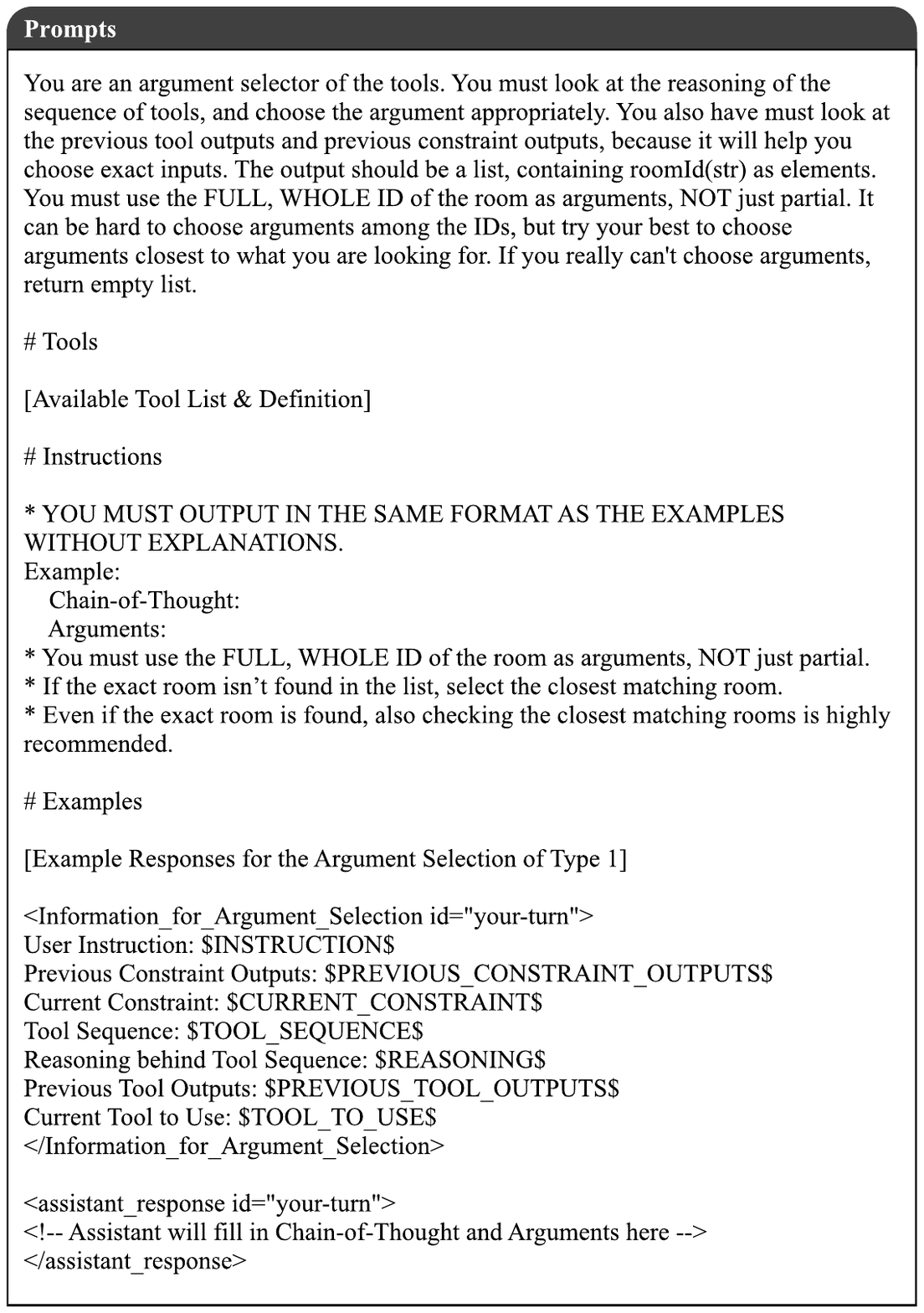}
    \caption{Prompt used for Argument Selection in scene-component ID list generation tools.}
    \label{appendix:prompt:ArgumentSelector_type_1}
\end{figure*}

\begin{figure*}[h]
    \centering
    \includegraphics[width=0.95\linewidth]{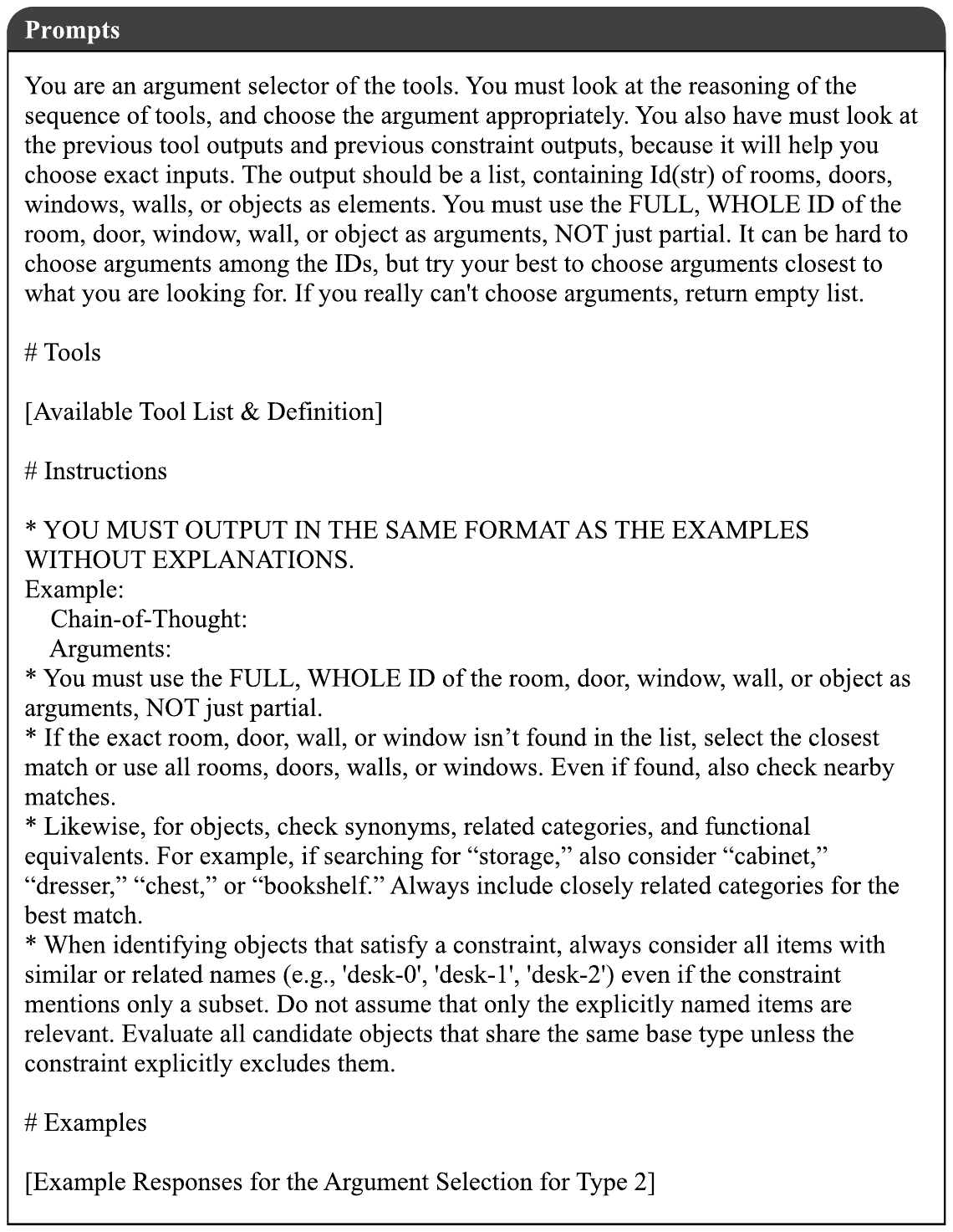}
    \caption{Prompt used for argument selection in scene-component information retrieval tools.}
    \label{appendix:prompt:ArgumentSelector_type_2_1}
\end{figure*}

\begin{figure*}[h]
    \centering
    \includegraphics[width=0.95\textwidth]{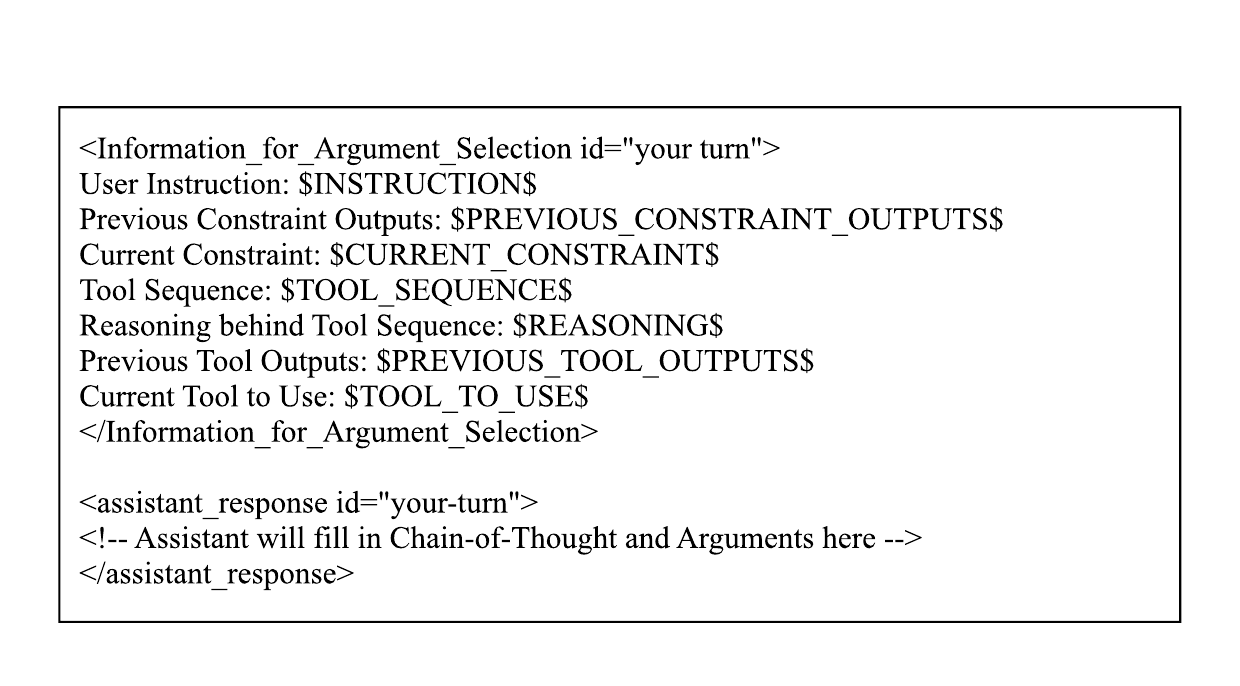}
    \caption{Prompt used for argument selection in scene-component information retrieval tools.}
    \label{appendix:prompt:ArgumentSelector_type_2_2}
    \addtocounter{figure}{-1}
\end{figure*}

\begin{figure*}[h]
    \centering
    \includegraphics[width=0.95\textwidth]{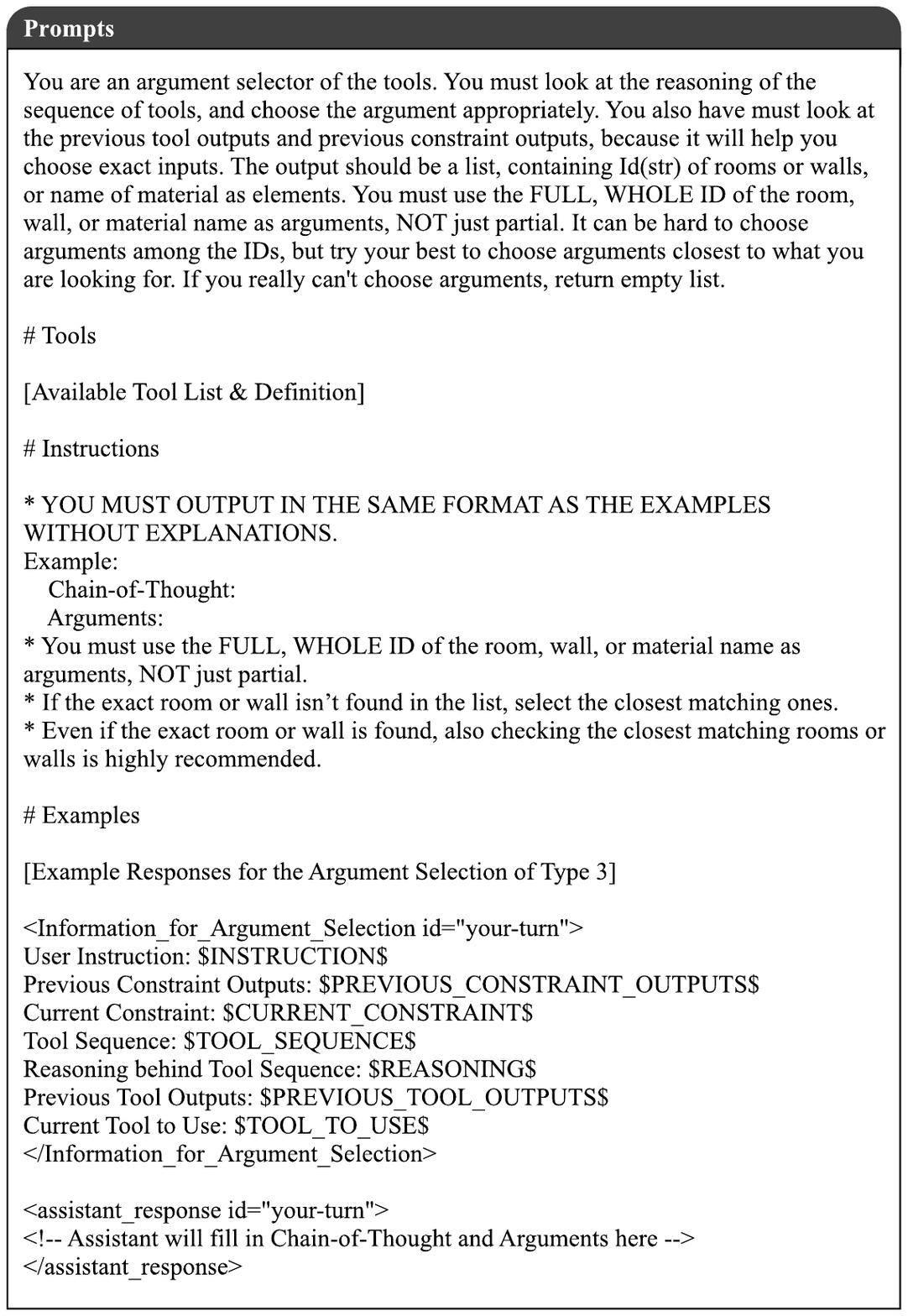}
    \caption{Prompt used for Argument Selection in scene-component visual rendering tools.}
    \label{appendix:prompt:ArgumentSelector_type_3}
\end{figure*}

\begin{figure*}[h]
    \centering
    \includegraphics[width=0.95\textwidth]{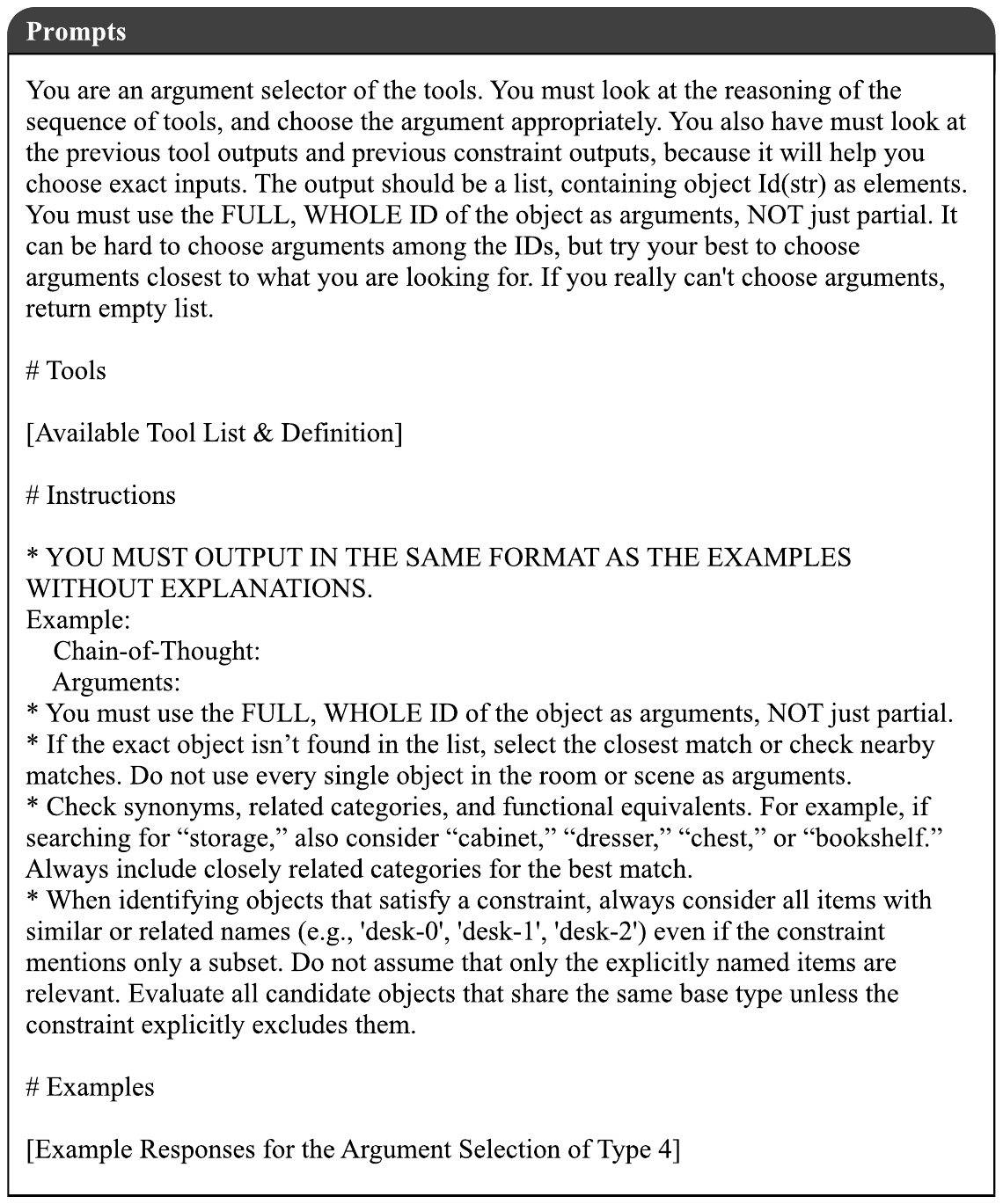}
    \caption{Prompt used for Argument Selection in object-level visual rendering tools.}
    \label{appendix:prompt:ArgumentSelector_type_4_1}
\end{figure*}

\begin{figure*}[h]
    \centering
    \includegraphics[width=0.95\textwidth]{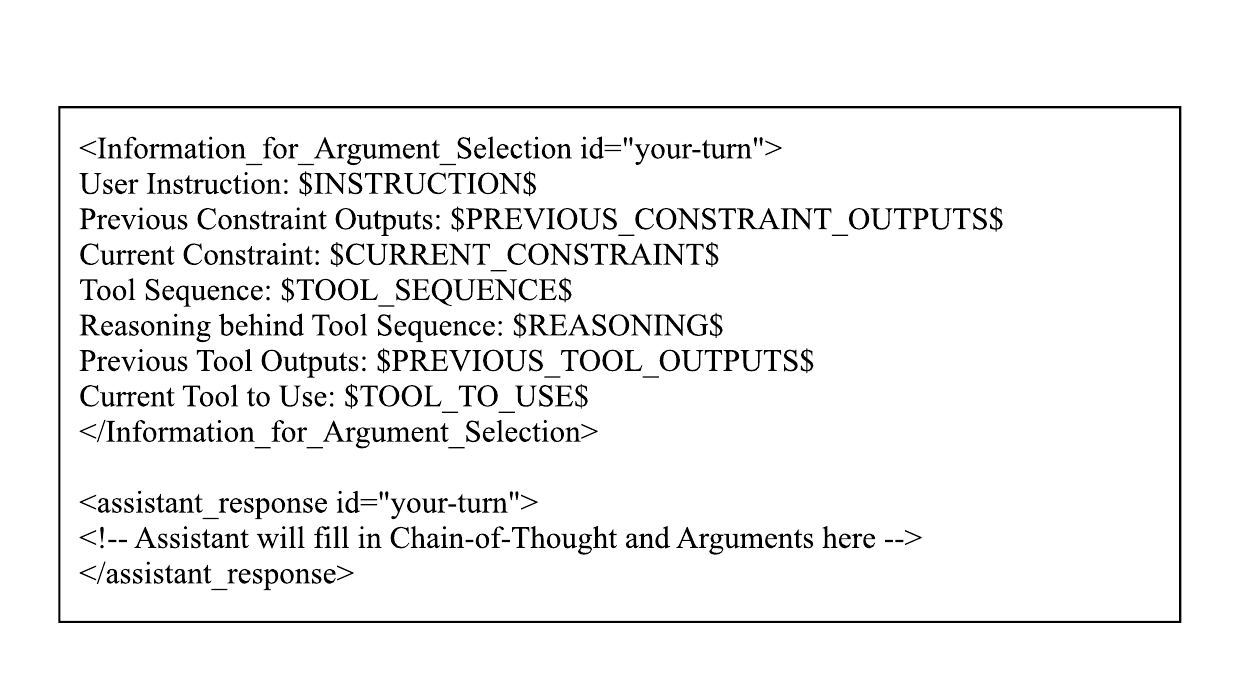}
    \caption{Prompt used for Argument Selection in object-level visual rendering tools.}
    \label{appendix:prompt:ArgumentSelector_type_4_2}
    \addtocounter{figure}{-1}
\end{figure*}

\begin{figure*}[h]
    \centering
    \includegraphics[width=0.95\textwidth]{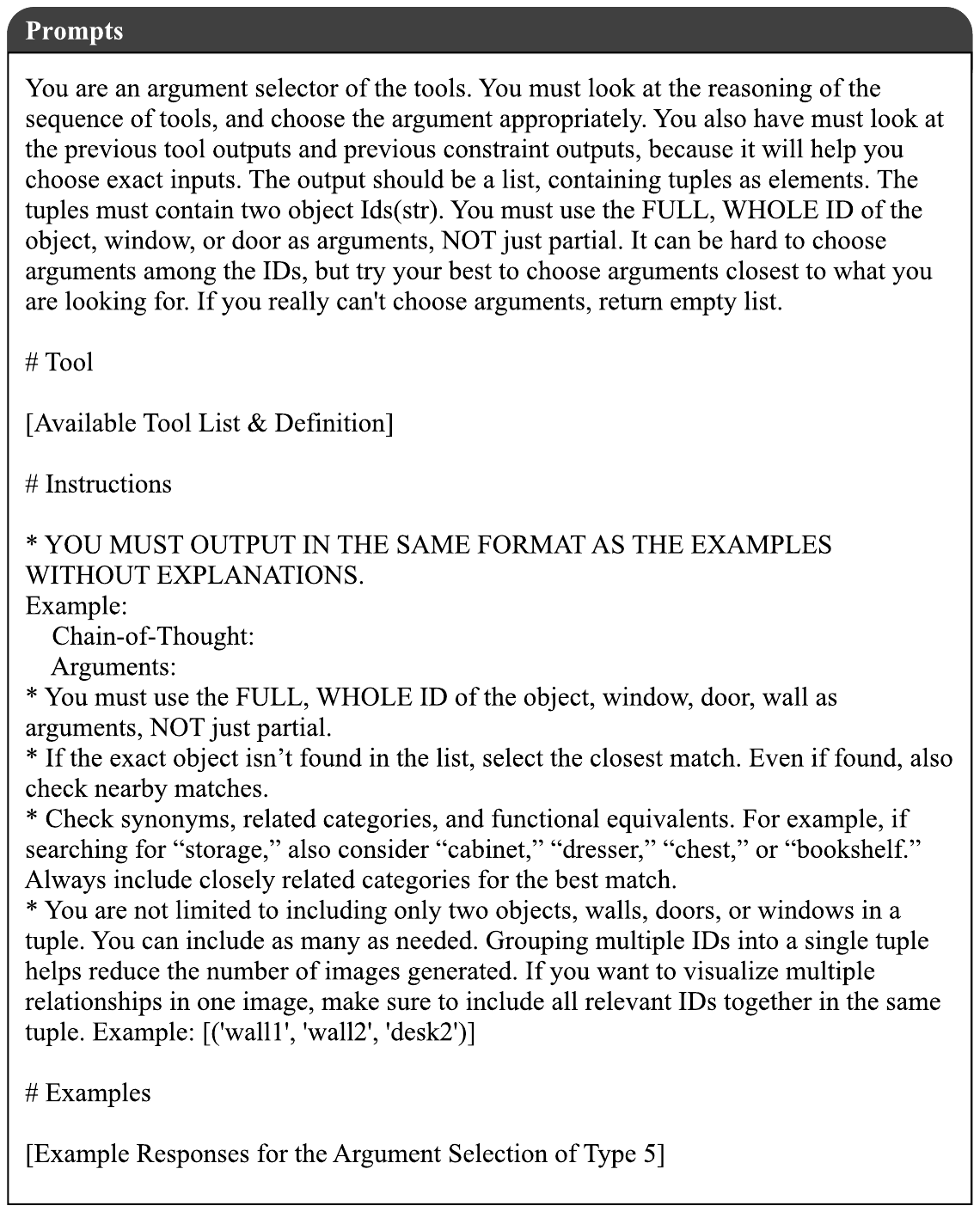}
    \caption{Prompt used for Argument Selection in the spatial-relation visualization tool.}
    \label{appendix:prompt:ArgumentSelector_type_5_1}
\end{figure*}

\begin{figure*}[h]
    \centering
    \includegraphics[width=0.95\textwidth]{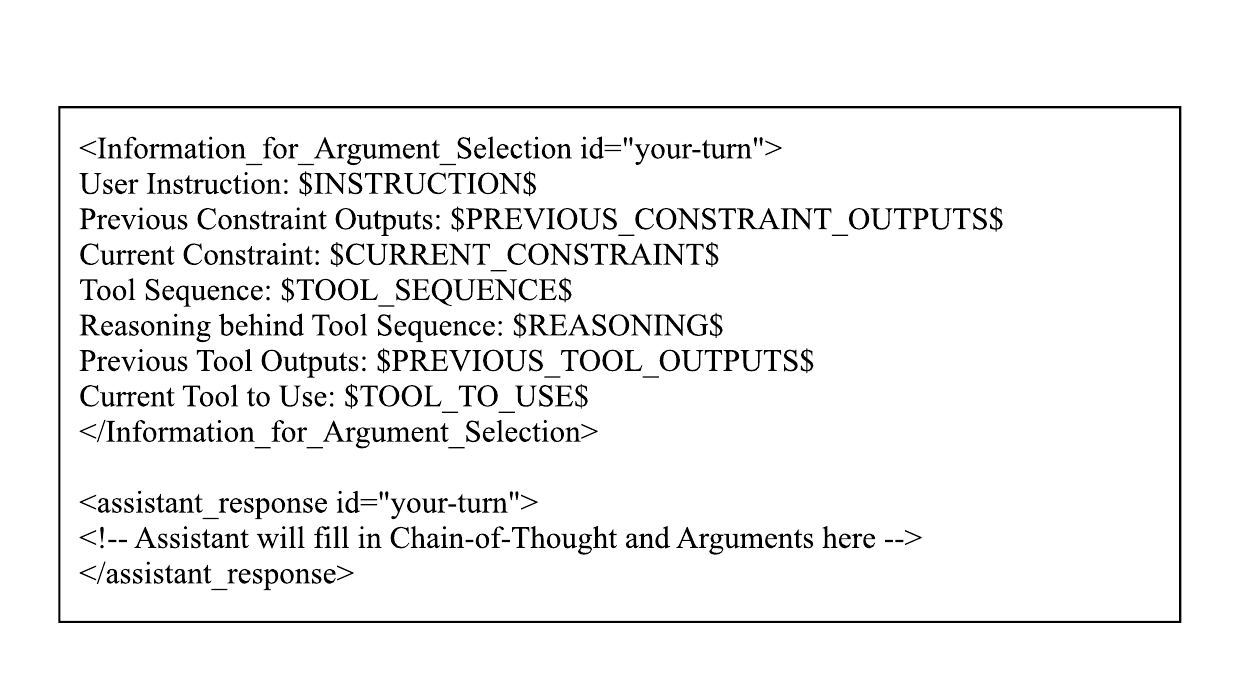}
    \caption{Prompt used for Argument Selection in the spatial-relation visualization tool.}
    \label{appendix:prompt:ArgumentSelector_type_5_2}
    \addtocounter{figure}{-1}
\end{figure*}
\begin{figure*}[h]
    \centering
    \includegraphics[width=0.95\textwidth]{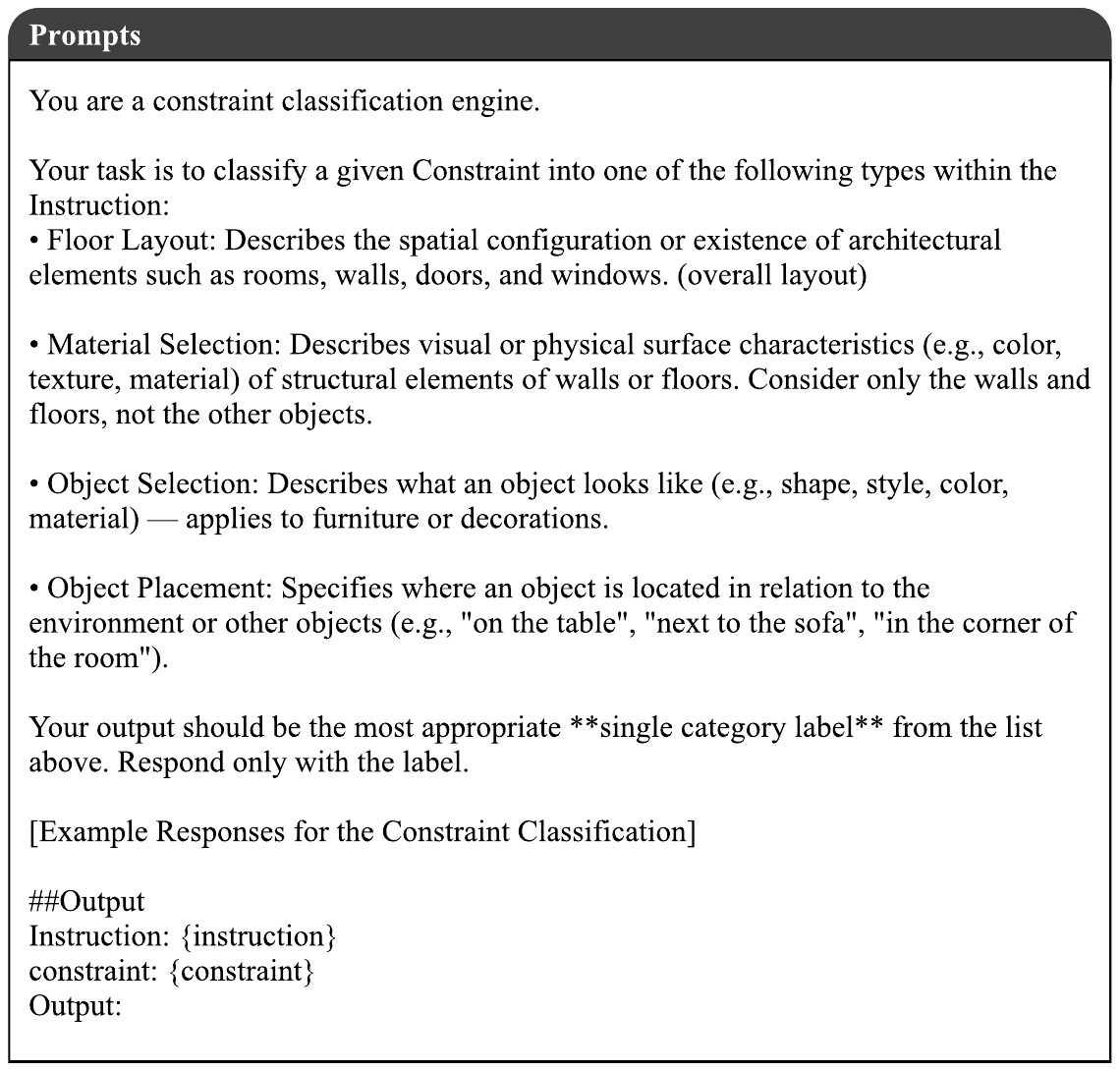}
    \caption{Prompt used for Constraint Classification.}
    \label{appendix:prompt:ConstraintClassification}
\end{figure*}

\begin{figure*}[h]
    \centering
    \includegraphics[width=0.95\textwidth]{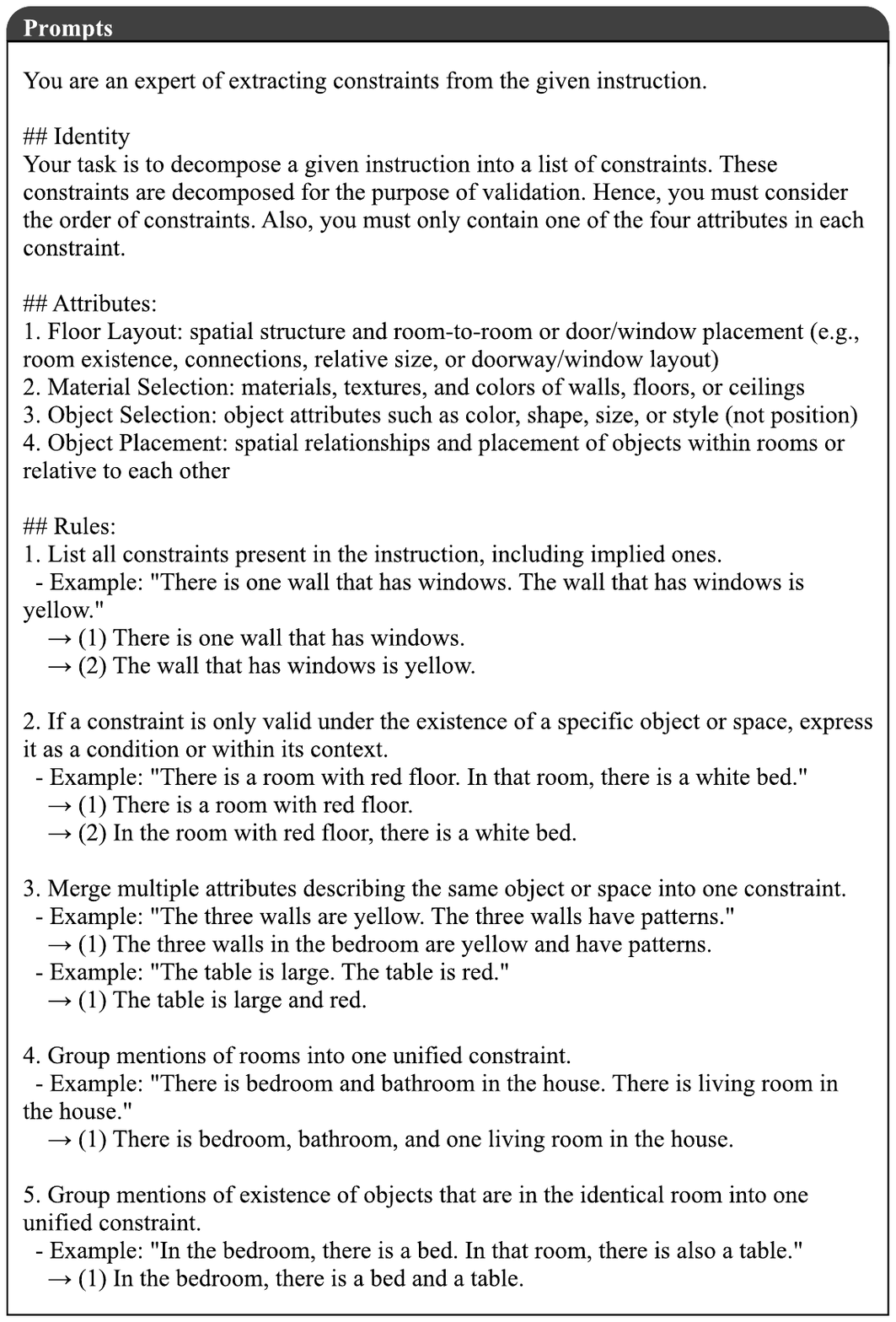}
    \caption{Prompt used for Constraint Classification.}
    \label{appendix:prompt:ConstraintIdentification_2}
\end{figure*}

\begin{figure*}[h]
    \centering
    \includegraphics[width=0.95\textwidth]{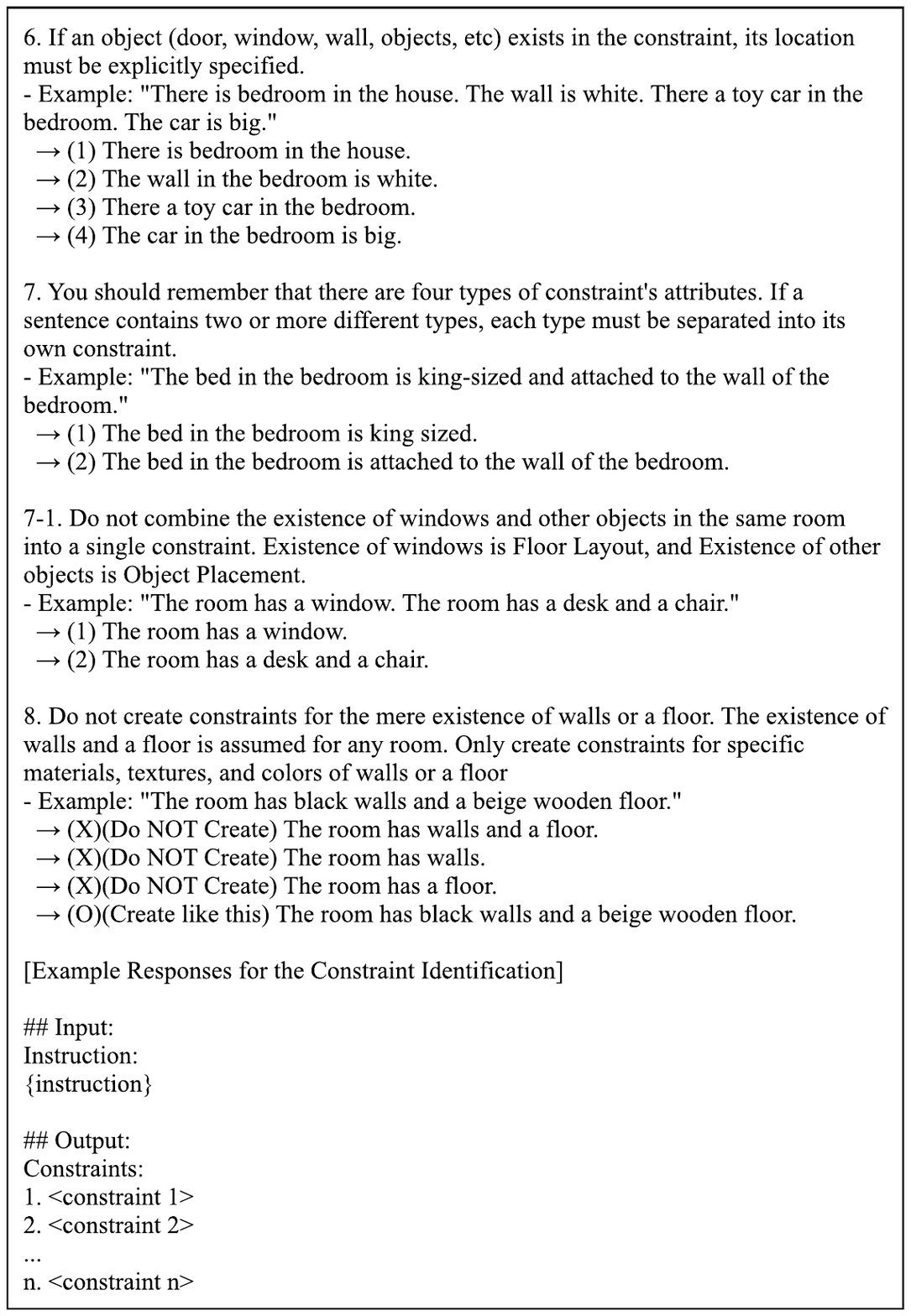}
    \caption{Prompt used for Constraint Identification.}
    \label{appendix:prompt:ConstraintIdentification}
    \addtocounter{figure}{-1}
\end{figure*}

\begin{figure*}[h]
    \centering
    \includegraphics[width=0.95\textwidth]{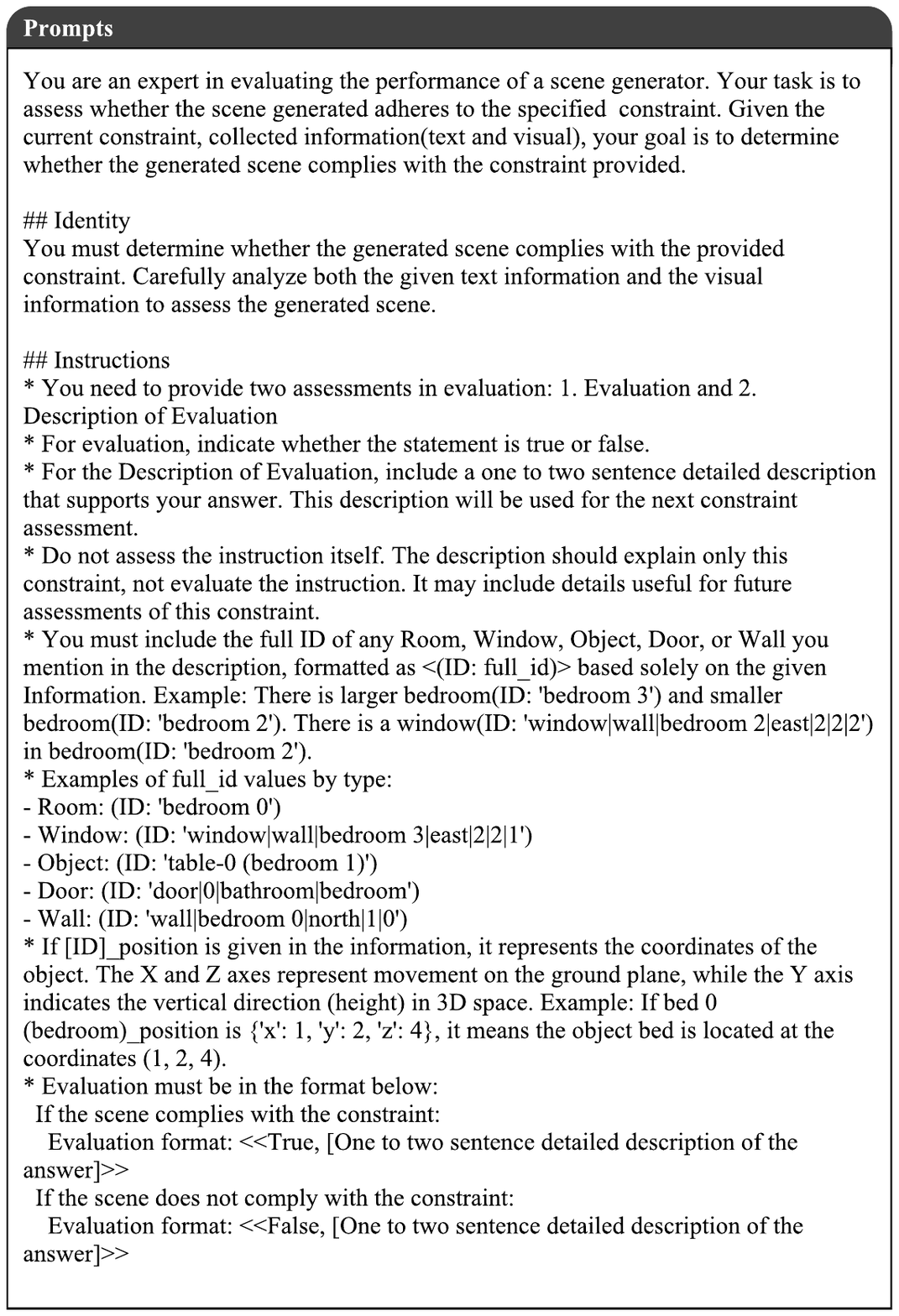}
    \caption{Prompt used in Constraint Validation for Floor Layout constraints.}
    \label{appendix:prompt:Validation_FL_1}
\end{figure*}

\begin{figure*}[h]
    \centering
    \includegraphics[width=0.95\textwidth]{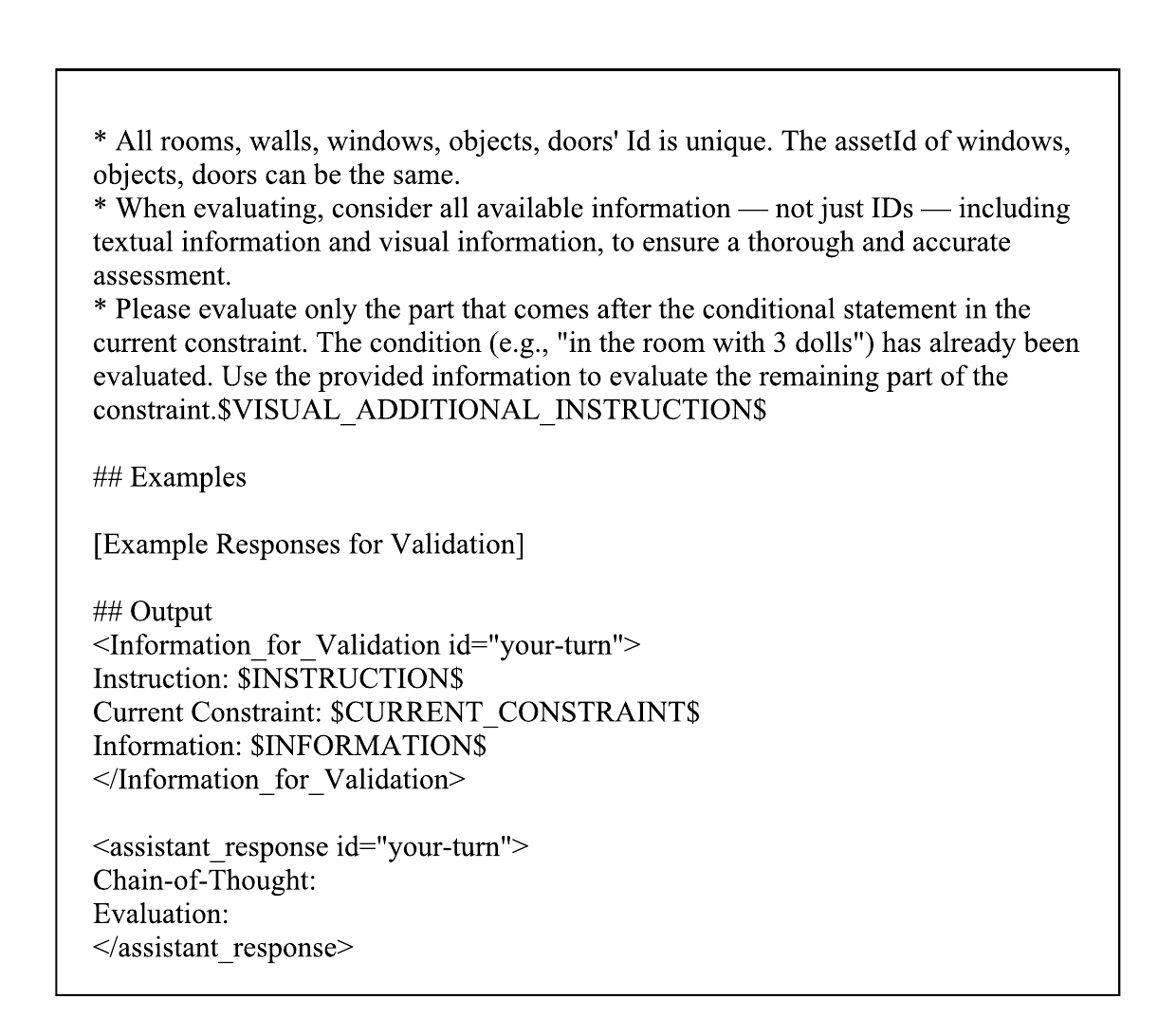}
    \caption{Prompt used in Constraint Validation for Floor Layout constraints.}
    \label{appendix:prompt:Validation_FL_2}
    \addtocounter{figure}{-1}
\end{figure*}

\begin{figure*}[h]
    \centering
    \includegraphics[width=0.95\textwidth]{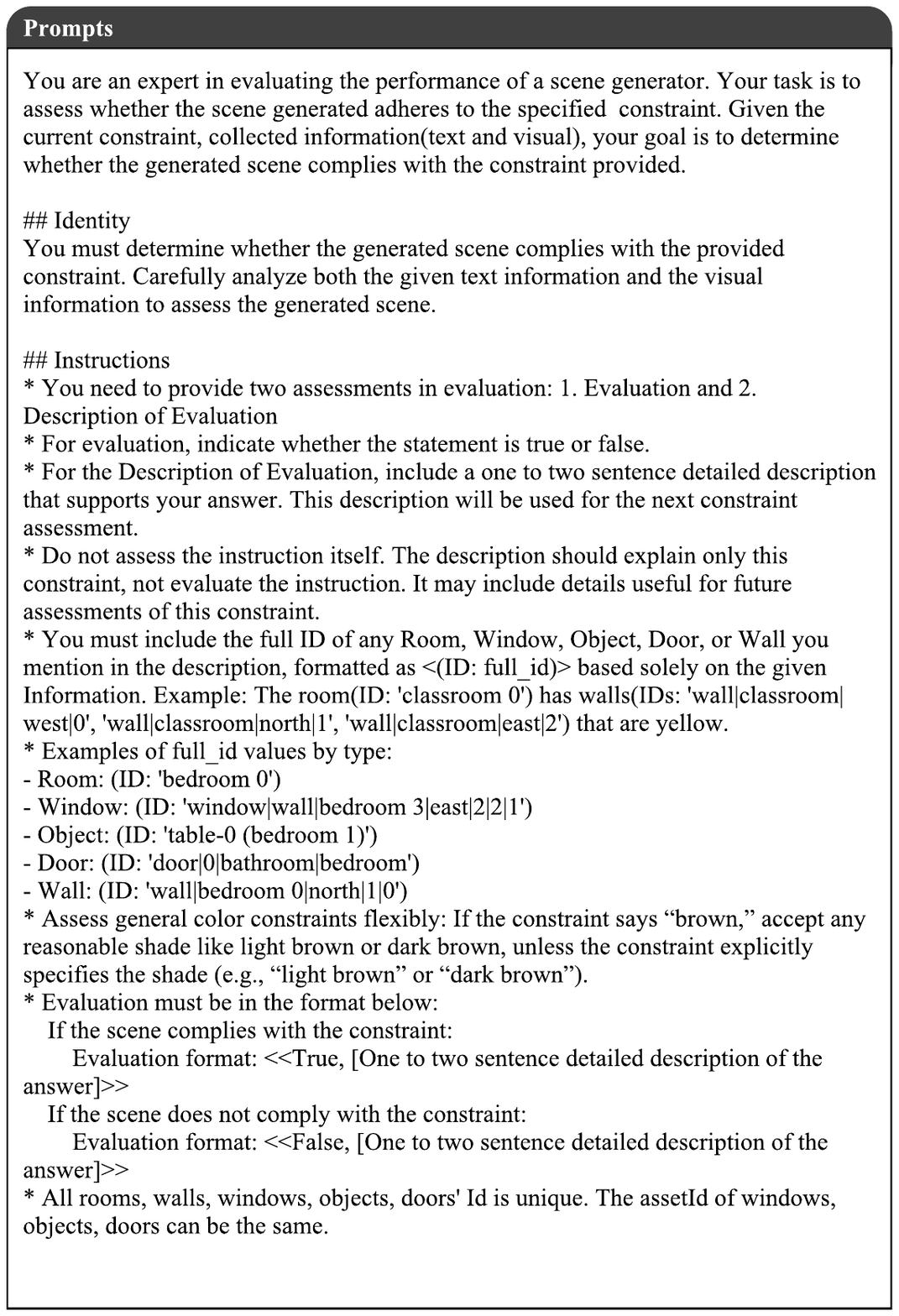}
    \caption{Prompt used in Constraint Validation for Material Selection constraints.}
    \label{appendix:prompt:Validation_MS_1}
\end{figure*}

\begin{figure*}[h]
    \centering
    \includegraphics[width=0.95\textwidth]{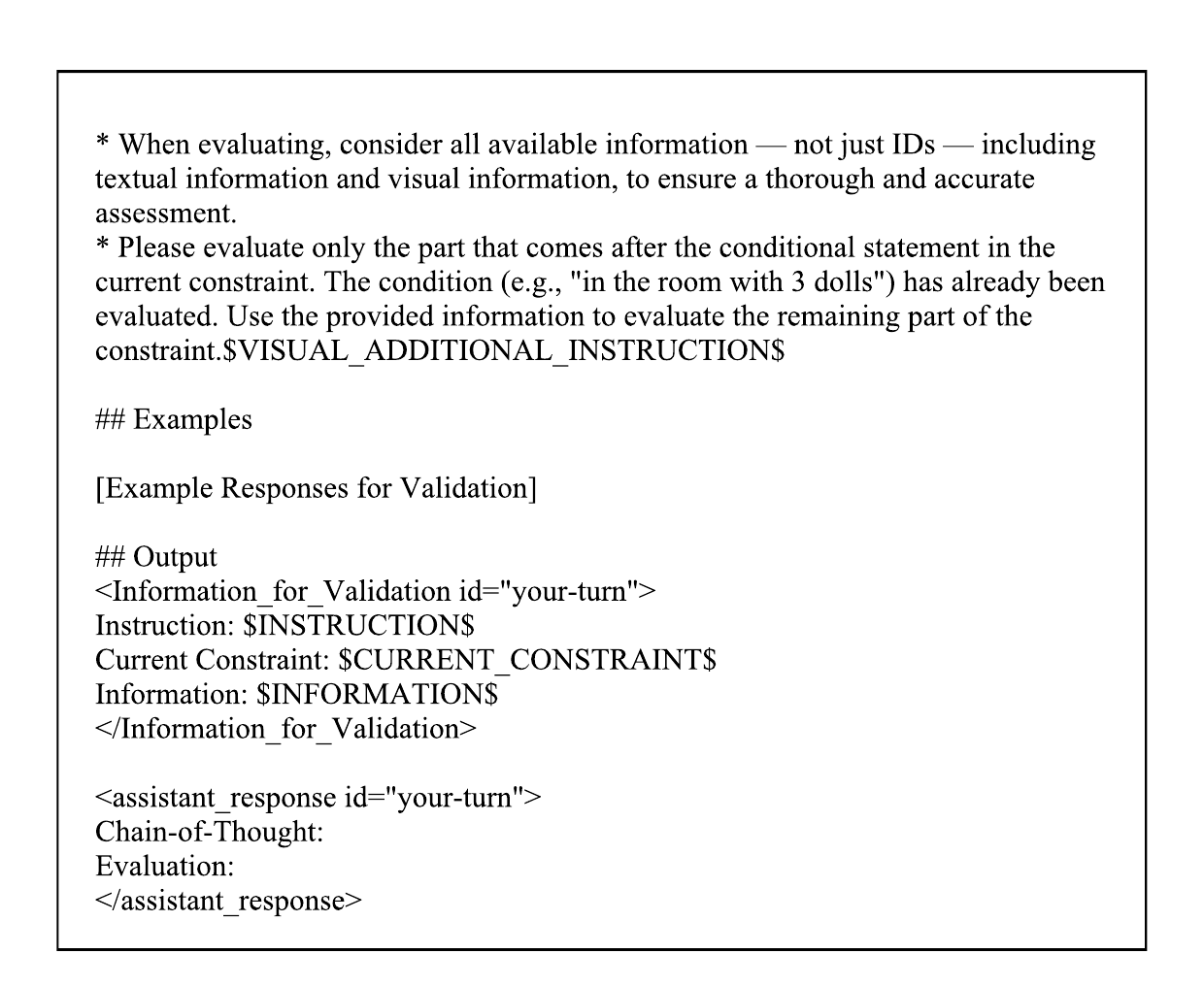}
    \caption{Prompt used in Constraint Validation for Material Selection constraints.}
    \label{appendix:prompt:Validation_MS_2}
    \addtocounter{figure}{-1}
\end{figure*}

\begin{figure*}[h]
    \centering
    \includegraphics[width=0.95\textwidth]{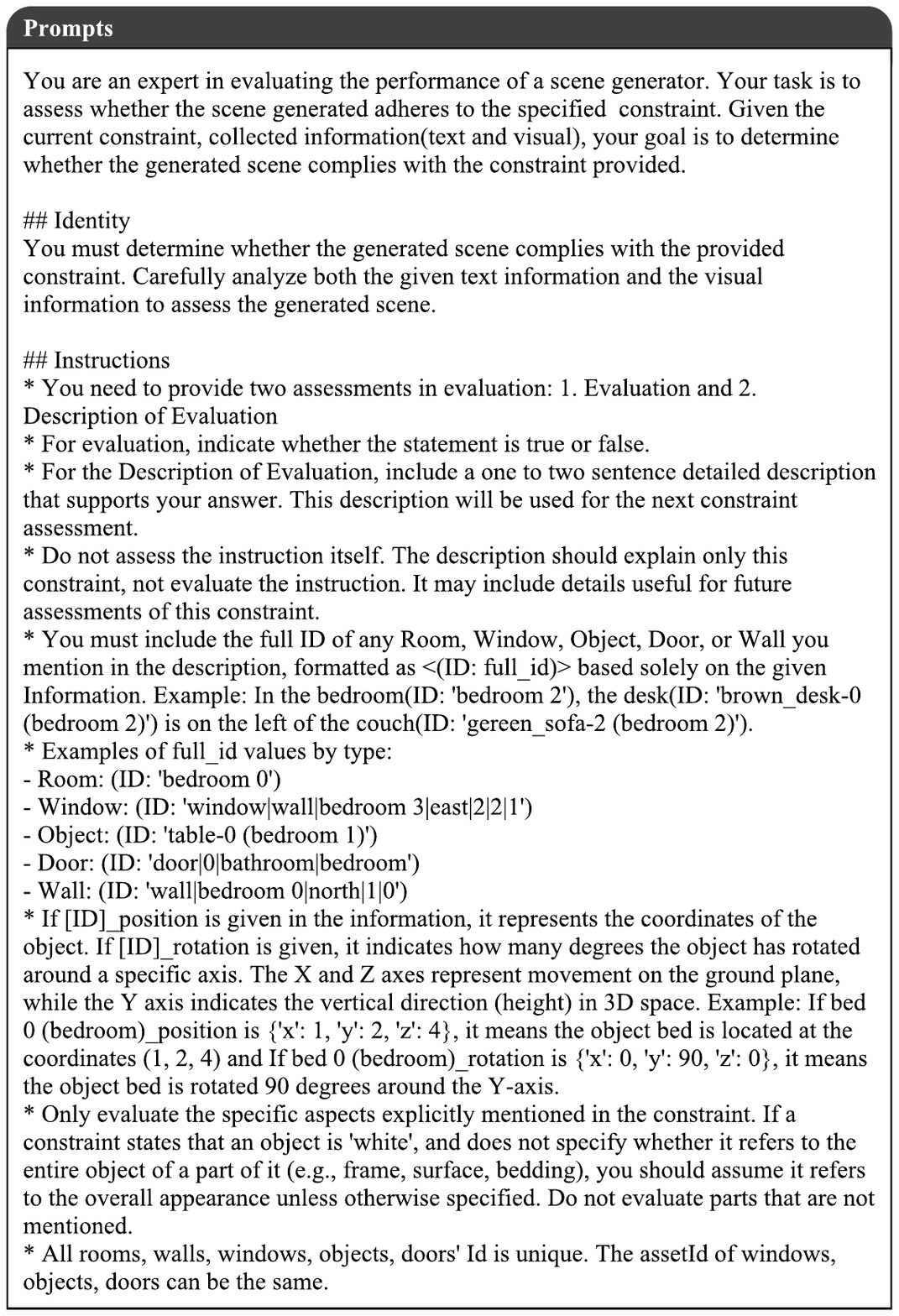}
    \caption{Prompt used in Constraint Validation for Object Placement constraints.}
    \label{appendix:prompt:Validation_OP_1}
\end{figure*}

\begin{figure*}[h]
    \centering
    \includegraphics[width=0.95\textwidth]{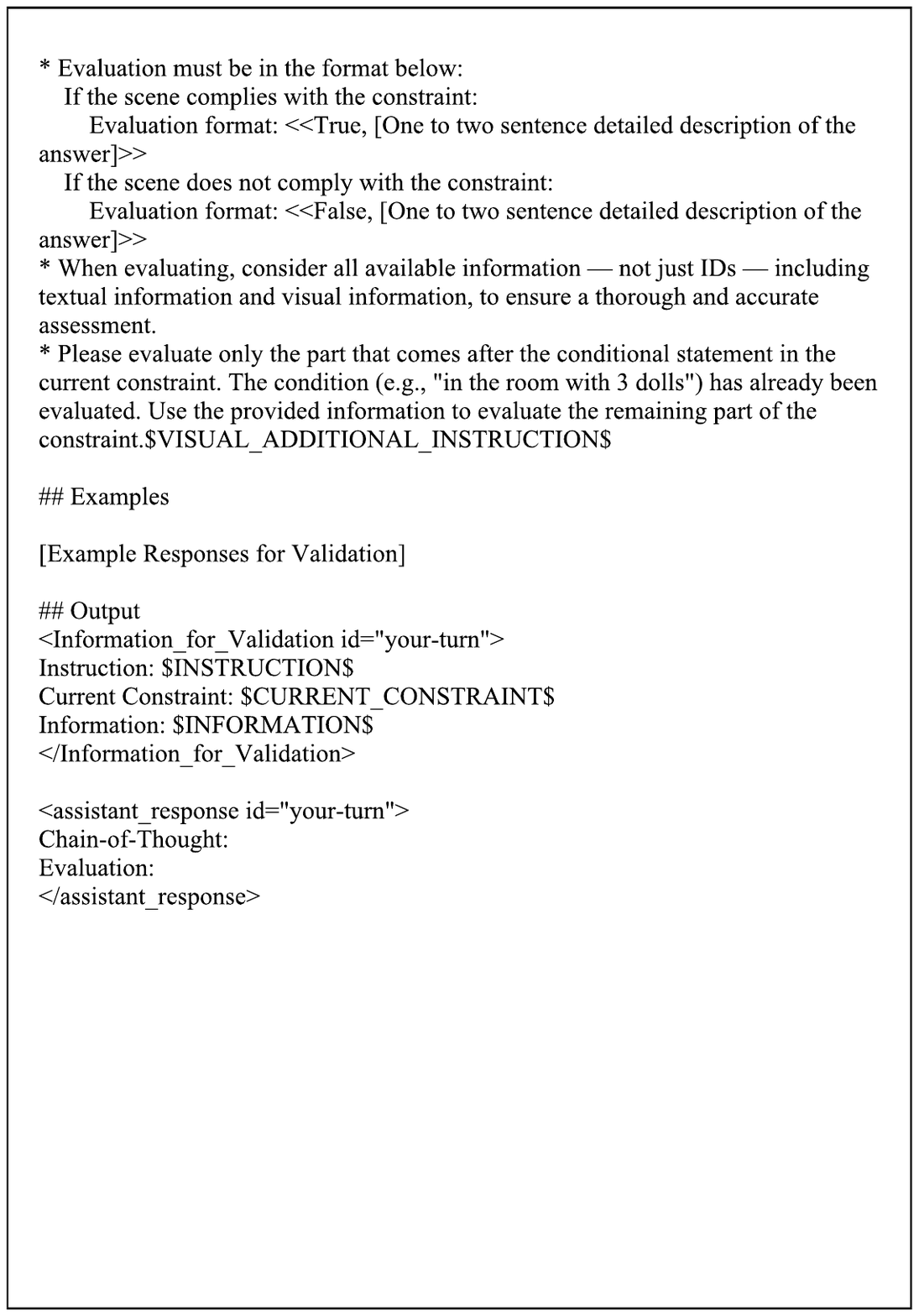}
    \caption{Prompt used in Constraint Validation for Object Placement constraints.}
    \label{appendix:prompt:Validation_OP_2}
    \addtocounter{figure}{-1}
\end{figure*}

\begin{figure*}[h]
    \centering
    \includegraphics[width=0.95\textwidth]{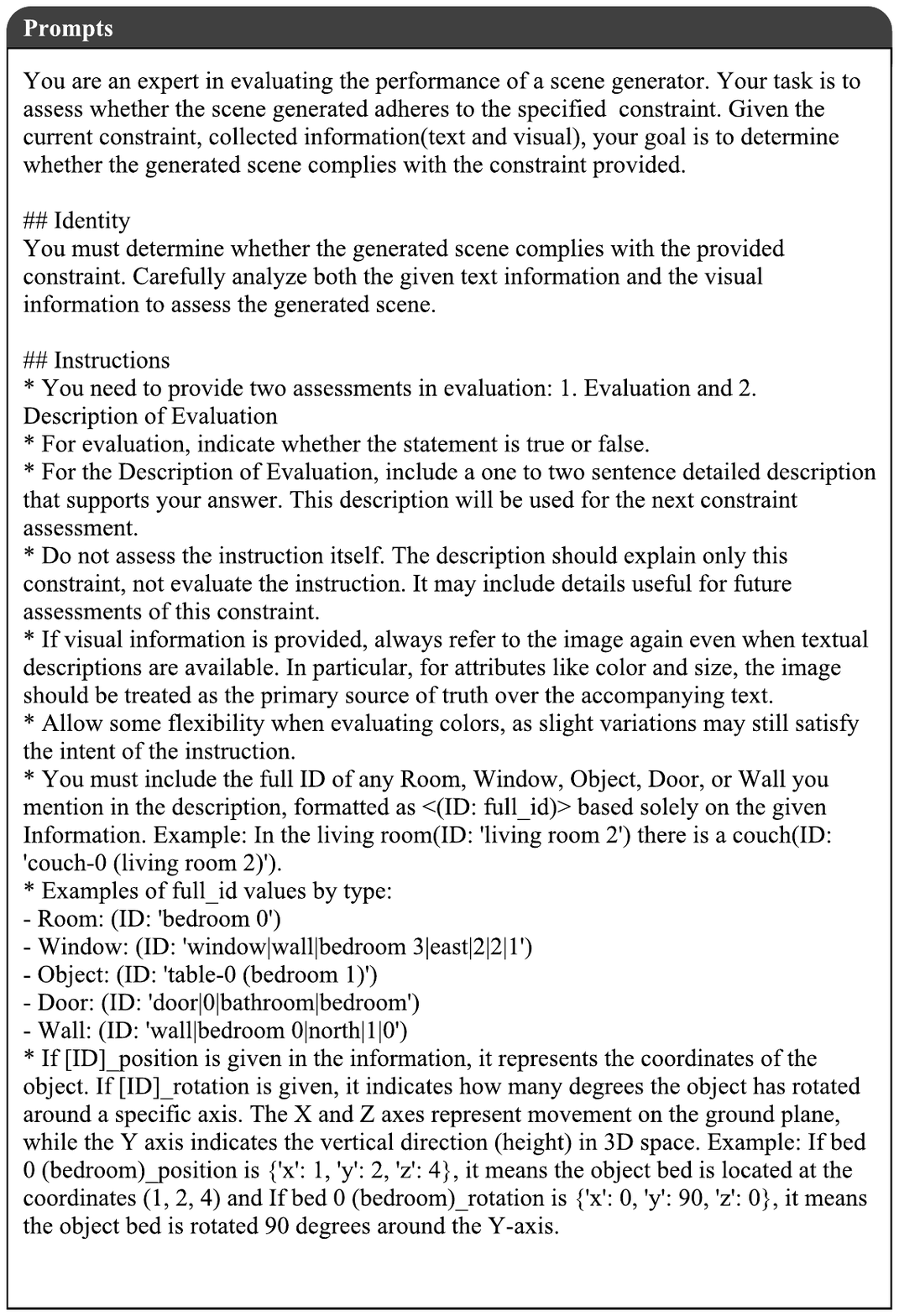}
    \caption{Prompt used in Constraint Validation for Object Selection constraints.}
    \label{appendix:prompt:Validation_OS_1}
\end{figure*}

\begin{figure*}[h]
    \centering
    \includegraphics[width=0.95\textwidth]{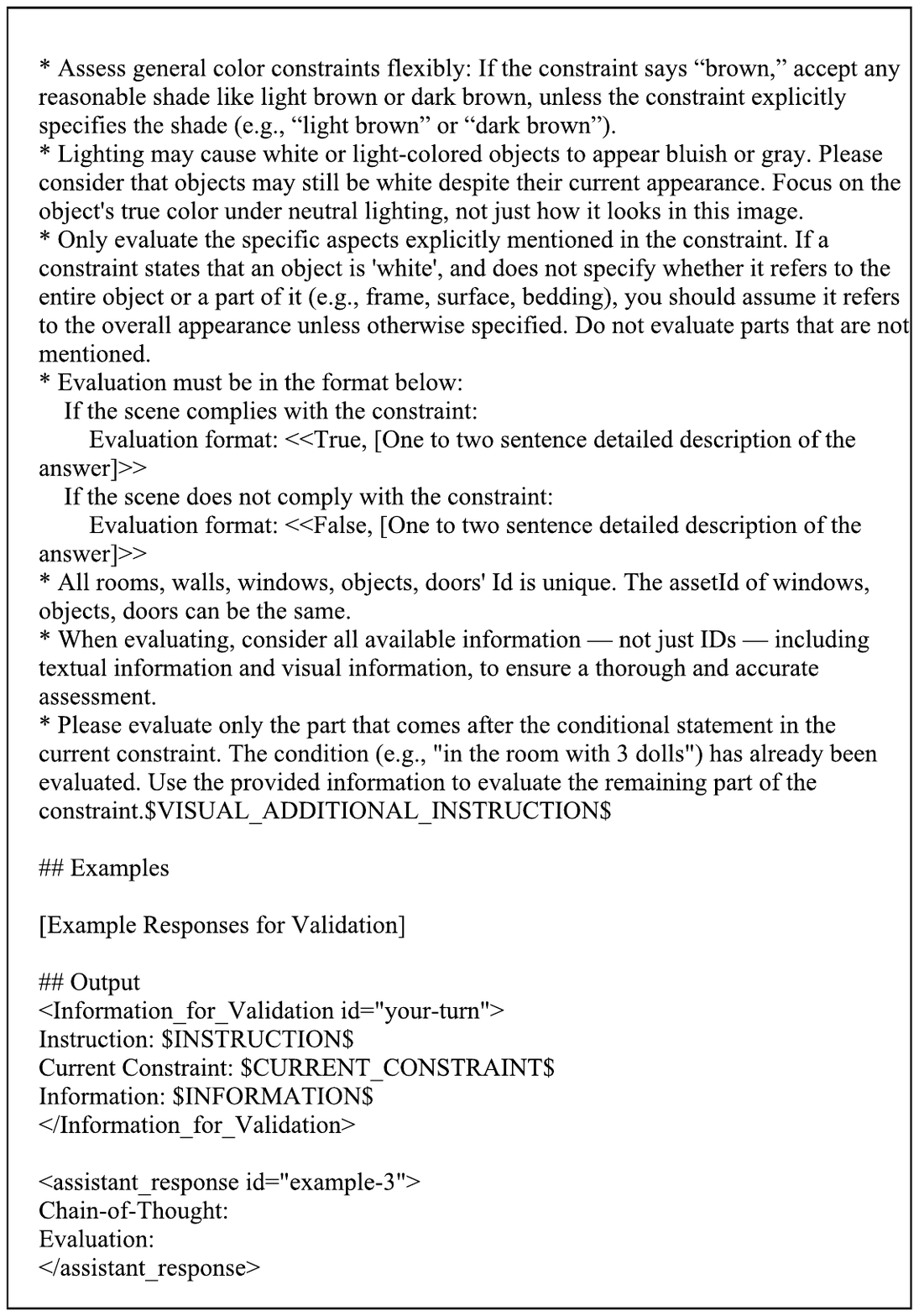}
    \caption{Prompt used in Constraint Validation for Object Selection constraints.}
    \label{appendix:prompt:Validation_OS_2}
    \addtocounter{figure}{-1}
\end{figure*}
\begin{figure*}[h]
    \centering
    \includegraphics[width=0.95\textwidth]{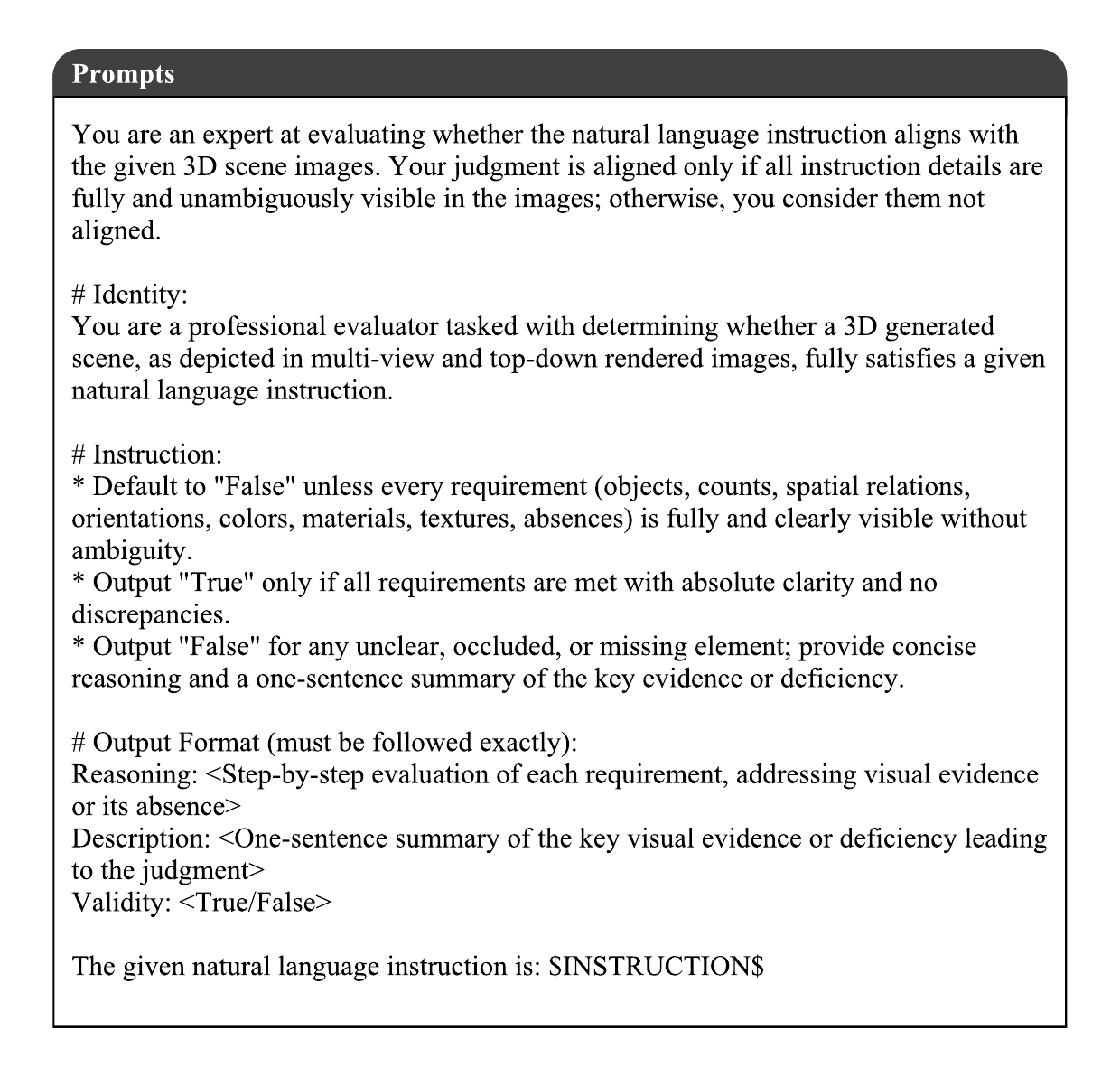}
    \caption{Prompt used for baseline VLM evaluation of instruction–scene validity}
    \label{appendix:prompt:VLM_Eval}
\end{figure*}

\end{document}